\documentclass[10pt,twocolumn,letterpaper]{article}

\usepackage[pagenumbers]{cvpr} 
\usepackage{graphicx}
\usepackage{siunitx}
\usepackage{enumitem}
\usepackage{multirow}
\usepackage{fp}
\usepackage{amsmath}
\usepackage{amssymb}
\usepackage{amssymb}
\usepackage{textcomp}
\usepackage{etoolbox}
\usepackage{algorithm}
\usepackage{algorithmicx}
\usepackage{algpseudocode}
\usepackage{wrapfig}
\usepackage{textcomp, gensymb}
\usepackage{calc}
\usepackage{booktabs}       
\usepackage{amsfonts}       
\usepackage{wrapfig}
\usepackage{graphicx}
\usepackage{caption}
\usepackage{subcaption}
\usepackage{booktabs}
\usepackage{multirow}
\usepackage{colortbl}
\usepackage{float}
\usepackage[most]{tcolorbox}  
\usepackage[super]{nth}
\usepackage{tikz}
\usepackage{pgfplots}
\usepackage{stmaryrd}
\usepackage{soul}
\usepackage{graphicx}   
\usepackage{placeins}
\usepackage{pifont}

\usepackage{amssymb} 
\usepackage{graphicx} 
\usepackage{resizegather} 
\usepackage{booktabs,multirow,makecell,array}

\usepackage[normalem]{ulem}
\usepackage{afterpage}
\usepackage{float}
\usepackage{fontawesome5}  
\usepackage{xspace}         
\usepackage{pifont}                  
\definecolor{indianred}{rgb}{0.8, 0.36, 0.36}
\definecolor{bleudefrance}{rgb}{0.19, 0.55, 0.91}
\definecolor{forestgreen}{rgb}{0.0, 0.5, 0.0}
\definecolor{ashgrey}{rgb}{0.7, 0.75, 0.71}
\definecolor{darkorange}{rgb}{1.0, 0.55, 0.0}
\definecolor{darkorchid}{rgb}{0.6, 0.2, 0.8}
\definecolor{ashgrey}{rgb}{0.7, 0.75, 0.71}
\definecolor{backred}{rgb}{1.0, 0.6, 0.6}
\definecolor{wdcolor}{RGB}{128, 0, 255}

\DeclareRobustCommand{\MTEMpc}{\texorpdfstring{\ensuremath{\mathrm{MT\text{-}EM}\%}}{MT-EM\%}}

\definecolor{wd_question_color}{RGB}{255, 0, 0}

\usepackage{pifont}
\newcommand{\cmark}{{\color{forestgreen}\ding{51}}}
\newcommand{\xmark}{{\color{red}\ding{55}}}

\usepackage{graphicx}  
\usepackage{pifont}    
\makeatletter

\newcommand{\wdorangecircle}{%
  {\color{ashgrey}%
    \mbox{%
      \ooalign{%
        \hfil\scalebox{1.18}{$\bigcirc$}\hfil\cr  
        \hfil\scalebox{1.08}{$\bigcirc$}\hfil\cr  
      }%
    }%
  }%
}

\DeclareRobustCommand{\icoyes}{%
  \wdorangecircle
  \xspace}

\DeclareRobustCommand{\icono}{%
  {\color{forestgreen}\ding{51}}%
  \xspace}

\DeclareRobustCommand{\icohalf}{%
  \mbox{%
    \wdorangecircle
    \kern-0.9em
    \raisebox{0.05ex}{\color{forestgreen}\ding{51}}%
  }%
  \xspace}

\makeatother

\newcommand{\Gray}[0]{\rowcolor{gray!20}}
\newcommand{\Lgray}[0]{\rowcolor{gray!10}}

\definecolor{cvprblue}{rgb}{0.21,0.49,0.74}
\usepackage[pagebackref,breaklinks,colorlinks,allcolors=cvprblue]{hyperref}

\def\confName{CVPR}
\def\confYear{2026}

\title{A Benchmark for Ultra-High-Resolution Remote Sensing MLLMs}

\author{
Yunkai Dang\thanks{Equal contribution.}\textsuperscript{1},
Meiyi Zhu\footnotemark[1]\textsuperscript{1},
Donghao Wang\textsuperscript{1},
Yizhuo Zhang\textsuperscript{1},
Jiacheng Yang\textsuperscript{1},\\
Qi Fan\textsuperscript{1},
Yuekun Yang\textsuperscript{1},
Wenbin Li\thanks{Corresponding author.}\textsuperscript{1},
Feng Miao\textsuperscript{2},
Yang Gao\textsuperscript{1}\\
\textsuperscript{1}School of Artificial Intelligence Science and Technology, Nanjing University\\
\textsuperscript{2}School of Physics, Nanjing University\\
{\tt\small yunkaidang@smail.nju.edu.cn}, 
{\tt\small liwenbin@nju.edu.cn}
}

\begin{document}
\maketitle

\begin{abstract}
Multimodal large language models (MLLMs) show strong perception and reasoning abilities on existing remote sensing (RS) benchmarks. 
However, these benchmarks mostly rely on low-resolution RS images, and few high-resolution benchmarks have flawed reasoning-task designs. 
We show that text-only LLMs can perform competitively with multimodal VLMs on RS reasoning tasks without seeing the images. 
This reveals a serious mismatch between current benchmarks and the intended evaluation of visual understanding. 
To enable faithful assessment, we introduce RSHR-Bench, a super-high-resolution benchmark for visual understanding and reasoning in RS. RSHR-Bench comprises 5,329 full-scene images whose long side is at least $4000$ pixels. 
Each image contains up to $\sim 3\times 10^{8}$ pixels and is drawn from widely used RS corpora and UAV collections. 
We design four task families: multiple-choice VQA, open-ended VQA, image captioning, and single-image evaluation. 
These tasks cover nine perception categories and four reasoning types, supporting both multi-turn and multi-image dialog. 
To reduce reliance on language priors, we use adversarial filtering with strong LLMs followed by rigorous human verification. 
Overall, we construct 3,864 VQA tasks (including closed-set and open-ended settings), 3,913 image captioning tasks, and 500 fully human-written/verified single-image evaluation VQA pairs. 
Evaluating a broad suite of open- and closed-source, as well as RS-specific, VLMs on RSHR-Bench reveals persistent performance gaps in high-resolution scenarios.
Our code is publicly available at \url{https://github.com/Yunkaidang/RSHR}.

 
\end{abstract}

\vspace{-2em}

\section{Introduction}
\label{sec:intro}

\begin{figure}[t]
\centering
\includegraphics[width=0.95\linewidth]{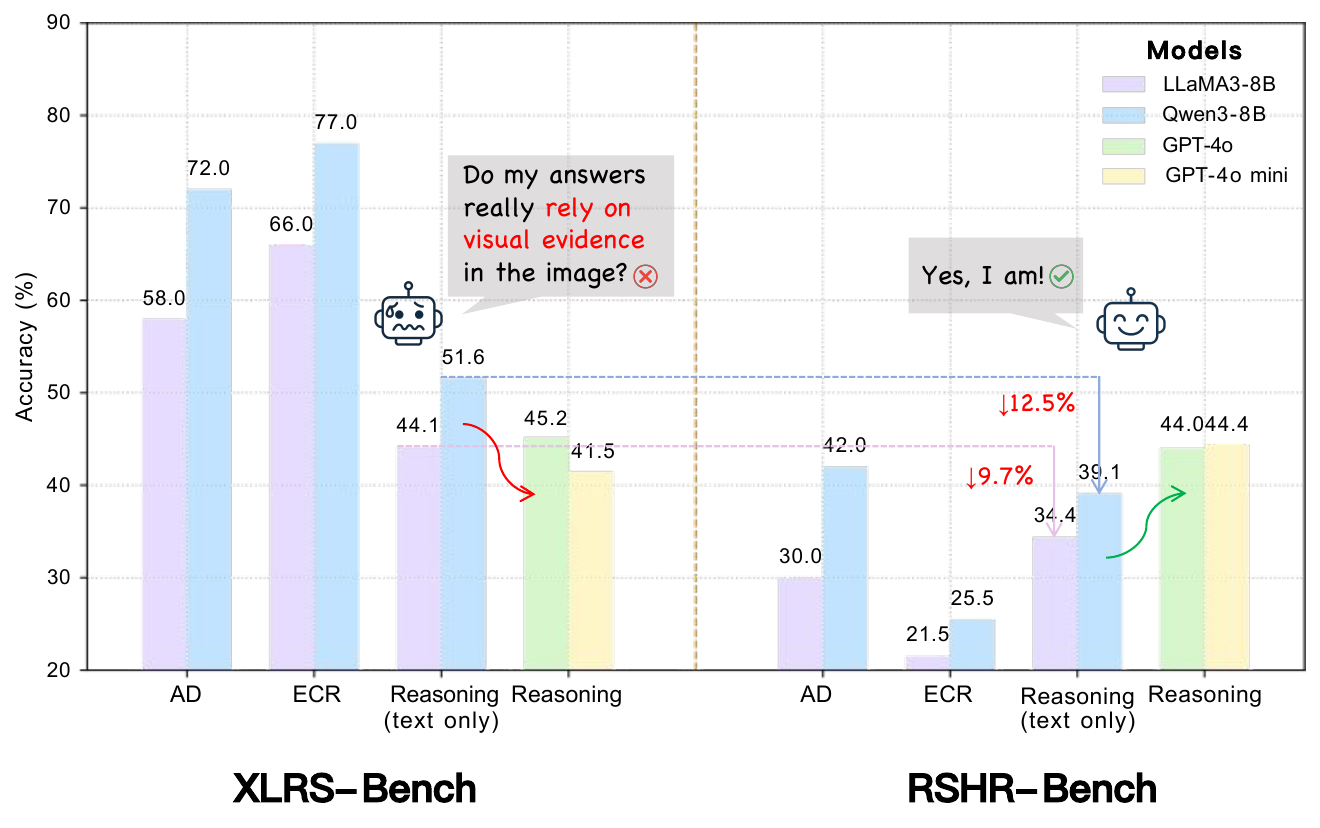}
\caption{Accuracy on XLRS-Bench~\citep{Wang2025XLRSBench} and RSHR-Bench. 
Tasks: AD (Anomaly Detection) and ECR (Existence \& Counting Reasoning). 
We report the average reasoning accuracy under two input settings: \emph{text-only} (Llama3-8B and Qwen3-8B) and \emph{multimodal} (image+text; GPT-4o and GPT-4o mini). 
RSHR-Bench exhibits a larger gap between text-only and multimodal settings, indicating stronger reliance on visual information.}
\label{fig:low_resolution_XLRSBench_llm}
\end{figure}

In recent years, multimodal large language models (MLLMs) have markedly advanced visual understanding and reasoning~\citep{chen2023internvl,wang2025internvl3,yao2024minicpm,abdin2024phi3,Qwen2.5-VL,anthropic2024claude,openai2025gpt5thinking,hurst2024gpt} ability. Driven by applications such as high-resolution video and autonomous driving~\citep{liu2024survey}, a parallel line of work scales MLLMs~\citep{llava-uhd,li2024mini,llava-next,zhang2024beyond,shi2025scaling} to handle high-resolution inputs, and several benchmarks~\citep{mmerealworld2025,wang2025traceable,wu2024v} admit 4K/8K images to evaluate these models. 
In contrast, remote sensing often involves ultra-high-resolution images captured by satellites and UAVs across large geographic areas. These scenes exhibit extreme multi-scale variation and contain small, sparsely distributed objects within cluttered backgrounds~\citep{ball2017comprehensive}.
However, despite progress on high-resolution inputs, general-purpose models remain insufficiently evaluated on operational remote sensing imagery, particularly at native spatial resolutions and geographic scales. 
Consequently, domain-specific MLLMs for remote sensing~\citep{kuckreja2024geochat,soni2025earthdial,wang2025geollava8k,pang2025vhm} are typically fine-tuned on remote-sensing data and employ either patch-based processing or direct resizing to address diverse tasks (perception, reasoning, captioning).

A series of remote sensing benchmarks~\citep{wang2025rseval,muhtar2024lhrs,Wang2025XLRSBench,li2024vrsbench,danish2025geobench} support evaluation across diverse remote-sensing scenarios. However, these studies still emphasize low-resolution settings, relying on small tiles that obscure scene-level context: VRSBench~\citep{li2024vrsbench} primarily uses \(512{\times}512\) slices; RSVQA~\citep{lobry2020rsvqa} adopts \(499{\times}499\); HRVQA~\citep{li2024hrvqa} reaches \(1024{\times}1024\). Recently, higher-resolution multimodal benchmarks have emerged, scaling to \(8500{\times}8500\) (XLRS-Bench~\citep{Wang2025XLRSBench}) and \(7099{\times}6329\) (LRS-VQA~\citep{luo2025lrsvqa}). However, limitations remain in reasoning-task design: XLRS-Bench yields high accuracy even \emph{without} visual input, and LRS-VQA focuses mainly on perception rather than reasoning.
As shown in Fig.~\ref{fig:low_resolution_XLRSBench_llm}, text-only LLMs achieve \(77\%\) accuracy on existence and counting reasoning tasks on XLRS-Bench. 
Moreover, across all reasoning tasks, a text-only LLM (Qwen3-8B) attains \(51.6\%\) accuracy, surpassing the GPT-4o at \(45.2\%\) (multimodal setting).
These results suggest that models may answer many questions by exploiting textual cues and prior world knowledge, rather than faithfully interpreting the visual content.

Therefore, evaluating the high-resolution remote sensing requires careful redesign. 
We highlight four key challenges:
\textbf{(1) Benchmark resolution.} 
Most existing benchmarks assess models on small images, whereas real-world RS scenes are much larger—for example, a single DOTA~\citep{xia2018dota} image can contain up to $\sim2\times10^{8}$ pixels, and PANDA~\citep{Bulten2022PANDA} reaches gigapixel-level resolution (i.e., $\sim2\times2.5^{8}$ pixels) per frame.
\textbf{(2) Long-range structure under high resolution.} 
Common tiling pipelines use inputs ranging from 512 to 2,000 pixels, which fragment the global layout and object constellations. As a result, they rarely provide a direct evaluation of a model’s understanding of the entire image.
\textbf{(3) Reasoning-task design that fails to control for LLM priors.} 
Many high-resolution benchmarks~\citep{Wang2025XLRSBench,luo2025lrsvqa,mmerealworld2025} rely on multiple-choice formats, allowing models to exploit text-only priors and inflate scores.
\textbf{(4) Insufficient human verification.} Large portions of question–answer pairs are generated automatically without rigorous human verification, leading to unrealistic items (e.g., “How many cars are in the image?” with an answer “300”).

To address these gaps, we present the \emph{\underline{R}emote-\underline{S}ensing \underline{H}igh-\underline{R}esolution Benchmark (\textbf{RSHR-Bench})}—a super-high-resolution remote sensing benchmark for understanding and complex reasoning tasks. RSHR-Bench curates a corpus of 5{,}329 full-scene remote-sensing images while preserving native resolutions: each image has a long side $\geq$4K, with pixel counts up to $\sim3\times10^{8}$ ($\approx$300 MP).
Source images span widely-used ultra–large-scene datasets—DOTA~v1.0/v2.0~\citep{xia2018dota,dota2022}, XLRS-Bench~\citep{Wang2025XLRSBench}, MiniFrance~\cite{CastilloNavarro2022MiniFrance}, FAIR1M~\cite{Sun2022FAIR1M}, HRSCD~\citep{Daudt2019HRSCD}—as well as our UAV-captured imagery (up to $10^{8}$ pixels).
We then design 13 prompt templates and use Qwen2.5-VL-7B~\cite{Qwen2.5-VL} and GPT-5~Thinking~\citep{openai2025gpt5thinking} to generate four types of tasks: multiple-choice VQA, open-ended VQA (free-form), image captioning, and single-image evaluation.
For the VQA task suite, we comprise nine perception categories and four reasoning types, spanning both single- and multi-image settings.
Different from prior RS benchmarks that emphasize single-turn choice, \emph{RSHR-Bench} additionally supports multi-turn dialog and multi-image dialog, better reflecting realistic remote-sensing analysis workflows.

To verify the quality, we built a semi-supervised human-LLM verification pipeline that ensures the question and answer are correct, vision-based, and have no ambiguity or hints.
In the first stage, we use LLM-only adversarial filtering to remove items that can be solved without viewing the image.
And then we add a full human pass, iteratively revising and rechecking until language priors no longer suffice.
To assess overall comprehension at native resolution, we include an image-captioning task on images spanning 4K to $10^{8}$ pixels, and provide ten VQA items per image to evaluate visual question answering and captioning tasks jointly.
We then evaluated a broad set of models in RSHR-Bench, including remote sensing VLMs, general-purpose VLMs (open- and closed-source), and text-only LLMs.
The evaluation result shows that all fourteen models exhibit poor performance across four types of tasks.
Overall, our contributions can be summarized as follows:
\begin{itemize}[leftmargin=*]
\item We introduce \emph{RSHR-Bench}, a new remote-sensing benchmark designed to fairly and comprehensively evaluate VLMs on ultra-high-resolution imagery.
\item We design four representative task types—multiple-choice and open-ended VQA, image captioning, and single-image evaluation—to assess the performance of general-purpose VLMs and RSVLMs.
\item We evaluate models on RSHR-Bench and find that all perform poorly, highlighting the need for progress toward real-world remote-sensing applications.
\end{itemize}


\section{Related Work}
\label{sec:related}

\textbf{General Multimodal Benchmarks.}
Recent multimodal benchmarks have advanced the quantitative assessment of VLMs, yet many focus on narrow domains or a small set of tasks (e.g., captioning~\citep{coco_captions,flickr30k_entities} or VQA~\citep{gqa,okvqa,vqa2,vizwiz,textvqa,youcook2}). 
To broaden coverage, MME~\citep{mme} spans 14 perceptual and cognitive tasks, MMBench~\citep{mmbench} offers 3{,}000+ questions over 20 skill dimensions (e.g., localization and social reasoning), Seed-Bench~\citep{seedbench} further enlarges question volume (19k), and MMT-Bench~\citep{ying2024mmt} brings in application-oriented data from autonomous driving and embedded AI. 
Complementary to these, several \emph{high-resolution natural-scene} benchmarks stress-test long-context vision at large input sizes but remain outside the remote-sensing domain: 
HRBench~\citep{wang2025divide} targets 4K/8K-scale inputs, TreeBench~\citep{wang2025traceable} evaluates traceable visual reasoning at $\sim$2K-level inputs, and MME-RealWorld~\citep{mmerealworld2025} reports typical images around $2{,}000{\times}1{,}500$ with very high maximum. However, these general-purpose datasets contain little to no RS imagery, providing limited RS-specific semantics and annotations. And their average image sizes are still below real RS scenes (e.g., HRSCD~\citep{hrscd} uses images up to 100\,MP).

\textbf{Remote Sensing Multimodal Benchmarks.} Existing RS benchmarks for multimodal models fall into three routes. \emph{(i) Captioning/VQA at modest resolution:} early efforts such as RSIEval~\citep{wang2025rseval} and LHRS-Bench~\citep{muhtar2024lhrs} target image captioning and VQA with hundreds of human-curated items, focusing primarily on perception tasks with limited visual context. \emph{(ii) Reliability and fine-grained semantics:} RSSA~\citep{h2rsvlm2024} focuses on false information detection, while FIT\mbox{-}RSRC~\citep{skysensegpt2024} and VLEO\mbox{-}BENCH~\citep{vleo-bench2024} emphasize object relations and scene-level understanding (e.g., urban monitoring, disaster relief, counting, localization, change detection), expanding beyond generic Q\&A to focus on trustworthiness and structured semantics. \emph{(iii) Large-scale and high-resolution:} to match real-world RS—which routinely exceeds 4{,}000{\small$\times$}4{,}000 in detection (DOTA~\citep{xia2018dota}) and up to 10{,}000{\small$\times$}10{,}000 in segmentation (HRSCD~\citep{hrscd})—recent benchmarks push resolution and breadth: XLRS\mbox{-}Bench\citep{Wang2025XLRSBench} averages $\sim$8{,}500{\small$\times$}8{,}500 with 16 sub-tasks spanning VQA, captioning, and localization; VRSBench~\citep{li2024vrsbench} provides 29{,}614 images with rich QA and object references for versatile evaluation. MDAS~\citep{hu2023mdas} introduces high spectral resolution for classification and change detection. Overall, the field is progressing from low-resolution captioning and VQA toward ultra-high-resolution benchmarks that better capture the needs of real-world deployment.

\textbf{Multimodal Large Language Models.}
Multimodal large language models (MLLMs) have advanced rapidly in perception and reasoning \citep{learning,DecodingTrust}, yet most general-purpose systems (e.g., LLaMA \citep{llama}, Gemini \citep{gemini}, GPT-4o \citep{hurst2024gpt}, Qwen-VL \citep{qwen}, InternLM-XComposer \citep{InternLM-XComposer}, MiniCPM \citep{yao2024minicpm}, LLaVA \citep{visualllama}, MiniGPT-4 \citep{zhu2023minigpt}) are not specifically optimized for ultra-high-resolution inputs and typically support only 2K–4K images. To overcome this limitation, recent work has scaled MLLMs to high resolution using three main strategies: patching with global alignment (LLaVA-Next \citep{llava-next}), token/patch compression (Monkey \citep{monkey}; LLaVA-UHD \citep{llava-uhd}) and dual-encoder or multi-scale fusion with learnable queries (Mini-Gemini \citep{li2024mini}; Cambrian \citep{cambrian2024}; SliME \citep{sliME2024}). These strategies reduce sequence length while preserving global semantics. In the RS field, research mainly focuses on MLLMs for geospatial understanding~\citep{sliME2024,efficient2024,huang2022fine,roma2025,Bazi2024RSLLaVA,Zhan2025SkyEyeGPT,skysensegpt2024,yao2025falcon,zhang2025georsmllm}.
GeoChat~\citep{kuckreja2024geochat} adapts instruction tuning for multi-turn dialogue on RS imagery;
LHRS-Bot~\citep{muhtar2024lhrs} introduces multi-level vision--language alignment with curriculum learning and VGI for stronger grounding;
EarthGPT unifies multi-sensor (optical/SAR/multi-spectral) interpretation under one generative interface~\citep{zhang2024earthgpt}.
On reliability and scale, VHM~\citep{pang2025vhm} reduces hallucination and promotes calibrated uncertainty, while GeoLLaVA-8K~\citep{wang2025geollava8k} enables ultra-high-resolution (8K) reasoning via tiling and context aggregation.
Persistent gaps include faithful geospatial grounding, multi-temporal reasoning, large-scale sensor fusion, and adjusted uncertainty.

\section{Method} \label{sec:method}
\begin{figure*}[t]
\centering
\includegraphics[width=\linewidth]{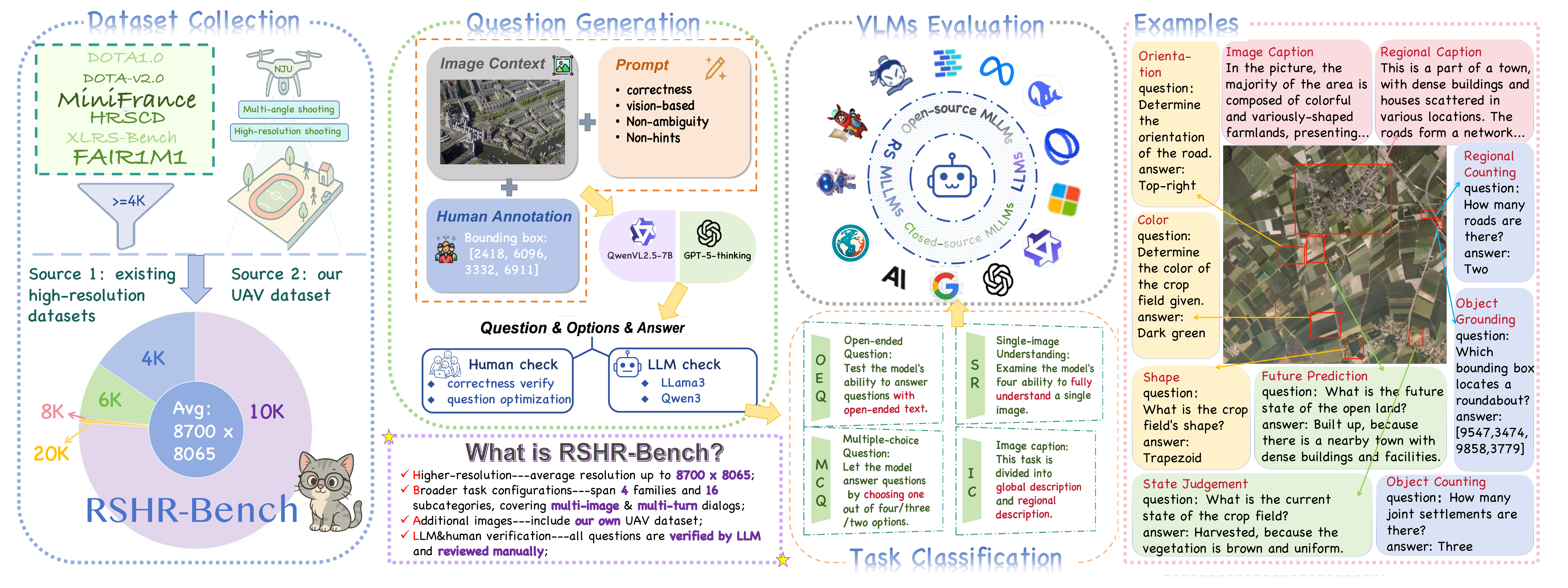}
\caption{
This overview shows the construction of our RSHR-Bench: We collect high-resolution imagery from multiple datasets and supplement it with images from our own UAV dataset. Then we generate questions, followed by LLM and human verification. The resulting tasks are categorized into four main types, covering various VLMs evaluation experiments. Finally, on the right, examples of single-image understanding tasks illustrate how the benchmark is applied.
}
\label{fig:low_resolution_XLRSBench}
\vspace{-3mm}
\end{figure*}

The overall pipeline (Fig.~\ref{fig:low_resolution_XLRSBench}) consists of three stages: dataset collection, question generation, and human and LLM verification. We also discuss the difficulties and compare with prior benchmarks in section~\ref{RSHR-Bench}. In section~\ref{Evaluation Dimensions}, we present the evaluation dimensions for our four main tasks.

\subsection{RSHR-Bench} \label{RSHR-Bench}
\textbf{Dataset Collection.}  
We assemble a high-resolution corpus by combining six public remote sensing benchmarks with a high-altitude UAV set of \(\sim100\,\mathrm{MP}\) single-shot frames. For every source, we keep native resolution and filter images so that the long side is at least 4K pixels. From DOTA v1.0 \cite{xia2018dota} and DOTA v2.0 \cite{dota2022}, we select large-format satellite scenes that contain dense human-made structures and long-range context. From XLRS-Bench \cite{Wang2025XLRSBench}, we sample ultra-large panoramas that reach tens of thousands of pixels on a side. From HRSCD \cite{Daudt2019HRSCD}, we retain paired views of the same areas to support temporal consistency analysis. From MiniFrance \cite{CastilloNavarro2022MiniFrance}, we choose very high-resolution metropolitan tiles with rich textures and many small objects. From FAIR1M 1.0 \cite{Sun2022FAIR1M}, we gather wide-area airport, port, and industrial scenes with dense fine-scale targets and varied viewing angles. Our UAV collection repeatedly surveys the same geographic area from diverse viewpoints and flight attitudes, with each frame captured in a single exposure at about \(100\,\mathrm{MP}\), which preserves native geometry. 
Compared with typical ultra-high-resolution benchmarks where targets occupy large pixel footprints~\citep{Bulten2022PANDA}, our UAV scenes are dominated by small objects under strong scale and perspective variation. 
Overall, our image collection is constructed based on the principles of high resolution, comprehensive scene coverage, detailed object diversity, and task-orientation.
Summary statistics and resolution distributions, including the UAV data, are shown in Fig.~\ref{fig:low_resolution_XLRSBench}. We provide detailed, per-dataset descriptions of data-collection procedures and quantities at multiple resolutions in the appendix.

\textbf{Question Generation.} 
Following the recent remote sense benchmark~\citep{li2024vrsbench, Wang2025XLRSBench}, we instantiate 13 prompt templates that cover perception and reasoning across four task types, including \emph{closed-set VQA} (multiple choice), \emph{open-ended VQA} (free-form), \emph{image captioning}, and \emph{single-image understanding}. Perception tasks cover nine common query families (e.g., color, orientation/position, region grounding, regional counting). 
And reasoning questions are image-specific, require explicit visual evidence, and avoid phrasing that hints at the correct option (e.g., future prediction, object state judgement). 
All the tasks in the design follow three principles—\emph{correctness}, \emph{unambiguous phrasing}, and \emph{visual answerability}. 
For closed-set and open-ended VQA tasks, we prioritize small but distinctive targets. This approach emphasizes high-resolution understanding while ensuring that each target remains uniquely identifiable in context during annotation. And then we implement these questions and choices through a two-stage drafting pipeline: perception-oriented items with given boxes and labels are drafted with Qwen2.5-VL-7B~\cite{Qwen2.5-VL}, whereas harder compositional or multi-image items are drafted with GPT-5 Thinking~\citep{openai2025gpt5thinking}. 
For full-image captioning, we partition each image into four directional sectors (top, bottom, left, right) and use GPT-5 Thinking~\citep{openai2025gpt5thinking} to generate image captions that separately describe each sector. 
For single-image tasks, we design ten questions per image covering both perception and reasoning, including eight open-ended questions, one local image caption, and one global image caption.
We finally generated 50 images and 500 questions to evaluate the model’s single-image understanding. 
All four-task questions and answers are fully authored by human annotators, while the image captions are produced by GPT-4o and verified for correctness by humans.
All prompts, templates, and examples are provided in the Appendix.


\textbf{Human and LLM Verification.} 
To ensure question quality, we use a two-stage human–LLM validation pipeline. In the first stage, we separate question writing from verification to enhance reliability. Annotator \textbf{A} drafts questions (and, when applicable, reviews model-generated questions\&answers) and ensures four criteria: correctness, strict visual grounding, unambiguous phrasing, and no unintentional hints. Reviewer \textbf{B} then independently audits A’s outputs and any model drafts, flags inconsistencies, and discards invalid items. A team of six trained annotators (each with a bachelor’s degree) completed this stage in about 300 hours. We also maintain complete provenance records (including image IDs, prompts, model outputs, and edit history) to support traceability and resolve ambiguities. 
In the second stage, we use text-only LLMs (Qwen3-8B~\cite{yang2025qwen3} and Llama3-8B~\citep{llama3modelcard}) to detect and eliminate the solvability of language priors alone. Specifically, the LLMs attempt to answer each question without access to the corresponding image. 
If a model attains higher accuracy by relying solely on linguistic cues or statistical regularities than by depending on images in the answer options, we remove or revise the affected items to mitigate this artifact. We iterate rewriting until correct solutions require genuine visual grounding and all options are free of suggestive cues.
After multiple rounds of editing and review, we obtained a high-quality dataset of 1,932 image–question pairs. 
The full generation and validation process consumed about 100 GPU hours.

\textbf{Difficulties in the Pipeline.}
We identify four main challenges in constructing reliable supervision and evaluation for our multitasking setting.
\ding{182}\ \textbf{Annotation Diversity.}
Many images contain clusters of highly similar targets, reducing instance-level diversity and encouraging overfitting.
We manually screen the corpus to prioritize a broader set of commonly observed remote-sensing scenes and objects( e.g.,land-use types, multiple transportation modes). Our goal is to increase both inter- and intra-class variability. At the same time, we aim to preserve the native scene context and maintain source distributions across splits.
\ding{183}\ \textbf{Question Hinting.}
Descriptive cues can leak labels (e.g., \textit{forest} $\rightarrow$ \textit{green}), enabling elimination by common sense rather than visual evidence. We first use text-only LLMs to assess language dependence in the prompts. If the question relies too much on language, we apply human post-editing to remove vocabulary shortcuts and standardize wording, while retaining the intended visual evidence and difficulty.\ding{184}\ \textbf{Distractor Design.}
Due to the limitations of static imagery and resolution, generating distractors for inference tasks becomes particularly challenging, as the interpretation of answers is often not unique. For example, a yellow patch of land could be either natural or man-made. However, if we observe signs of crop lodging, we can infer a natural disaster. To address this, we prioritize selecting objects with clear causal features, ensuring that answers are unique and well-defined. Additionally, we incorporate real-world, distinctive characteristics of objects into the answer design to reduce ambiguity and improve reliability.
\ding{185}\ \textbf{Answer Errors.}
Models exhibit systematic errors on small boxed targets and in counting: even when the correct answer is “0” and “0” appears among the options, predictions skew positive; for ultra-high-resolution images (\(>10^8\) pixels), detection recall drops sharply.
We mitigate these effects through targeted human review and corrections to improve label fidelity and evaluation fairness.

\textbf{Comparison with Prior Benchmarks.}
As shown in Table~\ref{tab:rs_vqa_overview_en_our}, we compare our benchmark with existing remote sensing datasets.
Specifically, we consider two categories of remote sensing datasets: non-VQA and VQA.
Non-VQA datasets (e.g., DOTA~\cite{xia2018dota,dota2022}, DIOR~\cite{li2021dior}) focus on recognition or captioning and generally lack multi-turn dialogue and multi-image reasoning.
Existing VQA datasets are primarily low-resolution (typically $512\times512$) and are limited to single-turn, single-image interactions (e.g., RSVQA~\cite{lobry2020rsvqa}, HRVQA~\cite{li2024hrvqa}, RSIVQA~\cite{zheng2021mutual}).
In contrast, our benchmark emphasizes:
\ding{182}\ \textbf{Higher-resolution imagery.} We curate ultra-high-resolution imagery up to $10^9$ pixels, pushing evaluation toward real-world remote sensing.
Our benchmark couples ultra-high-resolution images with interaction-rich evaluation and explicit reliability checks: images spanning \emph{tens to hundreds of megapixels} (avg.\ $8{,}700\times8{,}065$; max $29{,}200\times27{,}620$) from satellite, aerial, and UAV sources.
\ding{183}\ \textbf{LLM verification.} Perception and reasoning items are verified by LLMs under deliberately reduced-resolution views to stress visual competence rather than language priors.
We employ a human–LLM validation pipeline with LLM-based verification and a curated, challenging VQA set (1{,}578 items over 1{,}366 images) that targets perception-to-reasoning chains previously only partially covered.
\ding{184}\ \textbf{Global image perception and broader task configurations.} 
We vary interaction formats (e.g., single-image multi-turn and multi-image single-round) to probe multi-image fusion, memory, and decision-making, and broaden task coverage from fine-grained perception (color/shape cues) to cross-image reasoning (multi-region contrast, object-state judgment, future prediction).

\begin{table}[h]
\footnotesize
\centering
\caption{Comparisons between existing remote sensing benchmarks and our benchmark. 
\xmark and \cmark denote not included and included, 
\icoyes, \icono, and \icohalf denote machine-generated, human-written, and semi-automatic (\textit{i.e.}, machine generation followed by human verification), respectively. 
Top: Non-VQA; Bottom: VQA. Dashes (\textendash) indicate missing or not reported.}
\resizebox{\linewidth}{!}{%
\begin{tabular}{l|c|r|c|c|c|c|c}
\toprule
\rowcolor{gray!15}
\textbf{Dataset} & \textbf{Source} &
\textbf{Images} &
\textbf{Avg Res.} &
\textbf{Max Res.} &
\textbf{Multi-turn} &
\textbf{Annotation Method} &
\textbf{LLM Check} \\
\midrule
\rowcolor{gray!8}
\multicolumn{8}{l}{\textbf{Non-VQA}} \\
\midrule
GeoPixel~\cite{shabbir2025geopixel}           & Aerial             & 65{,}463  & 2{,}560$\times$2{,}560 & 4{,}960$\times$4{,}960  & \xmark & \icohalf & \xmark \\
refGeo~\cite{refgeo2023}                      & Aerial, UAV        & 80{,}000  & 640$\times$640         & 1{,}024$\times$1{,}024  & \xmark & \icohalf & \xmark \\
LuoJiaHOG~\cite{zhao2025luojiahog}            & TBD                & 94{,}856  & 640$\times$640         & 1{,}024$\times$1{,}024  & \xmark & \icohalf & \xmark \\
DOTA v1.0~\cite{xia2018dota}                       & Aerial, Satellite  & 2{,}806   & 4{,}000$\times$4{,}000 & 20{,}000$\times$20{,}000 & \xmark & \icono   & \xmark \\
DOTA v2.0~\cite{dota2022}                       & Aerial, Satellite  & 11{,}268  & 4{,}500$\times$4{,}500 & 20{,}000$\times$20{,}000 & \xmark & \icohalf & \xmark \\
FAIR1M 1.0~\cite{ Sun2022FAIR1M}                     & Satellite          & 15{,}000  & 5{,}300$\times$5{,}300 & 10{,}000$\times$10{,}000 & \xmark & \icohalf & \xmark \\
RSICD~\cite{lu2017rsicd}                      & Aerial, Satellite  & 10{,}921  & 224$\times$224         & 224$\times$224          & \xmark & \icono   & \xmark \\
RSVG~\cite{zhan2023rsvg}                      & Satellite          & 17{,}402  & 800$\times$800         & 800$\times$800          & \xmark & \icono   & \xmark \\
RRSIS-D~\cite{yuan2024rrsis}                  & Satellite          & 17{,}402  & 800$\times$800         & 800$\times$800          & \xmark & \icohalf & \xmark \\
RSIEval~\cite{wang2025rseval}                 & Aerial, Satellite  & 100       & 512$\times$512         & 512$\times$512          & \xmark & \icono   & \xmark \\
DIOR~\cite{li2021dior}                        & Satellite          & 23{,}463  & 1{,}200$\times$1{,}200 & 2{,}000$\times$2{,}000  & \xmark & \icono   & \xmark \\
MillionAID~\cite{li2022millionaid}           & Aerial, Satellite  & 1{,}000{,}848 & —                    & 31k$\times$31k          & \xmark & \icono   & \xmark \\
RSICap~\cite{lu2023rsicap}                    & Aerial, Satellite  & 2{,}585   & 224$\times$224         & 224$\times$224          & \xmark & \icono   & \xmark \\
\midrule
\rowcolor{gray!8}
\multicolumn{8}{l}{\textbf{VQA}} \\
\midrule
XLRS-Bench~\cite{Wang2025XLRSBench}                & Satellite, Aerial  & 3{,}079   & 8{,}500$\times$8{,}500 & 10{,}000$\times$10{,}000 & \xmark & \icohalf & \xmark \\
LRS-VQA~\cite{luo2025lrsvqa}                  & Satellite, Aerial  & 1{,}657   & 5{,}403$\times$4{,}935 & 26{,}176$\times$24{,}832 & \xmark & \icohalf & \xmark \\
RSIVQA~\cite{zheng2021mutual}                  & Aerial, Satellite  & 37{,}264  & 512$\times$512         & 512$\times$512          & \xmark & \icohalf & \xmark \\
VRSBench~\cite{li2024vrsbench}                & Aerial, Satellite  & 29{,}614  & 512$\times$512         & 512$\times$512          & \cmark & \icohalf & \xmark \\
RSVQA~\cite{lobry2020rsvqa}                   & Aerial, Satellite  & 11{,}431  & 499$\times$499         & 512$\times$512          & \xmark & \icoyes  & \xmark \\
HRVQA~\cite{li2024hrvqa}                   & Aerial             & 53{,}512  & 1{,}024$\times$1{,}024 & 1{,}024$\times$1{,}024  & \xmark & \icoyes  & \xmark \\
TAMMI~\cite{zhang2025tammi}                   & Aerial, Satellite  & 282{,}852 & —                      & —                       & \xmark & \icoyes  & \xmark \\
RSVLM-QA~\cite{zi2025rsvlm} & Aerial, Satellite  & 6{,}000   & 512$\times$512         & 512$\times$512          & \xmark & \icohalf & \xmark \\
EarthVQA~\cite{wang2024earthvqa}              & Satellite          & 6{,}000   & 512$\times$512         & 512$\times$512          & \xmark & \icohalf & \xmark \\
Segmentation VQA~\cite{tosato2024segguided}   & Aerial             & 16{,}274  & 2{,}560$\times$2{,}560 & 4{,}096$\times$4{,}096  & \xmark & \icoyes  & \xmark \\
\midrule
\textbf{RSHR-Bench} & \textbf{Satellite, Aerial, UAV} & 5{,}329 & 8{,}700$\times$8{,}065 & 29{,}200$\times$27{,}620 & \cmark & \icohalf & \cmark \\
\bottomrule
\end{tabular}%
}
\label{tab:rs_vqa_overview_en_our}
\vspace{-2mm}
\end{table}

\subsection{Evaluation Dimensions} \label{Evaluation Dimensions}

\begin{figure}[t]
  \centering
  \begin{subfigure}[b]{0.485\linewidth}
    \centering
    \includegraphics[width=\linewidth]{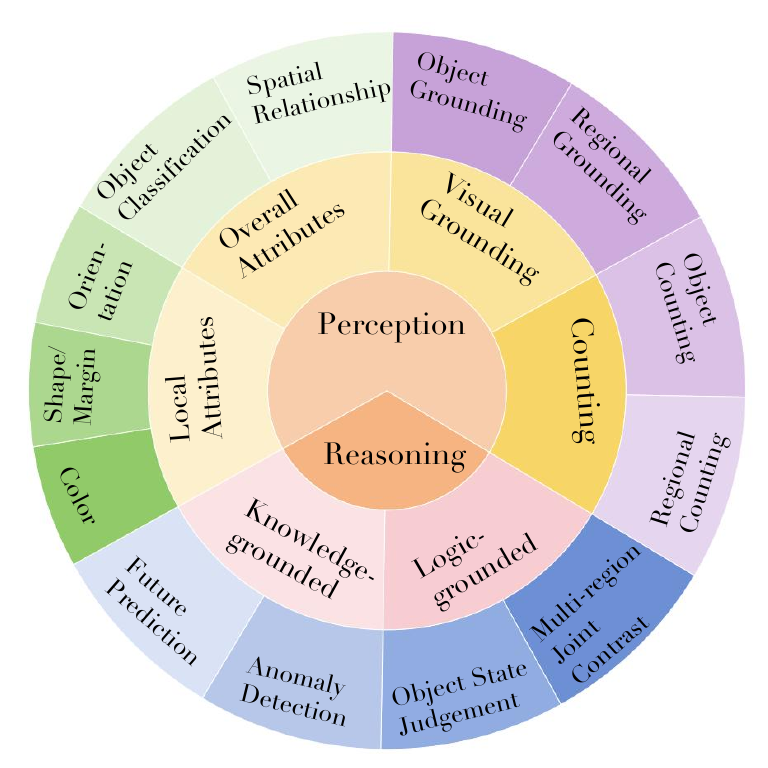}
  \end{subfigure}
  \hspace{0.1mm}
  \begin{subfigure}[b]{0.485\linewidth}
    \centering
    \includegraphics[width=\linewidth]{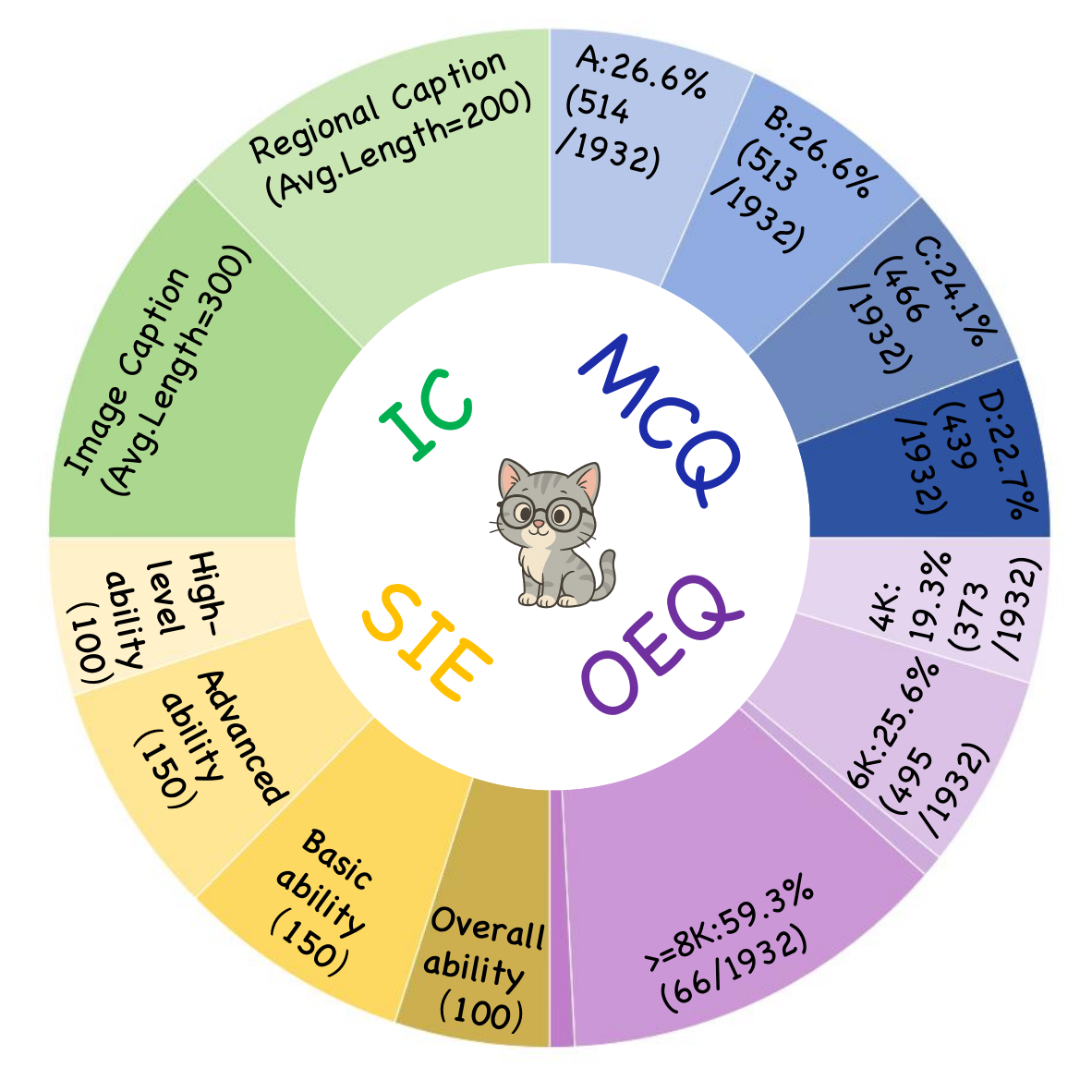}
  \end{subfigure}
    \caption{Overview of our benchmark composition. Left: task categories for perception and reasoning. Right: counts of tasks within the four capability groups—multiple-choice VQA (MCQ), open-ended questions (OEQ), image captioning (IC), and single-image evaluation (SIE).}
  \label{fig:two-in-one-column}
  \vspace{-0.5cm}
\end{figure}

\textbf{Overview.}
We evaluate models along two complementary dimensions—\emph{Perception} and \emph{Reasoning}(Fig.~\ref{fig:two-in-one-column}  (left))—via four tasks(Fig.~\ref{fig:two-in-one-column} (right)) that assess recognition, localization, grounding, and holistic understanding in ultra-high-resolution remote sensing imagery.
(i) \emph{Multiple-choice VQA}: evaluates decision-making within a fixed answer space.  
(ii) \emph{Open-ended VQA}: without multiple-choice options, this task assesses free-form visual understanding and compositionality, offering a more accurate measure of a model's capabilities.  
(iii) \emph{Image Captioning}: requires concise, accurate descriptions for regions and whole images.  
(iv) \emph{Single-Image Evaluation}: assesses single-image understanding, covering both perception and reasoning across resolutions, as well as image captioning.

\textbf{Perception.} Our benchmark decomposes perception into a suite of localized and scene-level tasks that stress-test recognition, localization, and spatial reasoning in high-resolution remote-sensing imagery. Concretely, we evaluate (i) \emph{Color Detection}, requiring identification and classification of colors confined to a provided bounding box; (ii) \emph{Shape/Margin Recognition}, which probes the model’s ability to delineate precise object boundaries; (iii) \emph{Orientation Detection}, assessing estimation of object heading or extension direction from cropped regions; (iv) \emph{Object Classification}, spanning land-use categories alongside fine-grained targets common in aerial settings (e.g., terminals, berthing areas, and ship types); (v) \emph{Object Spatial Relationship}, which elicits structured descriptions of relative positions among multiple entities; (vi) \emph{Object Grounding}, measuring localization accuracy from precise natural-language referents; (vii) \emph{Regional Grounding}, focusing attention on compositional subregions (e.g., the area where three roads intersect with a nearby fishpond); and (viii) \emph{Counting}, both at the global image level and regionally within a given box. 

\textbf{Reasoning.}  Complementing perception, our reasoning track targets temporal, causal, and contextual inference. We include (i) \emph{Anomaly Detection \& Interpretation}, in which models must not only localize abnormal or distinct regions but also infer plausible causes (e.g., clearance, flooding, construction, deforestation) under single-image and multi-turn settings; (ii) \emph{Future Prediction}, assessed in two configurations: Single-Image Multi-Round (50 groups, 3 questions each) and Multi-Image Single-Round (50 pairs of the same location across years), requiring extrapolating credible evolutions from current evidence; (iii) \emph{Multi-Region Joint Contrast}, which compares and synthesizes information across multiple regions within one or several images; and (iv) \emph{Object State Judgement}, determining whether objects are static or dynamic given contextual cues. By unifying these tasks under consistent inputs and prompts, the benchmark holistically measures a model’s capacity to connect “what is where” with “what it means and what comes next,” establishing a high bar for vision–language systems aspiring to real-world remote-sensing intelligence.

\textbf{VQA Task Suite.} We evaluate vision–language capabilities across nine \emph{Perception} and four \emph{Reasoning} tasks, defined in single- and multi-image settings with single- and multi-turn interactions. Following recent benchmarks~\citep{luo2025lrsvqa, Wang2025XLRSBench}, we adopt multiple-choice VQA (A/B/C/D) to evaluate all VLMs.
To mitigate language priors and random guessing inherent to multiple-choice formats, we further apply a standardized protocol for open-ended conversion: for each item, GPT-4o~\citep{hurst2024gpt} (i) removes the options; 
(ii) rewrites the question so the answer \emph{must} be inferred from visual evidence rather than textual cues or commonsense priors; 
and (iii) ensures the reformulated question cannot be answered without inspecting the image.
Each rewritten item undergoes human review to verify unambiguous visual grounding and clarity. Overall, we annotated 1,578 objects with bounding boxes across 1,366 images, generating a total of 1,932 question-answer pairs.
Accordingly, the open-ended VQA setting offers no answer cues and is more challenging than its multiple-choice counterpart.

\textbf{Image Captioning task.} For remote sensing datasets~\citep{lobry2020rsvqa,Wang2025XLRSBench,luo2025lrsvqa}, image captioning is a standard evaluation task that probes a model’s scene-understanding capabilities.
We assemble a corpus of ultra-high-resolution imagery (resolution $\geq$ 4K; predominantly $\sim$ 100-megapixel scenes) spanning ports, airports, forests, and urban and rural regions. We first design a prompt template and iteratively validate it on individual images to ensure robust caption generation. Applying \emph{GPT-5~Thinking}, we then generate (i) a global summary capturing the primary scene, salient objects, spatial layout, and functional use; and (ii) structured regional descriptions for the top, bottom, left, and right areas—each beginning with a region-level overview followed by fine-grained details that explicitly elicit attributes (color, position, size, shape) and spatial relations.
The generation guidelines avoid vague quantifiers (e.g., “many,” “some,” “few”).
In total, we produce reference captions for 3{,}913 images and randomly sample 100 for manual verification to assess correctness and reduce hallucinations.
We score with a reference judge, \emph{GPT-5~Thinking}~\citep{openai2025gpt5thinking}, and iteratively refine prompts and instructions to align judge scores with human assessments better.
The evaluation rubric emphasizes (i) visual fidelity and grounding; (ii) coverage of attributes and relations; (iii) coherence and fluency without redundancy; and (iv) factual consistency with low hallucination.
In addition, we use GPT-4o~\citep{hurst2024gpt} to evaluate the score between other models’ captions and those produced by GPT-5 Thinking~\citep{openai2025gpt5thinking}, evaluated separately for global (whole-image) summaries and regional (top/bottom/left/right) descriptions.


\textbf{Single-Image Evaluation.}
We design a single-image evaluation protocol to test whether models truly understand a single remote-sensing image. 
For this setting, we select 50 images with resolutions ranging from $\geq$4K up to $\sim 2\times10^{8}$ pixels and annotate each image with 10 representative subtasks covering four capability groups. 
All question–answer pairs are fully created and carefully checked by human annotators with at least a bachelor's degree.
The subtasks span region- and image-level understanding (image captioning, regional captioning) and perception of basic object attributes (color, shape/boundary, orientation, object classification). 
They further cover reasoning over object relations and challenging localization or counting (spatial relationships, object and regional grounding, object and regional counting), as well as higher-order abilities grounded in visual evidence and remote-sensing knowledge (state judgment, anomaly detection, future prediction).

\section{Experiment}

\begin{table*}[h]
\footnotesize
\vspace{-0.2cm}
\caption{\textbf{Results across visual \emph{Perception}, \emph{Reasoning}, and \emph{Multi-turn} tasks.}
Left-to-right order:
\emph{Perception}—\textbf{COL}=Color Detection, \textbf{SHP}=Shape Recognition, \textbf{DET}=Detection, \textbf{OC}=Object Classification, \textbf{REL}=Object Spatial Relationship, \textbf{OGD}=Object Grounding, \textbf{RG}=Regional Grounding, \textbf{OCN}=Object Counting, \textbf{RCN}=Regional Counting, \textbf{Avg.}=Perception average;
\emph{Reasoning}—\textbf{AD}=Anomaly (single-turn), \textbf{FP}=Future Prediction (multi-image), \textbf{MRJC}=Multi-region Joint Contrast (multi-image), \textbf{MRJCS}=Multi-region Joint Contrast (single-image, multi-box), \textbf{OSJ}=Object State Judgment (single-turn), \textbf{Avg.}=Reasoning average;
\emph{Multi-turn}—\textbf{MAD}=Anomaly, \textbf{MTFP}=Future Prediction, \textbf{MOSJ}=Object State Judgment; 
\emph{Highlighting:} column-wise maxima \emph{(excluding the two LLM rows)} are shown in \textcolor{blue}{blue}; row-wise maxima—computed only over \textbf{Perception} and \textbf{Reasoning} task columns \emph{(excluding both \emph{Avg.} columns, all \emph{Multi-turn}/\emph{MTEM@1}, and the two LLM rows)}—are shown in \textcolor{red}{red}. Ties are highlighted; if both rules apply, \textcolor{blue}{blue} takes precedence. 
https://github.com/Yunkaidang/RSHR
}
\label{tab:visual_reasoning_perception_table2}
\centering
\resizebox{\textwidth}{!}{
\begin{tabular}{l|ccccccccc|c|ccccc|c|ccc|c}
\toprule
\multirow{2}{*}{\textbf{Model}} &
\multicolumn{9}{c}{\textbf{Perception}} &
\multirow{2}{*}{\textbf{Avg.}} &
\multicolumn{5}{c}{\textbf{Reasoning}} &
\multirow{2}{*}{\textbf{Avg.}} &
\multicolumn{3}{c}{\textbf{Multi-turn}} &
\multirow{2}{*}{\shortstack{\textbf{MTEM}\\\textbf{@1}}}\\
\cmidrule(lr){2-10}\cmidrule(lr){12-16}\cmidrule(lr){18-20}
& \textbf{COL} & \textbf{SHP} & \textbf{DET} & \textbf{OC} & \textbf{REL} & \textbf{OGD} & \textbf{RG} & \textbf{OCN} & \textbf{RCN}
&
& \textbf{AD} & \textbf{FP} & \textbf{MRJC} & \textbf{MRJCS} & \textbf{OSJ}
&
& \textbf{MAD} & \textbf{MTFP} & \textbf{MOSJ}
&
\\
\midrule 
\multicolumn{21}{l}{\textit{\textcolor{gray}{Remote Sensing VLMs}}} \\
EarthDial~\cite{soni2025earthdial} & 41.0 & 22.0 & 21.0 & 30.0 & 32.5 & 30.5 & 27.1 & 18.0 & 31.0 & 28.1 & 42.0 & 30.0 & 29.5 & 32.0 & \textcolor{red}{52.0} & 37.1 & 56.7 & 60.0 & 73.5 & 30.7 \\
GeoChat~\cite{kuckreja2024geochat} & 32.5 & 22.0 & 24.0 & 29.5 & \textcolor{red}{40.0} & 25.0 & 22.9 & 22.5 & 29.0 & 25.9 & 30.0 & 24.0 & 25.5 & 30.0 & 32.0 & 28.3 & 48.3 & 46.0 & 62.9 & 19.3 \\
GeoLLaVA-8K~\cite{wang2025geollava8k} & 25.0 & 24.0 & 25.0 & 25.0 & 25.0 & 25.0 & 21.4 & 25.0 & 25.0 & 24.5 & 24.0 & 0.0 & 0.0 & \textcolor{blue}{34.0} & 22.0 & 16.0 & 25.0 & 24.7 & 47.7 & 7.0 \\
VHM~\cite{pang2025vhm} & 25.5 & 25.0 & 26.0 & 26.5 & \textcolor{blue}{55.0} & 25.0 & 22.9 & 25.0 & 25.0 & 25.7 & 26.0 & 24.0 & 26.5 & \textcolor{blue}{34.0} & 28.0 & 27.7 & 45.0 & 53.3 & 46.2 & 16.7 \\
\midrule
\multicolumn{21}{l}{\textit{\textcolor{gray}{Open-source VLMs}}} \\
InternVL2.5-8B~\cite{chen2023internvl} & 25.5 & 22.0 & 26.0 & 26.0 & 22.5 & 24.5 & 30.0 & 22.5 & 20.0 & 24.3 & 26.0 & 20.0 & 22.5 & \textcolor{blue}{34.0} & 20.0 & 24.5 & 25.0 & 28.7 & 35.6 & 1.8 \\
InternVL 3.5 8B~\cite{wang2025internvl3} & 21.5 & \textcolor{blue}{28.0} & 18.0 & 21.5 & 29.0 & 28.5 & 30.0 & \textcolor{blue}{26.5} & 25.0 & 25.3 & 20.0 & 16.0 & 29.0 & \textcolor{blue}{34.0} & 26.0 & 25.0 & 30.0 & 22.7 & 40.2 & 6.1 \\
MiniCPM2\_6~\cite{yao2024minicpm} & 21.5 & \textcolor{blue}{28.0} & \textcolor{blue}{30.0} & 24.0 & 19.5 & 29.5 & 34.3 & 22.0 & 29.0 & 27.4 & 26.0 & 30.0 & \textcolor{red}{35.0} & 32.0 & 30.0 & 30.6 & 26.7 & 23.3 & 31.1 & 1.8 \\
Phi-3.5-Vision~\cite{abdin2024phi3} & 25.0 & 24.0 & 25.0 & 25.0 & 23.5 & 25.0 & 22.9 & 25.0 & 25.0 & 24.5 & 24.0 & 22.0 & 23.5 & \textcolor{red}{30.0} & 22.0 & 24.3 & 28.3 & 24.7 & 47.0 & 7.0 \\
Qwen2.5-VL-7B~\cite{Qwen2.5-VL} & \textcolor{red}{29.5} & 25.0 & 22.0 & 28.0 & 25.0 & 24.5 & 24.3 & \textcolor{blue}{26.5} & 22.0 & 25.2 & 26.0 & 28.0 & 25.0 & 10.0 & 20.0 & 21.8 & 21.7 & 24.0 & 10.6 & 0.0 \\
Deepseek-VL~\citep{lu2024deepseek} & 22.5 & 22.0 & 21.0 & 25.0 & 20.5 & 26.0 & 28.6 & 20.5 & 22.0 & 23.1 & 22.0 & 28.0 & \textcolor{red}{50.0} & 32.0 & 20.0 & 30.4 & 20.0 & 23.3 & 33.3 & 9.1 \\
VILA-HD~\cite{shi2025scaling} & 40.0 & 22.0 & 22.0 & 37.0 & 35.5 & 26.0 & 21.4 & 24.5 & 24.0 & 28.0 & \textcolor{red}{58.0} & 30.0 & \textcolor{blue}{55.0} & 32.0 & \textcolor{red}{58.0} & 46.6 & 65.0 & 57.3 & 57.6 & 24.8 \\
\midrule
\multicolumn{21}{l}{\textit{\textcolor{gray}{Closed-source VLMs}}} \\
GPT5~\cite{openai2025gpt5thinking} & 29.0 & 10.0 & 23.0 & 23.0 & 37.0 & 24.5 & 31.4 & 20.0 & 23.0 & 24.5 & \textcolor{blue}{74.0} & \textcolor{blue}{58.0} & 35.0 & \textcolor{blue}{34.0} & \textcolor{blue}{66.0} & \textcolor{blue}{53.4} & \textcolor{blue}{78.3} & \textcolor{blue}{73.3} & \textcolor{blue}{86.4} & \textcolor{blue}{52.6} \\
GPT-4o~\cite{hurst2024gpt} & 49.5 & 23.0 & 15.0 & 35.5 & 30.5 & 28.0 & 27.1 & 22.5 & \textcolor{blue}{41.0} & 30.2 & \textcolor{red}{68.0} & 56.0 & 30.5 & 32.0 & 64.0 & 50.1 & 70.0 & 72.0 & 84.1 & 47.4 \\
GPT-4o-mini~\cite{hurst2024gpt} & 41.5 & 16.0 & 29.0 & 31.5 & 31.5 & 32.0 & 28.6 & 19.5 & 32.0 & 29.1 & \textcolor{red}{54.0} & \textcolor{red}{54.0} & 31.5 & 48.0 & \textcolor{red}{54.0} & 48.3 & \textcolor{blue}{78.3} & 68.0 & 75.0 & 41.2 \\
Gemini-2.5-pro~\cite{comanici2025gemini} & \textcolor{blue}{55.0} & 18.0 & 31.0 & \textcolor{blue}{40.0} & 41.5 & \textcolor{blue}{32.5} & \textcolor{blue}{45.7} & 25.0 & 25.0 & \textcolor{blue}{34.9} & \textcolor{red}{66.0} & 32.0 & 41.5 & 38.0 & 50.0 & 45.5 & 56.7 & 60.0 & 57.6 & 32.6 \\
\midrule
\multicolumn{21}{l}{\textit{\textcolor{gray}{LLMs}}} \\
\rowcolor{cyan!20} Llama3-8B~\cite{llama3modelcard} & 23.0 & 22.0 & 35.0 & 22.0 & 27.5 & 25.0 & 22.9 & 21.5 & 29.0 & 25.3 & 30.0 & 30.0 & 27.5 & 34.0 & 48.0 & 33.9 & 58.3 & 50.0 & 60.6 & 16.7 \\
\rowcolor{cyan!20} Qwen3-8B~\cite{yang2025qwen3} & 38.5 & 28.0 & 26.0 & 36.5 & 31.0 & 24.5 & 30.0 & 25.5 & 48.0 & 32.0 & 42.0 & 26.0 & 31.0 & 56.0 & 56.0 & 42.2 & 53.3 & 58.0 & 57.6 & 17.5 \\

\bottomrule
\end{tabular}
}
\end{table*}

\paragraph{Metrics.}
For multiple-choice VQA, we report \emph{accuracy} by comparing the model's predicted option to the ground-truth choice (A/B/C/D). For open-ended VQA, we use GPT-4o~\cite{hurst2024gpt} as an automatic judge to score each response on a 1--100 scale, measuring agreement with human-annotated references. Scoring follows an expert-judge rubric: it prioritizes consistency with the reference, accepts semantically equivalent phrasing and modest numeric/unit tolerance, and penalizes only material hallucinations. Responses with scores $\geq 80$ are counted as correct.
Following recent work~\cite{Wang2025XLRSBench,li2024vrsbench}, we report BLEU-1, 2, 3, and 4, METEOR, and ROUGE-L to assess caption quality for image captioning.
For multi-turn evaluation, we report a \emph{dialog-level exact match} (\MTEMpc). 
Let \(\mathcal{D}\) be the set of evaluated dialogs. For each dialog \(d\), let \(n_d\) be the number of valid turns, and for each turn \(i=1,\ldots,n_d\) define a per-turn correctness \(z_{d,i}\in[0,1]\) as \(z_{d,i}=\mathbf{1}\!\left[\hat{y}_{d,i}=y_{d,i}\right]\) for discrete choice, or \(z_{d,i}=s_{d,i}/100\) for scored evaluations with raw scores \(s_{d,i}\in\{1,\ldots,100\}\). The strict dialog-level “all-correct” metric that unifies MTEM@1 and MTEM@80 is
\begin{equation}
\label{eq:mt-em}
\mathrm{MTEM}@t
=\frac{100}{|\mathcal{D}|}\sum_{d\in\mathcal{D}}
\mathbf{1}\!\left[\min_{1\le i\le n_d} z_{d,i}\,\ge\, t\right].
\end{equation}
Here \(t\in\{1,\,0.8\}\) corresponds to MTEM@1 and MTEM@80, respectively (i.e., all valid turns must be exactly correct for \(t{=}1\), or each turn must score at least \(80\) on the raw 1–100 scale for \(t{=}0.8\)).

\textbf{Results on RSHR-Bench (multiple-choice).}
As shown in Table~\ref{tab:visual_reasoning_perception_table2}, we evaluate four model families (including remote-sensing VLMs, open-source VLMs, closed-source VLMs, and text-only LLMs) across nine \emph{Perception} subtasks, five \emph{Reasoning} subtasks, and three \emph{Multi-turn} tasks. 
We observe that Gemini-2.5-pro~\cite{comanici2025gemini} shows highest accuracy on low-level perception, and GPT5~\cite{openai2025gpt5thinking} delivers the best overall performance (AD/FP/OSJ~=~74.0/58.0/66.0). 
Open-source models mostly remain around 25\% accuracy and struggle on compositional reasoning tasks. 
GeoLLaVA-8K~\citep{wang2025geollava8k} handles only multiple-choice questions and heavily favors option~A, resulting in MRJC and MRJCS accuracies of 0.
Notably, \textsc{VILA-HD}~\cite{shi2025scaling}, which supports 4K inputs, markedly outperforms other open-source models on reasoning tasks (Avg.~58.0). 
For multi-turn dialog, \textsc{GPT-5} is strongest (MAD/MTFP/MOSJ~=~78.3/73.3/86.4; MTEM@1~=~52.6). 
Even after multiple rounds of manual review to ensure all questions require visual evidence, text-only LLMs (Qwen3-8B~\cite{yang2025qwen3} and Llama3-8B~\cite{llama3modelcard}) still reach over 30\% accuracy on reasoning.

\begin{table*}[h]
\footnotesize

\caption{\textbf{Results across visual \emph{Perception}, \emph{Reasoning}, and \emph{Multi-turn}.} 
Task abbreviations follow Table~\ref{tab:visual_reasoning_perception_table2}. 
\textbf{Avg.} (Perception/Reasoning) denotes the mean over all Perception/Reasoning task columns, respectively.
\emph{Highlighting:} column-wise maxima \emph{(excluding the two LLM rows)} are shown in \textcolor{blue}{blue}; row-wise maxima—computed only over \textbf{Perception} and \textbf{Reasoning} task columns \emph{(excluding both \emph{Avg.} columns, all \emph{Multi-turn}/\emph{MTEM@80}, and the two LLM rows)}—are shown in \textcolor{red}{red}. Ties are highlighted; if both rules apply, \textcolor{blue}{blue} takes precedence.}
\label{tab:visual_reasoning_perception_VQA}
\centering
\resizebox{\textwidth}{!}{
\begin{tabular}{l|ccccccccc|c|ccccc|c|ccc|c}
\toprule
\multirow{2}{*}{\textbf{Model}} &
\multicolumn{9}{c}{\textbf{Perception}} &
\multirow{2}{*}{\textbf{Avg.}} &
\multicolumn{5}{c}{\textbf{Reasoning}} &
\multirow{2}{*}{\textbf{Avg.}} &
\multicolumn{3}{c}{\textbf{Multi-turn}} &
\multirow{2}{*}{\shortstack{\textbf{MTEM}\\\textbf{@80}}}\\
\cmidrule(lr){2-10}\cmidrule(lr){12-16}\cmidrule(lr){18-20}
& \textbf{COL} & \textbf{SHP} & \textbf{DET} & \textbf{OC} & \textbf{REL} & \textbf{OGD} & \textbf{RG} & \textbf{OCN} & \textbf{RCN}
& 
& \textbf{AD} & \textbf{FP} & \textbf{MRJC} & \textbf{MRJCS} & \textbf{OSJ}
& 
& \textbf{MAD} & \textbf{MTFP} & \textbf{MOSJ}
& 
\\
\midrule 
\multicolumn{21}{l}{\textit{\textcolor{gray}{Remote Sensing VLMs}}} \\
EarthDial~\cite{soni2025earthdial} & 34.3 & 35.0 & \textcolor{blue}{42.4} & 50.8 & 52.7 & 28.6 & 25.5 & 24.0 & 45.3 & 37.6 & 55.4 & \textcolor{red}{62.6} & 58.0 & \textcolor{blue}{58.7} & 57.9 & 48.7 & 65.2 & 65.7 & 65.9 & 11.4 \\
GeoChat~\cite{kuckreja2024geochat} & 35.7 & 40.5 & 41.1 & 47.0 & \textcolor{red}{54.9} & 24.6 & 22.6 & 35.5 & 44.2 & 38.5 & 49.8 & 39.9 & 24.5 & 24.0 & 37.6 & 41.5 & 53.9 & 58.3 & 55.3 & 2.6 \\
VHM~\cite{pang2025vhm} & 49.3 & 42.9 & 39.6 & \textcolor{red}{63.0} & 48.2 & 24.4 & 23.5 & 29.1 & \textcolor{blue}{54.3} & 41.6 & 60.6 & 62.7 & 38.8 & 35.2 & 51.4 & 48.7 & 60.9 & 63.1 & 71.3 & 16.7 \\
\midrule
\multicolumn{21}{l}{\textit{\textcolor{gray}{Open-source VLMs}}} \\
InternVL2.5-8B~\cite{chen2023internvl} & 53.2 & 37.6 & 31.6 & 63.6 & 59.1 & 26.9 & 30.1 & \textcolor{blue}{46.1} & 41.8 & 43.3 & 74.3 & \textcolor{blue}{74.7} & 58.3 & 48.5 & 61.8 & 52.6 & 67.6 & 71.3 & 72.4 & 24.8 \\
InternVL3.5-8B~\cite{wang2025internvl3} & 37.1 & \textcolor{blue}{44.6} & 36.8 & \textcolor{blue}{68.1} & 54.2 & 26.9 & 31.7 & 41.1 & 34.3 & 41.6 & 66.0 & 67.6 & \textcolor{blue}{68.0} & 43.0 & 58.6 & 51.5 & 70.0 & 71.1 & 66.7 & 19.3 \\
MiniCPM2\_6~\cite{yao2024minicpm} & 49.3 & 38.6 & 37.8 & 60.8 & 58.9 & 23.2 & 29.0 & 35.0 & 33.0 & 40.6 & \textcolor{red}{71.3} & 70.9 & 51.8 & 43.1 & 60.2 & 51.7 & 63.5 & 75.1 & 74.1 & 30.7 \\
Phi-3.5-Vision~\cite{abdin2024phi3} & 40.4 & 32.2 & 39.8 & 61.1 & 56.5 & 28.9 & 26.6 & 36.3 & 32.3 & 39.3 & \textcolor{red}{67.5} & 67.1 & 43.6 & 40.1 & 55.2 & 50.3 & 64.5 & 70.8 & 75.1 & 22.8 \\
Qwen2.5-VL-7B~\cite{Qwen2.5-VL} & 35.1 & 34.6 & 30.0 & 55.5 & 57.9 & 26.0 & 30.2 & 33.2 & 28.3 & 36.8 & 58.8 & 37.4 & \textcolor{red}{67.1} & 33.5 & 51.4 & 40.2 & 55.9 & 49.6 & 54.2 & 2.6 \\
DeepSeek-VL~\citep{lu2024deepseek} & 54.1 & 35.2 & 34.2 & 63.5 & 60.2 & 25.4 & 24.5 & 38.5 & 33.9 & 41.1 & \textcolor{red}{70.4} & 60.1 & 46.8 & 35.2 & 55.1 & 49.5 & 67.7 & 68.4 & 60.2 & 19.3 \\
VILA-HD~\cite{shi2025scaling} & \textcolor{blue}{58.4} & 40.7 & 33.4 & 56.8 & \textcolor{blue}{67.8} & 27.7 & 29.6 & 43.9 & 37.9 & \textcolor{blue}{44.0} & 62.3 & 60.1 & 62.4 & 32.5 & 56.3 & \textcolor{blue}{53.0} & 71.1 & 74.1 & 78.6 & 67.5 \\
\midrule
\multicolumn{21}{l}{\textit{\textcolor{gray}{Closed-source VLMs}}} \\
GPT5~\cite{openai2025gpt5thinking} & 25.6 & 26.6 & 37.6 & 49.0 & 52.4 & 20.9 & 29.9 & 34.8 & 27.4 & 33.8 & \textcolor{red}{70.7} & 62.5 & 48.2 & 44.6 & 60.1 & 48.0 & 75.0 & \textcolor{blue}{76.8} & 59.8 & \textcolor{blue}{72.8} \\
GPT-4o~\cite{hurst2024gpt} & 55.6 & 35.0 & 33.8 & 56.5 & 46.6 & \textcolor{blue}{29.9} & \textcolor{blue}{35.4} & 35.1 & 40.8 & 41.0 & \textcolor{blue}{74.4} & 41.0 & 28.6 & 55.8 & 53.8 & 49.6 & 54.5 & 74.9 & 76.0 & 68.9 \\
GPT-4o-mini~\cite{hurst2024gpt} & 29.3 & 37.6 & 31.8 & 62.2 & 45.9 & 28.4 & 23.0 & 30.1 & 24.3 & 34.7 & 70.5 & \textcolor{red}{72.2} & 65.0 & 43.9 & \textcolor{blue}{63.9} & 51.0 & \textcolor{blue}{78.0} & 74.9 & \textcolor{blue}{82.3} & 72.5 \\
Gemini-2.5-pro~\cite{comanici2025gemini} & 34.4 & 25.5 & 23.3 & 38.0 & 51.2 & 18.4 & 17.3 & 25.4 & 16.6 & 27.8 & 42.7 & 30.8 & \textcolor{red}{60.4} & 29.0 & 41.7 & 29.6 & 35.0 & 20.3 & 41.9 & 20.6 \\
\midrule
\multicolumn{21}{l}{\textit{\textcolor{gray}{LLMs}}} \\
\rowcolor{cyan!20}
Llama3-8B~\cite{llama3modelcard} & 21.2 & 31.0 & 28.7 & 28.6 & 20.2 & 28.7 & 28.6 & 20.3 & 26.3 & 26.0 & 42.5 & 33.6 & 18.2 & 21.6 & 28.6 & 28.6 & 34.8 & 38.5 & 35.3 & 0.0 \\
\rowcolor{cyan!20}
Qwen3-8B~\cite{yang2025qwen3} & 26.3 & 29.1 & 25.7 & 54.8 & 34.1 & 23.2 & 24.1 & 18.2 & 21.9 & 28.6 & 65.9 & 68.1 & 21.6 & 22.6 & 45.0 & 38.5 & 57.8 & 57.2 & 57.3 & 5.3 \\
\bottomrule
\end{tabular}
}
\vspace{-0.3cm}
\end{table*}

\begin{table}[h]
\footnotesize
\caption{
Experimental results with text-generation metrics. Closed-source models are highlighted in gray.
}
\label{tab:gen-metrics_image_caption}
\centering
\resizebox{0.46\textwidth}{!}{%
\begin{tabular}{l|cccccc}
\toprule
\textbf{Method} &
\textbf{BLEU-1} & \textbf{BLEU-2} & \textbf{BLEU-3} & \textbf{BLEU-4} & \textbf{METEOR} & \textbf{ROUGE-L} \\
\midrule
InternVL2.5-8B~\cite{chen2023internvl}      & 11.0 & 4.7 & 1.7 & 0.6 & 17.0 & 17.3 \\
InternVL 3.5 8B~\cite{wang2025internvl3}    & 33.7 & 14.4 & 6.1 & 2.5 & 26.4 & 23.5 \\
MiniCPM2\_6~\cite{yao2024minicpm}          & 2.9 & 1.2 & 0.5 & 0.2 & 11.8 & 13.6 \\
Phi-3.5-Vision~\cite{abdin2024phi3}        & 36.5 & 14.0 & 5.6 & 2.1 & 25.1 & 22.3 \\
Qwen2.5-VL-7B~\cite{Qwen2.5-VL}               & 0.1 & 0.0 & 0.0 & 0.0 & 3.8 & 6.1 \\
DeepSeek-VL~\citep{lu2024deepseek}         & 2.7 & 1.2 & 0.5 & 0.2 & 11.8 & 14.6 \\
VILA-HD~\cite{shi2025scaling}              & 28.0 & 10.0 & 4.2 & 1.7 & 21.6 & 21.8 \\
\midrule
\Lgray GPT5~\cite{openai2025gpt5thinking}   & 43.3 & 15.3 & 5.0 & 1.6 & 30.1 & 20.6 \\
\Lgray GPT-4o-mini~\cite{hurst2024gpt}     & 49.7 & 21.8 & 10.0 & 4.8 & 33.7 & 25.8 \\
\Lgray GPT-4o~\cite{hurst2024gpt}          & 43.8 & 18.1 & 7.9 & 3.6 & 29.2 & 23.1 \\
\Lgray Gemini-2.5-pro~\cite{comanici2025gemini}                     & 9.1 & 3.0 & 0.9 & 0.3 & 7.5 & 7.1 \\
\bottomrule
\end{tabular}%
}
\end{table}

\begin{figure}[t]
  \centering
  \begin{subfigure}[b]{0.485\linewidth}
    \centering
    \includegraphics[width=\linewidth]{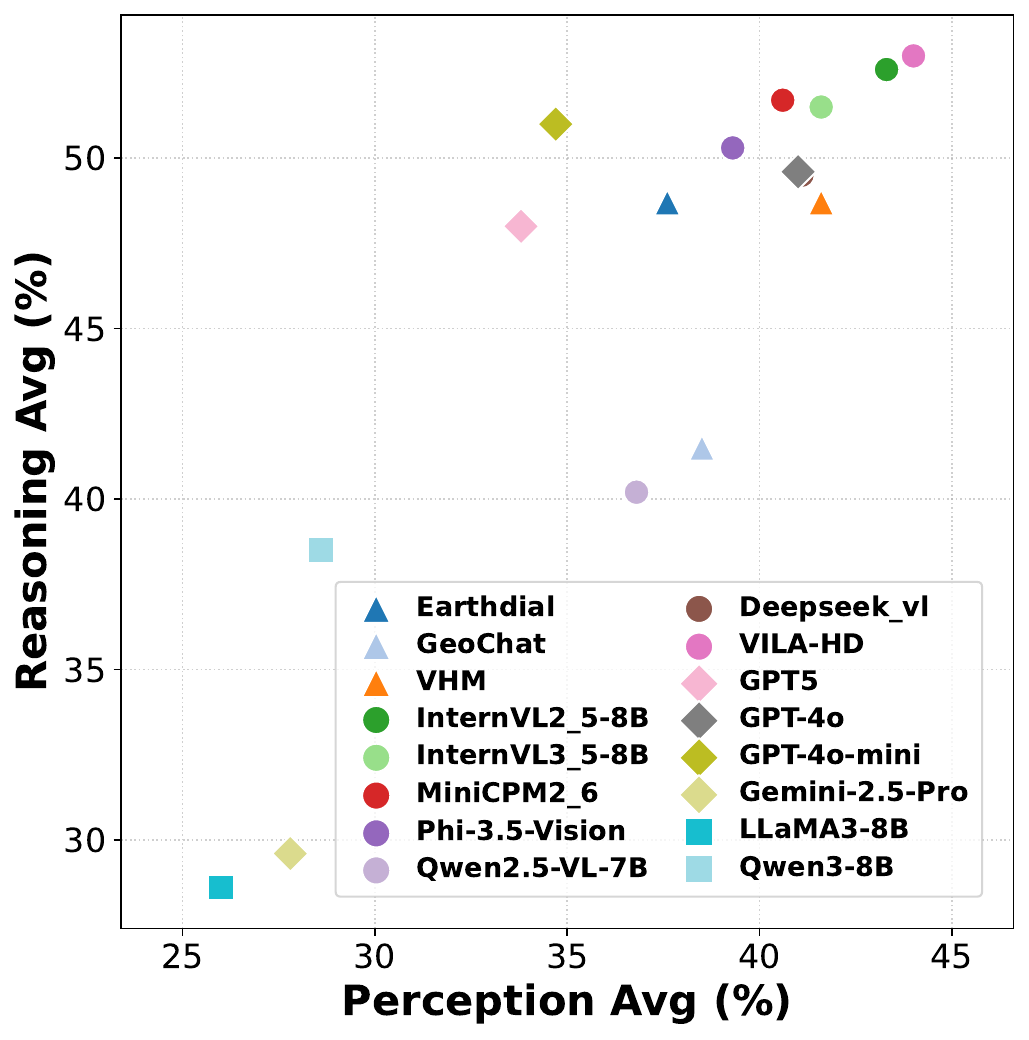}
  \end{subfigure}
  \hspace{0.1mm}
  \begin{subfigure}[b]{0.485\linewidth}
    \centering
    \includegraphics[width=\linewidth]{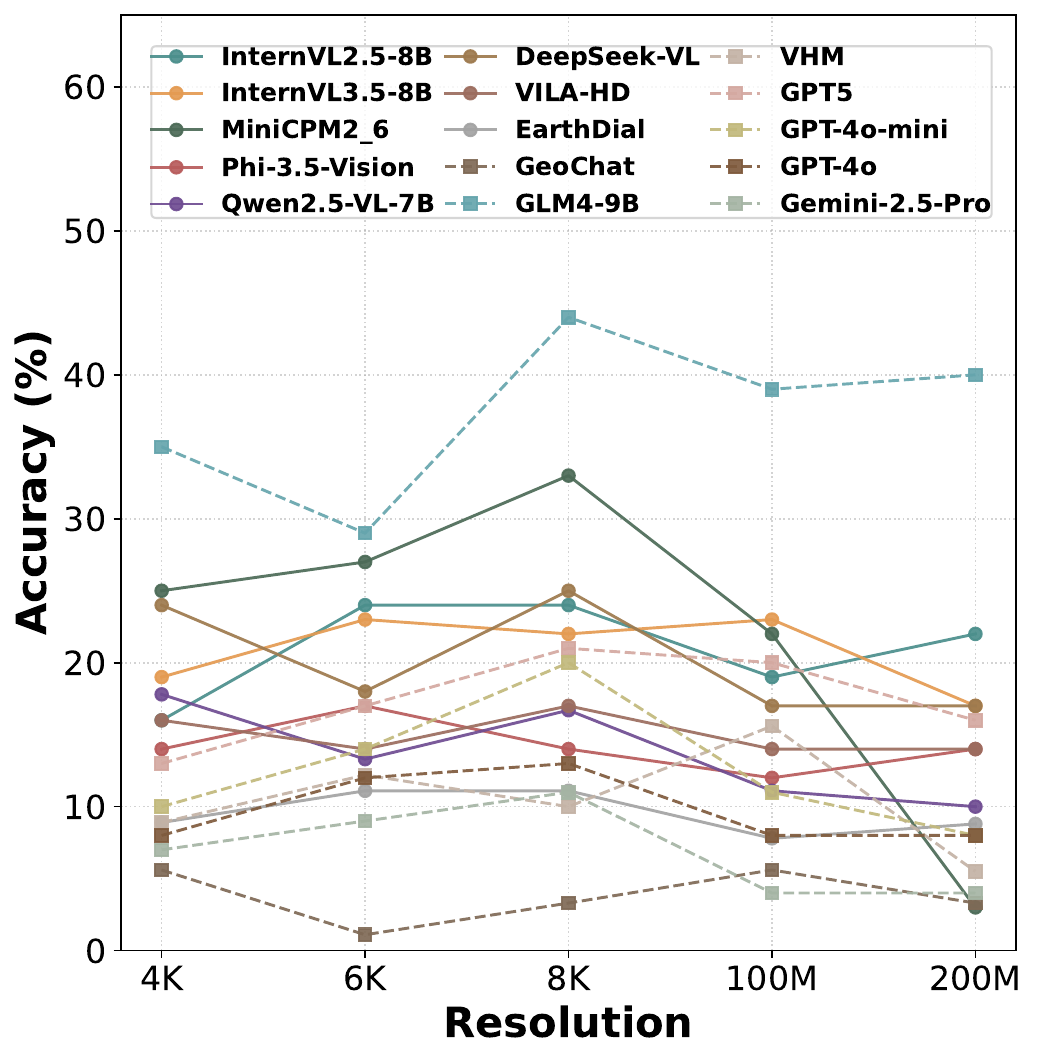}
  \end{subfigure}
    \caption{Experiment results of different models on RSHR-Bench. Left: Model performance on open-ended VQA. Right: Model performance on single-image evaluation.}
  \label{fig:211}
  \vspace{0mm}
\end{figure}

\vspace{-0.2cm}
\textbf{Results on RSHR-Bench (open-ended).}
To prevent models from using option priors in multiple-choice formats, we convert each item into an option-free, open-ended question.
Table~\ref{tab:visual_reasoning_perception_VQA} reports per-task accuracies, and the multi-turn exact match at the 80/100 threshold (\textbf{MTEM@80}). 
As shown in Fig.~\ref{fig:211} (left), all models score below 50 on perception tasks, indicating nearly no correct relevant responses and highlighting the lack of perception abilities on ultra-high-resolution images. On reasoning tasks, both closed-source models and open-source VLMs score around 50 on average, while performance drops notably on MRJC (Multi-Region Joint Contrast, multi-image) and MRJCS (Multi-Region Joint Contrast, single-image, multi-box).
Within \emph{remote-sensing VLMs}, open-source and closed-source VLMs achieve comparable scores.
For multi-turn dialogs, closed-source models achieve strong results: GPT5~\cite{openai2025gpt5thinking}, GPT-4o~\cite{hurst2024gpt}, and GPT-4o-mini~\cite{hurst2024gpt}, along with the open-source VILA-HD, perform well on the dialog-level metric MTEM@80.
For Qwen3-8B~\cite{yang2025qwen3} and Llama3-8B
~\cite{llama3modelcard}, the scores average around \(30\), which indicates incorrect responses with substantial hallucinations.

\textbf{Results on Image Caption tasks.}
As shown in Table~\ref{tab:gen-metrics_image_caption}, we evaluate widely-used metrics of image caption tasks.
The results show that the closed-source VLMs outperform open‐source baselines across all text–generation metrics. 
GPT-4o-mini~\cite{hurst2024gpt} achieves the best overall scores—BLEU-4 (4.8), METEOR (33.7), and ROUGE-L (25.8).
Notably,  GPT5~\cite{openai2025gpt5thinking} attains strong BLEU-1 (43.3) and METEOR (30.1) yet a comparatively low BLEU-4 (1.6), suggesting paraphrastic or syntactically freer captions that reduce exact n-gram overlap. 
The Gemini-2.5-pro~\cite{comanici2025gemini} and Qwen2.5-VL-7B~\cite{Qwen2.5-VL}  underperform by a large margin (yielding near-zero BLEU for the latter), indicating potential domain mismatches under our evaluation setup.

\textbf{Results on Single Image Evaluation.}
As summarized in Fig.~\ref{fig:211} (right), we evaluate all models across practical high resolutions (4K--8K) and the ultra-high-pixel regime (100M/200M). 
Performance is highly sensitive to resolution: accuracy is modest at 4K--8K and drops markedly at 100M/200M. 
Across open- and closed-source general models as well as remote-sensing VLMs, results remain uniformly low (around 30\%), indicating limited robustness to extreme pixel counts and large spatial contexts. 
Overall, current VLMs still struggle to understand high-resolution remote-sensing imagery reliably.


\section{Conclusion}
We present \textbf{RSHR-Bench}, a large-scale benchmark for vision--language understanding in ultra--high-resolution remote sensing imagery. 
RSHR-Bench preserves native resolutions up to $\sim 3\times10^{8}$ pixels and provides a comprehensive, fair evaluation of both general-purpose and remote-sensing VLMs. 
Experiments on a broad range of open- and closed-source models reveal uniformly low performance. 
We hope RSHR-Bench will serve as a challenging benchmark for future models that can bridge this gap toward real-world remote-sensing applications.

\section{Overview of the Appendix}
\label{sec:app-overview}
This appendix supplements the proposed \textbf{RSHR-Bench} with details that are omitted from the main paper due to space constraints.
The remainder of the appendix is organized as follows.

Sec.~\ref{sec:human-annotation} describes the human annotation pipeline and evaluation protocol, including the two-stage manual labeling and review process, as well as the human evaluation.

Sec.~\ref{sec:benchmark-details} provides additional details of our benchmark design, including the task taxonomy and question answer formats, the input resolution policies of different models, the image captioning evaluation, and the UAV data source.

Sec.~\ref{app-details} reports additional results on other benchmarks, including MME-RealWorld, XLRS-Bench, and LRS-VQA, and analyzes robustness to input resolution.

Finally, Sec.~\ref{sec:case-study} presents more case studies and the detailed prompt templates of various tasks.

\section{Human annotation and evaluation}
\label{sec:human-annotation}

\textbf{Details of annotation stage.}
The process consists of two phases. 
In the first phase, three undergraduate students spent approximately 50 person-hours drawing bounding boxes and authoring answers for a subset of questions. 
For the single-image setting, we first stratify candidate images into five resolution bins according to the long-side length: 4K, 6K, 8K, 100M, and 200M pixels (e.g., the 4K bin contains images whose long side is at least 4{,}000 pixels, with the remaining bins defined analogously). 
Different tasks sample images from these bins with task-dependent proportions: tasks that intrinsically rely on fine spatial detail, such as shape or margin recognition, are biased toward higher-resolution bins. 
Within each selected image, annotators choose target objects that are usable across multiple tasks and that exhibit a graded difficulty distribution, for example by varying the ratio of the object’s pixel area to the full image area.
In the second phase, a different group of undergraduate students from the same program (disjoint from the first three annotators) performs a second-round review. They check all annotated items and either revise or discard those that are potentially ambiguous or factually incorrect. This process ensures both correctness and strict visual grounding. After this two-stage process, each annotated task types contain approximately 100--200 validated items, with a controlled mixture of resolutions across the five bins.

\textbf{Details of human evaluation.}
We conduct a human evaluation to estimate human answer accuracy on our curated question sets. 
Concretely, we design 17 task types, grouped into \emph{Perception} (color (COL), shape (SHP), detection (DET), object classification (OC), relation (REL), object grounding (OGD), regional grounding (RG), object counting (OCN), regional counting (RCN)), \emph{Reasoning} (anomaly detection (AD), future prediction with two images (FP), multi-region joint contrast (MRJC), multi-region joint contrast single (MRJCS), object state judgement (OSJ)), and \emph{Multi-turn} (multi-turn anomaly detection (MAD), multi-turn future prediction (MTFP), multi-turn object state judgement (MOSJ)). 
For each task, we sample 15 image–query instances, resulting in 255 instances in total. 
Human annotators are shown the specific task description, the input images, and the questions, and are asked to answer them without seeing the reference answers or model predictions. For each question, the raters are distinct from both the original annotators and the reviewers who participated in its construction.
Accuracy for each task is computed as the fraction of instances judged correct (e.g., COL/SHP/DET/REL/RG/OCN all achieve 15/15, OGD 14/15, OC and RCN 13/15, AD 10/15, FP 12/15, MRJC and MRJCS 15/15, OSJ 14/15, MAD and MTFP 13/15, MOSJ 15/15). 
Overall, the model answers 237 out of 255 instances correctly, corresponding to an average human-evaluated accuracy of 92.94\%.

\textbf{Details of human\&LLM verification.} For all tasks, we first obtain candidate questions and options either by prompting models (Qwen2.5\mbox{-}VL\mbox{-}8B and GPT\mbox{-}5 Thinking) with human-annotated boxes and labels, or by directly feeding images and prompts to these models to draft question–answer pairs. We then apply a two-stage verification pipeline. In the first stage, human annotators check every question, option set, and answer. They correct any factual errors or mismatches between the image and the labeled answer. In the second stage, we perform text-only validation: LLMs are asked to answer each question without access to the image. If their accuracy in this setting is high (typically above \(30\%\)), we treat the item as overly solvable from language alone and revise it using the model explanations as guidance. During revision, we remove lexical hints and reduce dependence on commonsense priors; for example, in a prompt such as ``determine the color of the plane within the bounding box'' with the answer ``white'', we replace ``plane'' with a neutral phrase like ``main object''. We also correct for label biases observed in text-only runs: in the orientation task, for instance, Qwen3\mbox{-}8B tends to prefer ``top-right'', so we may change the correct orientation to ``bottom-left'' or select a different object. These adjustments lower text-only accuracy, typically to below the \(30\%\) threshold or to a point where further reduction becomes impractical. Overall, this two-stage procedure preserves strict visual grounding while maintaining a well-calibrated level of task difficulty.

\section{Details of our benchmark}
\label{sec:benchmark-details}

\textbf{Input resolution.}
Table~\ref{tab:model_resolution_policy} summarizes the maximum input resolution or official resolution reports of the evaluated models. 
Only GeoLLaVA-8K and Claude3.7 explicitly support very high-resolution inputs up to \(8000\times 8000\).
Most open-source VLMs remain constrained by tiling at around \(448\times 448\). 
For closed-source models, the providers do not reveal the exact limits, so we mark their resolution as \emph{Not disclosed}. 
Many remote sensing models also rely on CLIP-based and ViT-based visual encoders, which in practice operate on tiled inputs around \(336\times 336\). 
However, since the original papers usually do not state this policy clearly, we conservatively label these cases as \emph{Not disclosed}.

\begin{table}[h]
\centering
\scriptsize
\caption{Reported maximum image size for each model.}
\label{tab:model_resolution_policy}
\begin{tabular}{ll}
\toprule
\textbf{Model} & \textbf{Max input resolution} \\
\midrule
GeoLLaVA-8K~\cite{wang2025geollava8k} & \(8000\times 8000\) \\
GeoChat~\cite{kuckreja2024geochat} & Not disclosed \\
EarthDial~\cite{soni2025earthdial} & Not disclosed \\
VHM~\cite{pang2025vhm} & Not disclosed \\
InternVL3-8B~\cite{wang2025internvl3} &  \(448\times 448\) tiling \\
QwenVL2.5-7B~\cite{Qwen2.5-VL} & Dynamic tokenization \\
Claude3.7~\cite{anthropic2024claude} & \(8000\times 8000\) \\
GPT-4o~\cite{hurst2024gpt} & Not disclosed  \\
\bottomrule
\end{tabular}
\end{table}

\textbf{Image captioning.}
We further evaluate image captioning performance on a dataset of 3{,}913 images.
For efficiency, the main paper reports results on a 1{,}000-image subset, as querying commercial APIs for closed-source models is costly.
The full results on the 3{,}913-image set are provided in the supplementary material.
For each image, we first use GPT-5-Thinking to generate reference captions, and then compute standard image caption metrics---BLEU (1--4), METEOR, and ROUGE-L---between model outputs and these references.
Table~\ref{tab:gen-metrics_image_caption} summarizes the results, where higher scores indicate better caption quality.

\begin{table}[h]
\footnotesize
\caption{
Experimental results with image caption metrics. 
}
\label{tab:gen-metrics_image_caption}
\centering
\resizebox{0.4\textwidth}{!}{%
\begin{tabular}{l|cccccc}
\toprule
\textbf{Method} &
\textbf{BLEU-1} & \textbf{BLEU-2} & \textbf{BLEU-3} & \textbf{BLEU-4} & \textbf{METEOR} & \textbf{ROUGE-L} \\
\midrule
InternVL2.5-8B~\cite{wang2025internvl3}      & 33.3 & 14.4 & 6.2 & 2.8 & 26.2 & 24.1 \\
InternVL3.5-8B~\cite{wang2025internvl3}      & 33.4 & 14.3 & 6.0 & 2.4 & 26.2 & 23.4 \\
DeepSeek-VL~\citep{lu2024deepseek}                 & 2.7  & 1.2  & 0.5 & 0.2 & 11.8 & 14.6 \\
GLM4-9B~\cite{glm2024chatglm}                          & 37.3 & 11.5 & 3.5 & 1.1 & 23.8 & 19.0 \\
MiniCPM2\_6~\cite{yao2024minicpm}                  & 2.9  & 1.2  & 0.5 & 0.2 & 11.7 & 13.6 \\
Phi-3.5-Vision~\cite{abdin2024phi3}         & 36.5 & 14.0 & 5.6 & 2.1 & 25.1 & 22.3 \\
Qwen2.5-VL-7B~\cite{Qwen2.5-VL}            & 0.1  & 0.0  & 0.0 & 0.0 & 3.8  & 6.1  \\
VILA-HD~\cite{shi2025scaling}                        & 28.1 & 10.0 & 4.2 & 1.7 & 21.6 & 21.8 \\

\bottomrule
\end{tabular}%
}

\end{table}

\textbf{Task taxonomy and QA formats.}
Table~\ref{tab:task_taxonomy} summarizes the hierarchical task design of RSHR-Bench, from high-level L1 categories (Perception, Reasoning, Image Caption) to finer-grained L2/L3 tasks. 
For each task, we list the task code, annotation protocol (all-human vs.\ semi-automated), question generator, interaction pattern (e.g., single-/multi-image, single-/multi-turn, or multi-box grounding), answer format (multiple-choice and/or open-ended), and the number of question instances. 
This taxonomy jointly covers low-level perceptual skills (e.g., color, shape, counting), higher-level logic- and knowledge-grounded reasoning (e.g., anomaly interpretation, future prediction), and both overall and regional captioning, providing a comprehensive evaluation of remote-sensing VLM capabilities across perception and reasoning.

\textbf{UAV data source.} 
We collect the ultra-high-resolution data captured by a drone (UAV) in outdoor public spaces. 
All data are collected in a public park and a nearby drone operation site. 
Each sequence is recorded in a single continuous flight at heights between 120\,m and 150\,m. 
The UAV follows a smooth trajectory and observes the same region for an extended period. 
The camera looks at the scene from multiple headings, with both upward-looking and downward-looking views, so the frames form a long, continuous, multi-view stream of the same scene.  
Each frame contains about $10^{8}$ pixels (around 100 megapixels), and the scenes are mainly park scenes and drone operation site scenes. 
We use the images for image captioning and visual question answering tasks, where the continuous multi-view structure provides rich context for temporal and multi-view reasoning.

\begin{table*}[htbp]
\footnotesize
\caption{Overview of task taxonomy, annotation sources, and QA formats in our benchmark. L1/L2/L3 denote coarse-, mid-, and fine-grained task levels, respectively, and \emph{Attr.} lists the attribute codes for each L3 task. We also summarize the annotation source, question generator, question type, answer type, and the number of questions (\#Q) for each task.}
\label{tab:task_taxonomy}
\centering
\resizebox{0.9\textwidth}{!}{
\begin{tabular}{l|l|l|c|c|l|l|l|r}
\toprule
\Gray
\textbf{L1-task} &
\textbf{L2-task} &
\textbf{L3-task} &
\textbf{Attr.} &
\textbf{Annotation} &
\textbf{Question generator} &
\textbf{Question type} &
\textbf{Answer type} &
\textbf{\#Q} \\
\midrule
\multirow{9}{*}{Perception}
  & \multirow{3}{*}{Local Attributes}
    & Color Detection                 & COL & All human      & Qwen2.5-VL-7B   & Single image Single turn & \makecell[l]{Multiple Choice (A/B/C/D)\\Open-ended text} & 443\\ \cmidrule(lr){3-9}
  & & Shape/Margin Recognition        & SHP & All human      & Qwen2.5-VL-7B   & Single image Single turn & \makecell[l]{Multiple Choice (A/B/C/D)\\Open-ended text} & 243\\ \cmidrule(lr){3-9}
  & & Orientation Detection           & DET & All human      & GPT-5 Thinking  & Single image Single turn & \makecell[l]{Multiple Choice (A/B/C/D)\\Open-ended text} & 228\\ \cmidrule(lr){2-9}

  & \multirow{2}{*}{Overall Attributes}
    & Object Classification           & OC  & All human      & Qwen2.5-VL-7B   & Single image Single turn & \makecell[l]{Multiple Choice (A/B/C/D)\\Open-ended text} & 436\\ \cmidrule(lr){3-9}
  & & Object Spatial Relationship     & REL & semi-automated & Qwen2.5-VL-7B   & Single image Single turn & \makecell[l]{Multiple Choice (A/B/C/D)\\Open-ended text} & 444\\ \cmidrule(lr){2-9}

  & \multirow{2}{*}{Visual Grounding}
    & Object Grounding                & OGD & All human      & GPT-5 Thinking  & Single image Single turn & \makecell[l]{Multiple Choice (A/B/C/D)\\Open-ended text} & 430\\ \cmidrule(lr){3-9}
  & & Regional Grounding              & RG  & All human      & GPT-5 Thinking  & Single image Single turn & \makecell[l]{Multiple Choice (A/B/C/D)\\Open-ended text} & 219\\ \cmidrule(lr){2-9}

  & \multirow{2}{*}{Counting}
    & Object Counting                 & OCN & semi-automated & GPT-5 Thinking  & Single image Single turn & \makecell[l]{Multiple Choice (A/B/C/D)\\Open-ended text} & 412\\ \cmidrule(lr){3-9}
  & & Regional Counting               & RCN & All human      & Qwen2.5-VL-7B   & Single image Single turn & \makecell[l]{Multiple Choice (A/B/C/D)\\Open-ended text} & 245\\
\midrule

\multirow{4}{*}{Reasoning}
  & \multirow{2}{*}{Logic-grounded Reasoning}
    & Multi-region Joint Contrast     & \makecell{MRJC\\MRJCS} & All human      & GPT-5 Thinking &
      \makecell[l]{Multi image Single turn\\Single image Multi box} &
      \makecell[l]{Multiple Choice (A/B/C)\\Open-ended text} & 140\\ \cmidrule(lr){3-9}
  & & Object State Judgement          & \makecell{OSJ\\MOSJ}  & semi-automated & GPT-5 Thinking &
      \makecell[l]{Single image Single turn\\Single image Multi turn} &
      \makecell[l]{Multiple Choice (A/B/C/D)\\Open-ended text} & 409\\ \cmidrule(lr){2-9}

  & \multirow{2}{*}{Knowledge-grounded Reasoning}
    & Anomaly Detection  & \makecell{AD\\MAD} & semi-automated & GPT-5 Thinking &
      \makecell[l]{Single image Single turn\\Single image Multi turn} &
      \makecell[l]{Multiple Choice (A/B/C/D)\\Open-ended text} & 246\\ \cmidrule(lr){3-9}
  & & Future Prediction               & \makecell{FP2I\\MTFP} & semi-automated & GPT-5 Thinking &
      \makecell[l]{Single image Single turn\\Single image Multi turn\\Multi image Single turn} &
      \makecell[l]{Multiple Choice (A/B/C/D)\\Multiple Choice (A/B)\\Open-ended text} & 429\\
\midrule

\multirow{2}{*}{Image Caption}
  & Overall Caption  &  & OCAP & semi-automated & GPT-5 Thinking & Single image Single turn & Open-ended text & \\
  & Regional Caption &  & RCAP & semi-automated & GPT-5 Thinking & Single image Single turn & Open-ended text & 3913\\
\bottomrule
\end{tabular}}
\end{table*}

\section{Result on other benchmark}
\label{app-details}

\textbf{Results on MME-RealWorld.}
As shown in Table~\ref{tab:mme_realworld-test}, we evaluate several language models and multimodal models without images on the remote sensing subset of MME-RealWorld, where all images are removed and the models receive only text inputs. 
Surprisingly, the best pure language model (Llama3-8B) still answers 31.22\% of the questions correctly, indicating that a substantial portion of the benchmark can be solved without access to the underlying imagery. 
This highlights the need to re-verify remote sensing tasks with strong LLMs and to construct benchmarks that genuinely require visual verification.

\begin{table}[t]
\footnotesize
\caption{Accuracy on the remote sensing subset of MME-RealWorld. All models are evaluated on remote sensing tasks using text-only inputs (no images).}
\label{tab:mme_realworld-test}
\centering
\resizebox{0.4\textwidth}{!}{
\begin{tabular}{lccc}
\toprule
\textbf{Method} & \textbf{Input Type} & \textbf{Perception (\%)} & \textbf{\#Correct / \#All} \\
\midrule
Llama3-8B~\cite{llama3modelcard}                 & No Image & 31.22 & 1167 / 3738 \\
GPT-5-All~\cite{openai2025gpt5thinking}          & No Image & 22.45 & 839 / 3738  \\
Gemini2.0-Flash~\cite{comanici2025gemini}        & No Image & 16.19 & 605 / 3738  \\
Qwen3-8B~\cite{yang2025qwen3}                    & No Image & 15.09 & 564 / 3738  \\
GPT-4o~\cite{hurst2024gpt}                       & No Image &  2.03 &  76 / 3738  \\
\bottomrule
\end{tabular}}
\end{table}

\textbf{Results on XLRS\mbox{-}Bench.}
As is shown in table~\ref{tab:xlrsbench_merged-XLRSBench}, we evaluate XLRS\mbox{-}Bench on image\mbox{-}based VQA using both image\mbox{-}conditioned multimodal models and text\mbox{-}only LLM baselines, including GPT-4o~\cite{hurst2024gpt}, Qwen3\mbox{-}8B~\cite{yang2025qwen3}, and Llama3-8B~\cite{touvron2023llama}. 
Under our unified evaluation protocol, text\mbox{-}only LLMs can be surprisingly competitive on Reasoning tasks: Qwen3\mbox{-}8B attains an average accuracy of \(\mathbf{51.6\%}\), surpassing the image\mbox{-}conditioned GPT\mbox{-}4o baseline (\(45.2\%\)). 
Drilling down by sub\mbox{-}task, Qwen3\mbox{-}8B achieves \(72.0\%\) on AD (Anomaly Detection) task and \(77.0\%\) on ECR (Existence \& Counting Reasoning) task, while Llama3-8B reaches \(48.0\%\) on RP (Route Planning) task, outperforming most vision\mbox{-}language models despite having no image input. 
These results highlight that high\mbox{-}level spatial reasoning and counting can be solved via robust priors and linguistic cues alone, although vision remains crucial for other tasks.

\begin{table*}[h]
\footnotesize
\caption{
Results on XLRS-Bench reasoning and perception dimensions.
Avg. denotes the average accuracy over the sub-tasks in that dimension.
A dash (--) indicates results not reported in the original paper.
An asterisk (*) indicates models evaluated without image input.}
\label{tab:xlrsbench_merged-XLRSBench}
\centering
\resizebox{0.8\textwidth}{!}{
\begin{tabular}{l|ccccc|c|cccccc|c}
\toprule
\Gray
\multicolumn{1}{l}{\textbf{Method}} &
\multicolumn{6}{c|}{\textbf{Reasoning}} &
\multicolumn{7}{c}{\textbf{Perception}} \\
\midrule
\Gray
 & \textbf{AD} & \textbf{ECR} & \textbf{RP} & \textbf{RCCD} & \textbf{CCR} & \textbf{Avg.} 
 & \textbf{OC} & \textbf{RC} & \textbf{RLUC} & \textbf{OCC} & \textbf{OCL} & \textbf{OSR} & \textbf{Avg.} \\
\midrule

\multicolumn{14}{l}{\textit{\textcolor{gray}{Remote Sensing MLLMs}}} \\
GeoChat~\cite{kuckreja2024geochat} & 33.0 & 43.0 & 10.0 & - & 21.0 & 26.8 & 16.7 & 29.0 & 23.0 & 21.1 & 16.8 & 24.2 & 21.8\\
GeoLLaVA-8K~\cite{wang2025geollava8k} & 67.0 & 72.0 & 67.0 & 28.3 & 21.0 & 51.1 & 16.7 & 29.0 & 66.0 & 37.4 & 28.5 & 35.4 & 35.5\\
Earthdial~\cite{soni2025earthdial} & 62.0 & 71.0 & 43.0 & 48.3 & 50.0 & 54.9 & 18.3 & 42.0 & 36.0 & 31.3 & 31.0 & 24.8 & 30.6\\
VHM~\cite{pang2025vhm} & 42.0 & 53.0 & 46.0 & 28.3 & 21.0 & 38.1 & 16.7 & 30.0 & 26.0 & 21.4 & 16.8 & 25.6 & 22.8\\

\midrule
\multicolumn{14}{l}{\textit{\textcolor{gray}{Closed-source MLLMs}}} \\
GPT-4o~\cite{hurst2024gpt} & 73.0 & 73.0 & 35.0 & 20.0 & 25.0 & 45.2 & 25.0 & 32.0 & 66.0 & 9.5 & 11.3 & 24.6 & 28.1\\
GPT-4o-mini~\cite{hurst2024gpt} & 71.0 & 71.0 & 29.0 & 6.7 & 30.0 & 41.5 & 23.3 & 25.0 & 59.5 & 40.9 & 31.0 & 23.6 & 33.9\\

\midrule
\multicolumn{14}{l}{\textit{\textcolor{gray}{Open-source MLLMs}}} \\
InternVL3-8B~\cite{wang2025internvl3} & 77.0 & 82.0 & 36.0 & 21.7 & 50.0 & 53.3 & 40.0 & 39.0 & 71.5 & 44.5 & 30.8 & 25.2 & 41.8\\
Qwen2.5-VL-7B~\cite{Qwen2.5-VL} & 68.0 & 72.0 & 27.0 & 38.3 & 45.0 & 50.1 & 33.3 & 40.0 & 77.0 & 40.6 & 40.5 & 36.2 & 44.6\\
InternVL3-78B~\cite{wang2025internvl3} & 76.0 & 81.0 & 40.0 & 45.0 & 42.0 & 56.8 & 23.3 & 49.0 & 74.0 & 42.5 & 37.4 & 30.0 & 42.7\\
\midrule
\multicolumn{14}{l}{\textit{\textcolor{gray}{LLM (text-only)}}} \\
\Lgray Llama3-8B~\cite{touvron2023llama} & 58.0 & 66.0 & 48.0 & 28.3 & 20.0 & 44.1 & 16.8 & 30.0 & 23.0 & 16.7 & 21.4 & 25.8 & 22.3 \\
\Lgray Qwen3-8B~\cite{yang2025qwen3}  & 72.0 & 77.0 & 43.0 & 35.0 & 31.0 & 51.6 & 30.5 & 28.0 & 40.5 & 30.0 & 29.0 & 25.4 & 30.6 \\
\Lgray GPT-4o~\cite{hurst2024gpt}      & 74.0 & 75.0 & 55.0 & 35.0 & 24.0 & 52.6 & 21.9 & 28.0 & 30.0 & 13.3 & 32.1 & 30.4 & 26.0 \\
\Lgray GPT-4o-mini~\cite{hurst2024gpt} & 73.0 & 79.0 & 41.0 & 30.0 & 29.0 & 50.4 & 30.5 & 32.0 & 27.5 & 16.7 & 36.1 & 26.0 & 28.1 \\
\Lgray Claude3.7-Sonet~\cite{anthropic2024claude} & 62.0 & 74.0 & 45.0 & 18.3 & 22.0 & 44.3 & 18.9 & 29.0 & 28.0 & 16.7 & 30.4 & 25.4 & 24.7 \\
\midrule
\multicolumn{14}{l}{\textit{\textcolor{gray}{VLM*(text-only)}}} \\
\Lgray GeoLLaVA-8K$^{*}$~\cite{wang2025geollava8k} & 63.0 & 71.0 & 63.0 & 25.0 & 26.0 & 49.6 & 23.5 & 24.0 & 48.0 & 16.7 & 25.1 & 35.6 & 28.8 \\
\Lgray EarthDial$^{*}$~\cite{soni2025earthdial}    & 58.0 & 74.0 & 33.0 & 55.0 & 53.0 & 54.6 & 31.1 & 37.0 & 27.0 & 23.3 & 35.6 & 24.4 & 29.7 \\
\Lgray VHM$^{*}$~\cite{pang2025vhm}                & 35.0 & 51.0 & 46.0 & 28.3 & 21.0 & 36.3 & 16.8 & 29.0 & 25.0 & 16.7 & 21.2 & 25.2 & 22.3 \\
\Lgray GeoChat* ~\cite{kuckreja2024geochat}        & 44.0 & 52.0 & 46.0 & 28.3 & 23.0 & 38.7 & 16.8 & 29.0 & 23.0 & 16.7 & 21.1 & 29.0 & 22.6 \\
\bottomrule

\end{tabular}
}
\end{table*}

\textbf{Results on LRS-VQA.}
As shown in Table~\ref{tab:lrs_vqa}, we evaluate several LLMs on LRS-VQA~\citep{luo2025lrsvqa}. In contrast, text-only models perform substantially worse: GPT-4 (text-only) achieves 19.71\%, Qwen (text-only) 16.62\%, and GPT-4o-mini (text-only) only 8.40\%. Llama~3~8B (text-only) almost completely fails, with near-zero accuracy across all categories. These results indicate that LRS-VQA cannot be solved by language priors alone and requires visual information from remote sensing imagery.

\begin{table}[t]
\footnotesize
\caption{Per-category accuracy (\%) of general-purpose VLMs on LRS-VQA. 
RU: rural/urban; ObjStatus: object status; ObjCat: object category; ObjBg: object background.}
\label{tab:lrs_vqa}
\centering
\resizebox{\linewidth}{!}{
\begin{tabular}{lccccccccc}
\toprule
\textbf{Method} & \textbf{RU} & \textbf{Count} & \textbf{ObjStatus} & \textbf{Reason.} & \textbf{ObjCat} & \textbf{ObjShape} & \textbf{ObjColor} & \textbf{ObjBg} & \textbf{Overall} \\
\midrule
GPT-4~\cite{hurst2024gpt}       & 32.11 &  0.00 & 13.70 & 12.60 & 12.77 & 44.63 & 30.98 &  8.98 & 19.71 \\
Qwen3-8B~\cite{yang2025qwen3}        & 37.94 &  0.17 &  5.10 &  9.30 & 10.23 & 37.40 & 19.61 &  6.53 & 16.62 \\
GPT-4o-mini~\cite{hurst2024gpt} & 29.15 &  0.00 &  0.40 &  3.40 &  5.78 & 15.14 &  1.31 &  4.90 &  8.40 \\
\bottomrule
\end{tabular}}
\end{table}


\textbf{Robustness to input resolution.}
Table~\ref{tab:xlsbench_resolution} reports results on XLRSBench at 2K, 4K, and 8K. 
We evaluate Qwen2.5-VL-7B, InternVL~3.5, and Phi-3.5-Vision under identical settings. 
For all three models, the accuracy curves are almost flat: changing the resolution from 8K to 4K or 2K shifts performance by at most about 2\%. 
The averaged scores over models show the same trend (38.33\% at 2K vs. 39.75\% at 8K). 
These results suggest that XLRSBench performance is mainly determined by resolution-invariant semantic cues rather than raw pixel density.

\begin{table}[t]
\footnotesize
\caption{Accuracy on XLSBench at different resolutions. Rows are models, columns are input resolutions.}
\label{tab:xlsbench_resolution}
\centering
\resizebox{0.3\textwidth}{!}{
\begin{tabular}{lcccc}
\toprule
\textbf{Model} & \textbf{2K} & \textbf{4K} & \textbf{8K} & \textbf{Avg} \\
\midrule
QwenVL2.5-7B~\cite{Qwen2.5-VL}        & 39.51 & 42.19 & 41.49 & 41.06 \\
InternVL 3.5~\citep{wang2025internvl3}  & 38.41 & 38.60 & 39.84 & 38.95 \\
Phi3.5-Vision~\cite{abdin2024phi3}    & 37.08 & 37.11 & 37.92 & 37.37 \\
\midrule
Avg                                   & 38.33 & 39.30 & 39.75 & 39.13 \\
\bottomrule
\end{tabular}}
\end{table}

\section{Case Studies and Prompt Templates}
\label{sec:case-study}

\textbf{Case studies of various tasks.}
We design a family of task-specific prompts to construct high-quality VQA data that systematically covers fine-grained perception and reasoning in remote sensing imagery. As shown in Figure~\ref{Cases from different tasks: color detection, shape/margin recognition, orientation detection, and classification.}, we begin with basic perceptual abilities, including color detection, shape/margin recognition, orientation detection, and object classification within precisely specified reference regions. Figure~\ref{Cases from different tasks: object spatial relationship, object grounding,  regional grounding, and object counting.} then targets relational and localization-centric perception, such as object spatial relationships, object grounding, regional grounding, and object counting at the scene level. Building on this, Figure~\ref{Cases from different tasks: regional counting, multi-region joint contrast (multi-image), multi-region joint contrast( single-image multi-box).} focuses on regional perception, including regional counting and multi-region joint contrast under both multi-image and single-image multi-box settings, where the model must compare multiple land parcels. Figure~\ref{anomaly_detection(single-image_single-turn)} introduces prompts for object state judgement (single-image, single- and multi-turn) and single-image anomaly detection. 
Figure~\ref{anomaly_detection} further extends anomaly reasoning to multi-turn settings and introduces double-image future prediction. 
Finally, Figure~\ref{future_prediction} presents single-image multi-turn future prediction, requiring the model to first localize key industrial complexes and then reason about their potential expansion from surrounding land-use patterns.

\textbf{Prompt design for fine-grained perception and reasoning.}
To analyze model behavior in a fine-grained manner, we design task-specific prompt templates covering all perception and reasoning types in our benchmark. For basic perception, we provide prompts for color detection, shape/margin recognition, orientation estimation, and object classification (Figures~\ref{fig:prompt_color}--\ref{fig:prompt_classification}), as well as relative spatial relationship modeling, object grounding, regional grounding, and both object-level and regional counting (Figures~\ref{fig:prompt_spatial}--\ref{fig:prompt_reg_count}). Beyond single-region perception, we introduce multi-region joint contrast prompts in both multi-image and single-image settings (Figures~\ref{fig:prompt_multi_contrast_three} and~\ref{fig:prompt_multi_contrast_single}). For higher-level state reasoning, we construct object state judgement and anomaly detection templates in both single-turn and multi-round dialogue forms (Figures~\ref{fig:prompt_state_single}, \ref{fig:prompt_state_multi}, \ref{fig:prompt_anomaly_single}, and~\ref{fig:prompt_anomaly_multi}). Finally, we design future prediction prompts for single-image multi-turn dialogues and two-image temporal comparison scenarios (Figures~\ref{fig:prompt_future_multi} and~\ref{fig:prompt_future_twoimg}). These unified prompt templates offer an intuitive view of how models handle diverse remote sensing tasks, and complement the quantitative evaluation with systematic qualitative analysis.

\textbf{Prompt templates of VQA scores.}
Figure~\ref{score_1-100} shows the scoring prompt we designed for open-ended visual question answering without answer options.
The prompt casts GPT-5-Thinking as an expert judge that assigns a score from 1–100, prioritizing correctness with respect to the reference answer while also considering usefulness, completeness, and hallucination penalties.
It enumerates lenient but explicit matching guidelines, covering semantic equivalence, numeric and range tolerances, unit handling, directional terms, and special treatment of yes/no and unanswerable cases, so that harmless paraphrases or minor deviations are not over-penalized.
We additionally provide suggested score bands (e.g., 80–100 for semantically equivalent and precise answers) to calibrate the judge and promote consistent use of the full scoring scale.
Finally, the template constrains the output to exactly two lines containing only the numeric score, which facilitates robust automatic parsing and aggregation of VQA scores.

\textbf{Prompt templates of image caption.}
Figure~\ref{image_caption} and ~\ref{image_caption_four_corner} illustrate our prompt design for GPT-5-Thinking. 
The first prompt guides the model to produce global scene descriptions for high-resolution aerial or satellite images, including an overall summary and quadrant-level subregion analysis. The second prompt turns the model into a quadrant-structured captioner for high-resolution remote-sensing images, enforcing a global summary description with explicit natural/man-made object usage breakdowns and precise spatial–functional constraints. 

{
    \small
    \bibliographystyle{ieeenat_fullname}
    \bibliography{main}
}

\setcounter{figure}{0}
\begin{figure*}[t]
    \centering

   \begin{tcolorbox}[
    width=\textwidth, 
    enhanced, 
    colframe=black, 
    boxrule=0.5mm, 
    colback=gray!20, 
    arc=3mm, 
    overlay={
        \node[anchor=north east, xshift=0mm, yshift=-6mm] (image) at (frame.north east) 
            {\includegraphics[width=6cm]{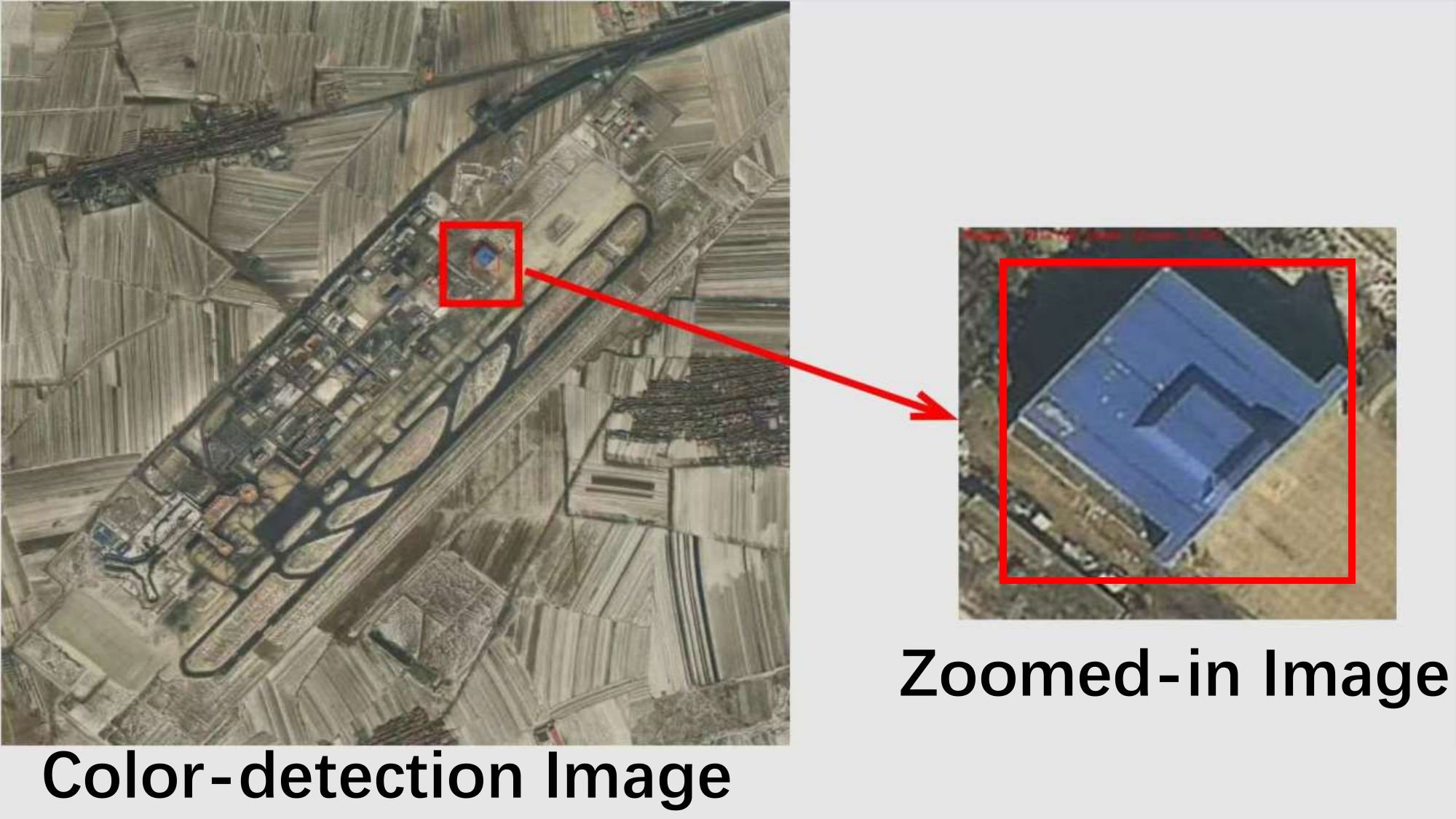}};        
    }]
    \small
    \textbf{Category: Color Detection}\\
    \textbf{ Image Resolution:} 4701 x 4946\\
    \textbf{Question:} Determine the \textcolor{red}{color} of the airport within the given reference bounding\\ box in the image.Bounding box: [2814, 1628, 2975, 1790].\\
    \textbf{Options:}
    \begin{itemize}[label={}]
        \item A: brown
        \item B: \textcolor{blue}{blue}
        \item C: white
        \item D: gray
    \end{itemize}
    \textbf{Correct Answer:} B \\
\end{tcolorbox}

\vspace{0.1em}  
    \begin{tcolorbox}[width=\textwidth, enhanced, colframe=black, boxrule=0.5mm, colback=gray!20, arc=3mm, 
         overlay={
        \node[anchor=north east, xshift=0.8mm, yshift=-6mm] (image) at (frame.north east) 
            {\includegraphics[width=6.2cm]{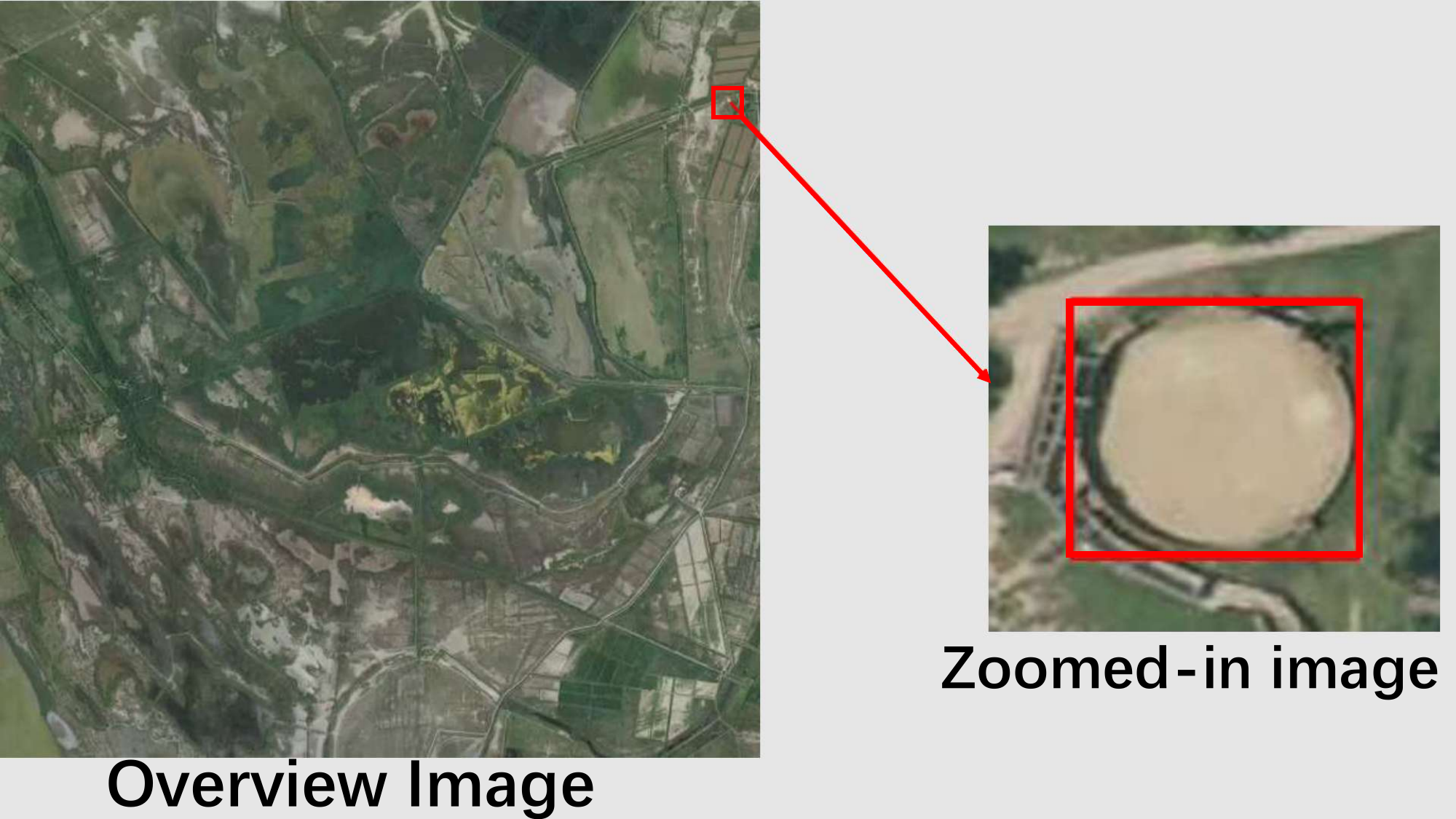}};        
    }]
        \small
        \textbf{Category: Shape/Margin Recognition}\\
        \textbf{Image resolution:} 10000 x 10000

        \textbf{Question:} Which is the \textcolor{red}{most precise} description of the main object's boundary\\ \textcolor{red}{characteristic/shape} within the given reference bounding box in the image.\\Bounding box: [9541, 1273, 9610, 1344].\\
        \textbf{Options:}
            \begin{itemize}[label={}]
                \item A: irregular
                \item B: \textcolor{blue}{circular}
                \item C: elliptical
                \item D: square
            \end{itemize}
        \textbf{Correct Answer:} B \\
    \end{tcolorbox}

    \begin{tcolorbox}[width=\textwidth, enhanced, colframe=black, boxrule=0.5mm, colback=gray!20, arc=3mm, 
        overlay={\node[anchor=north east, xshift=0.8mm, yshift=-6mm] (image) at (frame.north east)  {\includegraphics[width=6.4cm]{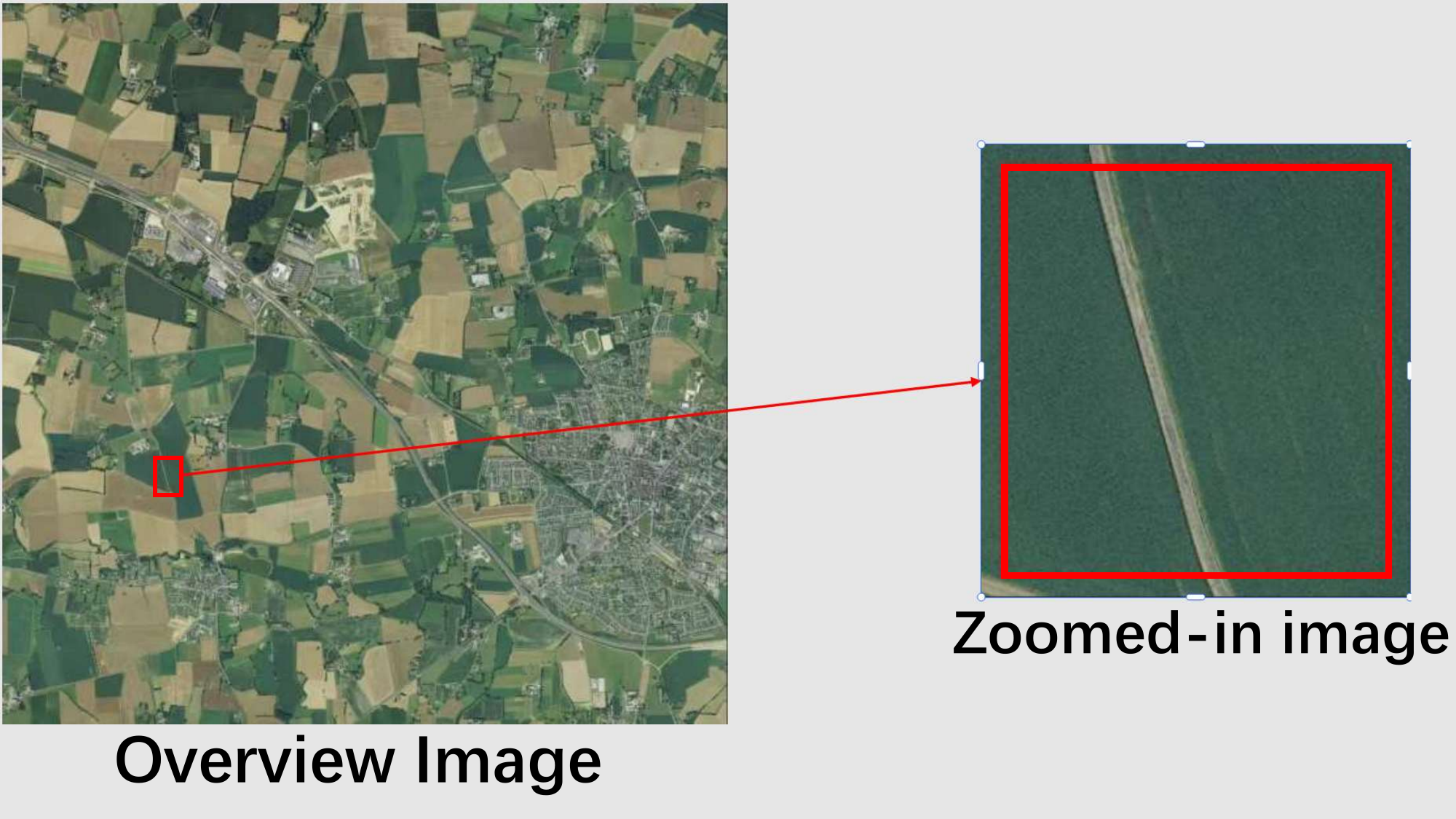}};}]
        \small
        \textbf{Category: Orientation Detection}\\
        \textbf{Image resolution:} 10000 x 10000
        
        \textbf{Question:} Determine the \textcolor{red}{orientation} of the road network within the given\\ reference bounding box in the image.Bounding box: [2096, 6332, 2454, 6732]. \\
        \textbf{Options:}
            \begin{itemize}[label={}]
                \item A: right
                \item B: top
                \item C: top-right
                \item D: \textcolor{blue}{bottom-right}
            \end{itemize}
        \textbf{Correct Answer:} D \\
    \end{tcolorbox}

    \begin{tcolorbox}[width=\textwidth, enhanced, colframe=black, boxrule=0.5mm, colback=gray!20, arc=3mm, 
        overlay={\node[anchor=north east, xshift=0.8mm, yshift=-6mm] (image) at (frame.north east)  {\includegraphics[width=6.2cm]{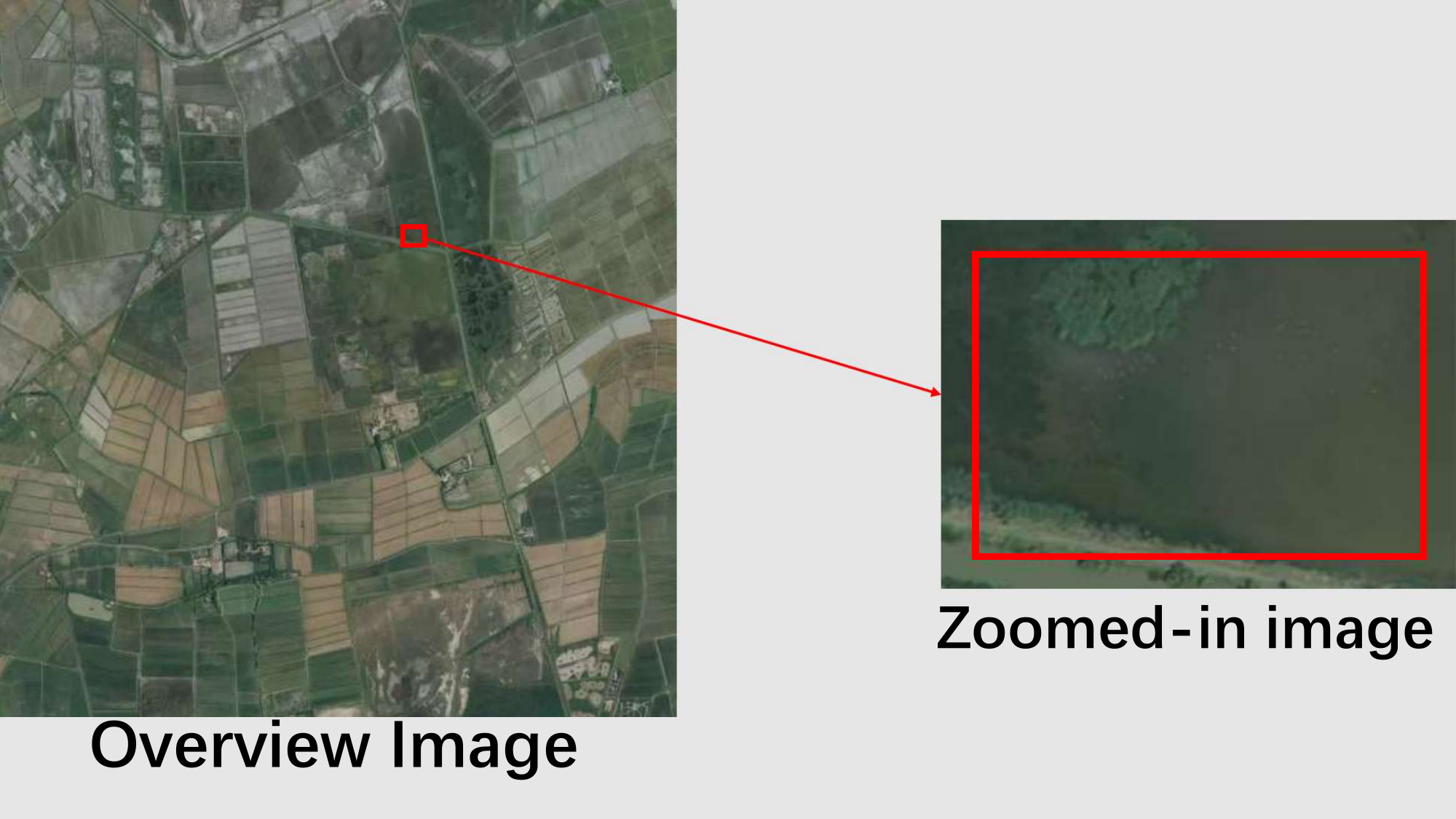}};}]
        \small
        \textbf{Category: Object Classification}\\
        \textbf{Image Resolution:} 10000 x 10000
        
        \textbf{Question:} Determine the \textcolor{red}{category} of the \textcolor{red}{main object} within the given reference\\ bounding box in the image.  Bounding box:[2382, 5403, 4489, 6853].\\
        \textbf{Options:}
        \begin{itemize}[label={}]
                \item A: coastal wetland
                \item B: \textcolor{blue}{sparse cropland area}
                \item C: urban residential zone
                \item D: dense forest
            \end{itemize}
        \textbf{Correct Answer:} B \\  
    \end{tcolorbox}
    \caption{Cases from different tasks: color detection, shape/margin recognition, orientation detection, and classification.}
    \label{Cases from different tasks: color detection, shape/margin recognition, orientation detection, and classification.}
\end{figure*}

\begin{figure*}[t]
    \centering
    \begin{tcolorbox}[width=\textwidth, enhanced, colframe=black, boxrule=0.5mm, colback=gray!20, arc=3mm, 
        overlay={\node[anchor=north east, xshift=0.8mm, yshift=-6mm] (image) at (frame.north east)  {\includegraphics[width=6cm]{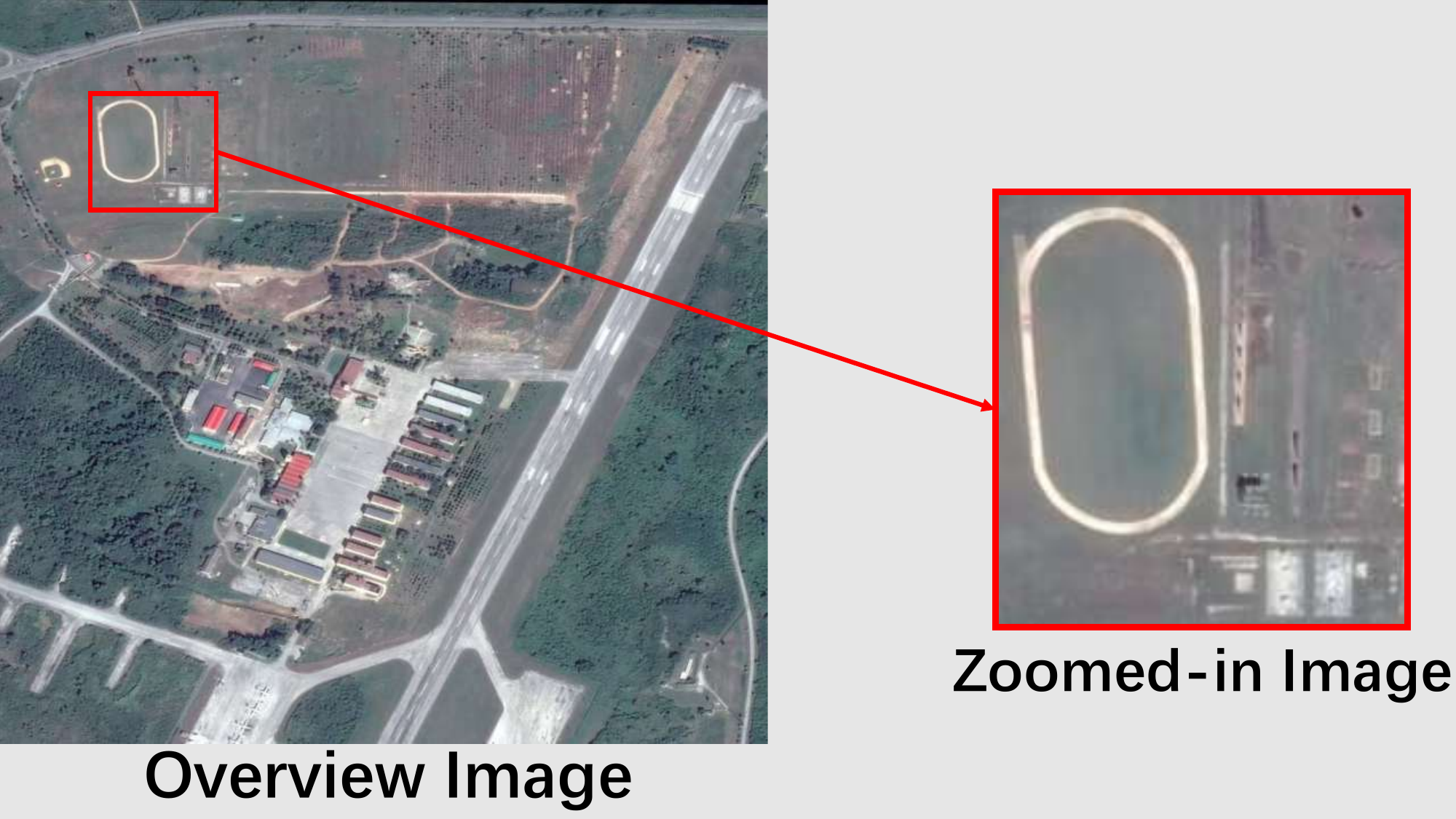}}; }]
        \small
        \textbf{Category: Object Spatial Relationship}\\
        \textbf{Image resolution:}  4183 x 8115

        \textbf{Question:} In the image, \textcolor{red}{where} is the elliptical stadium with the track located\\ \textcolor{red}{relative to} the rectangular gray building with a white roof? \\
        \textbf{Options:}
        \begin{itemize}[label={}]
            \item A: \textcolor{blue}{To the left}
            \item B: To the right
            \item C: Below
            \item D: Above
        \end{itemize}
        \textbf{Correct Answer:} A \\  
    \end{tcolorbox}

    \begin{tcolorbox}[width=\textwidth, enhanced, colframe=black, boxrule=0.5mm, colback=gray!20, arc=3mm, 
        overlay={\node[anchor=north east, xshift=-1mm, yshift=-4.5mm] (image) at (frame.north east)  {\includegraphics[width=7cm]{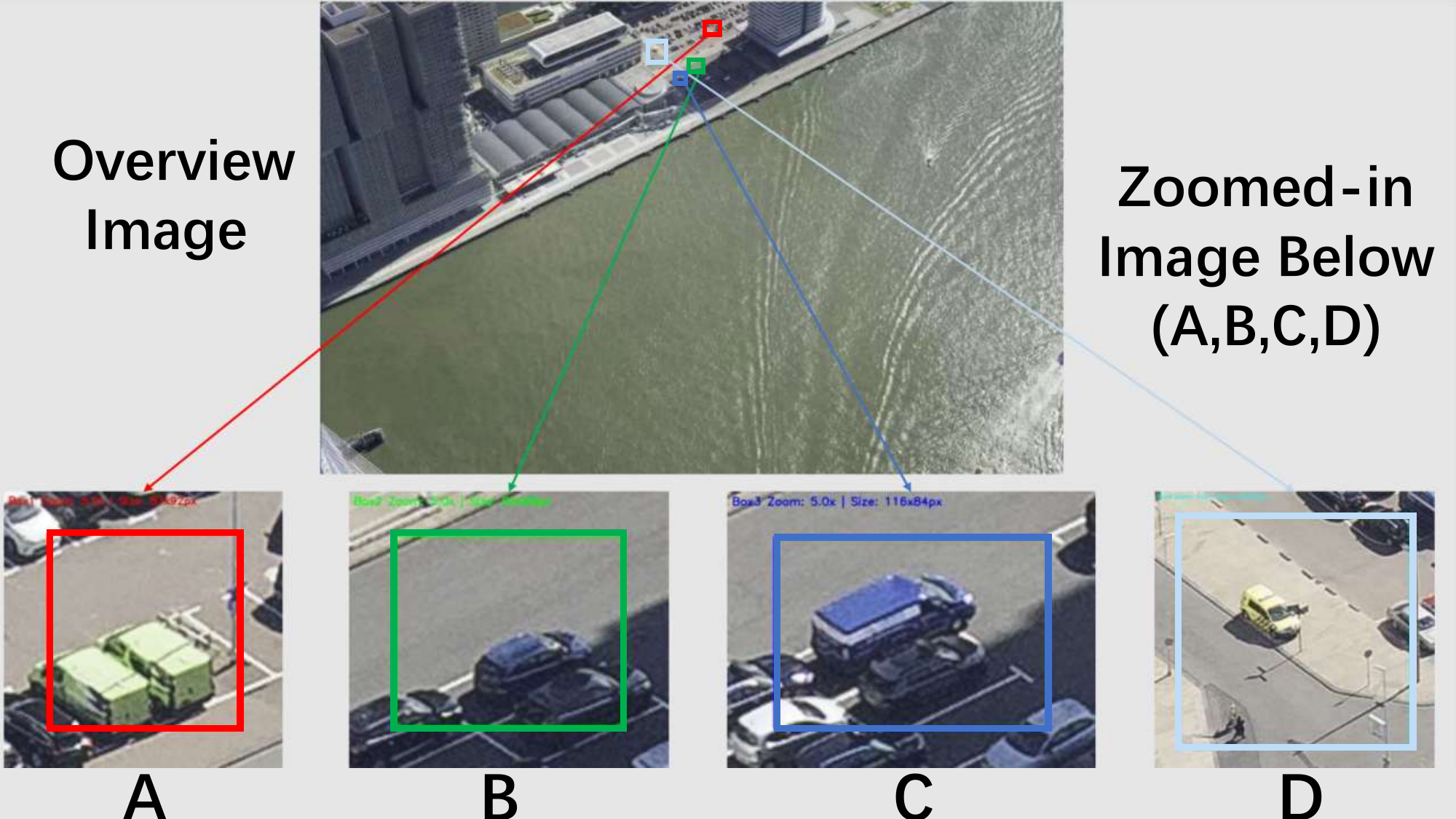}};}]
        \small
        \textbf{Category: Object Grounding}\\
        \textbf{Image resolution:} 7360 x 4912\\
        
        \textbf{Question:} Which bounding box \textcolor{red}{best localizes} a green van?\\
        \textbf{Options:}
        \begin{itemize}[label={}]
            \item A: \textcolor{blue}{[3806, 270, 3893, 362]}
            \item B: [3662, 656, 3761, 742]
            \item C: [3500, 758, 3616, 842]
            \item D: [3237, 418, 3430, 619]
        \end{itemize}
        \textbf{Correct Answer:} A \\  
    \end{tcolorbox}
\vspace{0.1em} 

     \begin{tcolorbox}[width=\textwidth, enhanced, colframe=black, boxrule=0.5mm, colback=gray!20, arc=3mm, 
        overlay={\node[anchor=north east, xshift=-5mm, yshift=-4.5mm] (image) at (frame.north east)  {\includegraphics[width=8cm]{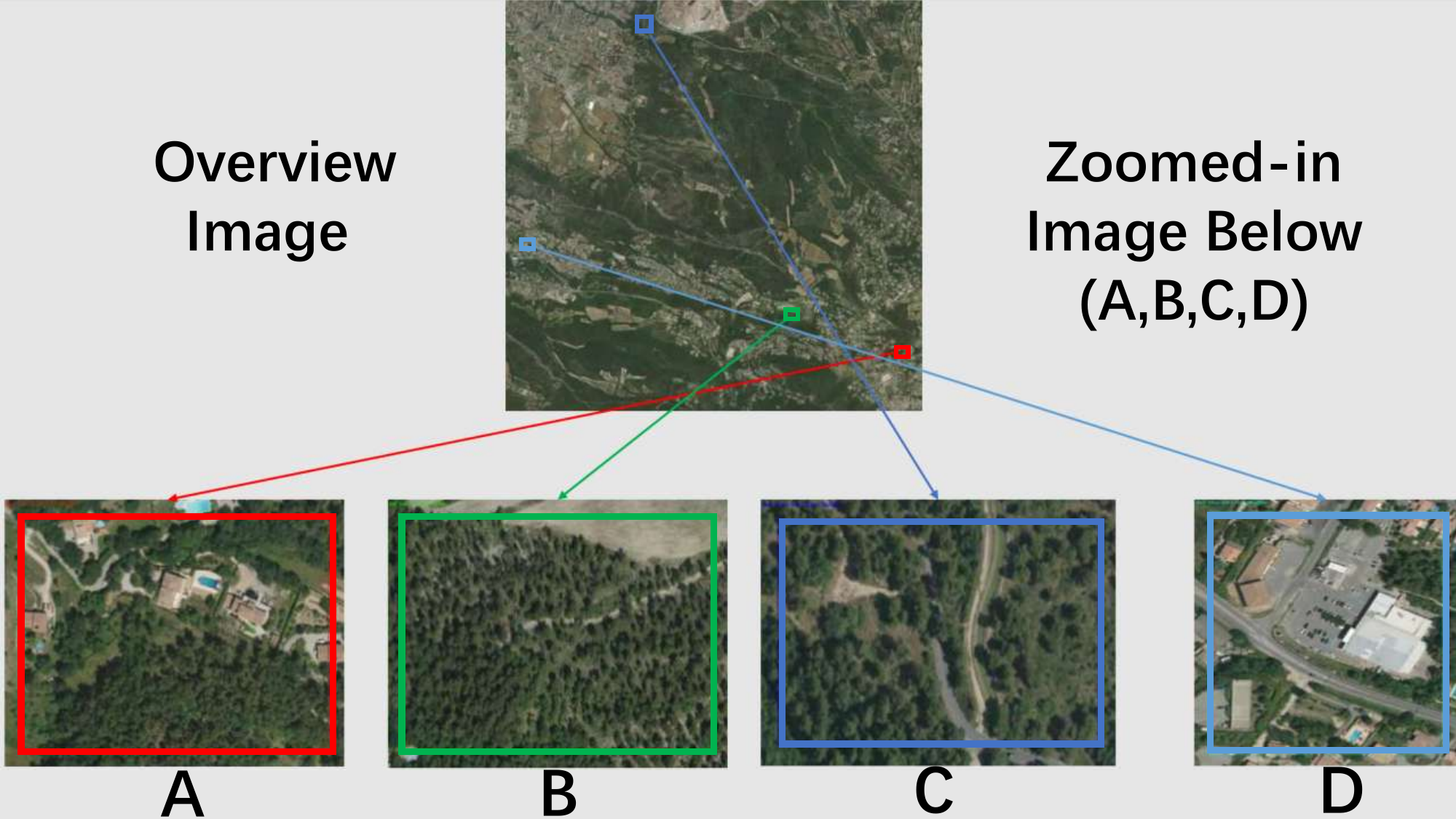}}; }]
        \small
        \textbf{Category: Regional Grounding}\\
        \textbf{Image Resolution:} 10000 x 10000\\
        
        \textbf{Question:}Which bounding box \textcolor{red}{best localizes} the area that is\\ 
        a commercial complex dominated by roads and parking lots,\\ surrounded by vegetation and residential areas?\\
        \textbf{Options:}
        \begin{itemize}[label={}]
            \item A: [9072, 8359, 9467, 8637]
            \item B: [6514, 7541, 7036, 7805]
            \item C: [3102, 436, 3480, 640]
            \item D: \textcolor{blue}{[336, 5876, 636, 6271]}
        \end{itemize}
        \textbf{Correct Answer:} D \\  
    \end{tcolorbox}
\vspace{0.1em} 

     \begin{tcolorbox}[width=\textwidth, enhanced, colframe=black, boxrule=0.5mm, colback=gray!20, arc=3mm, 
        overlay={
        \node[anchor=north east, xshift=0mm, yshift=-6mm] (image) at (frame.north east)  
        {\includegraphics[width=6cm]{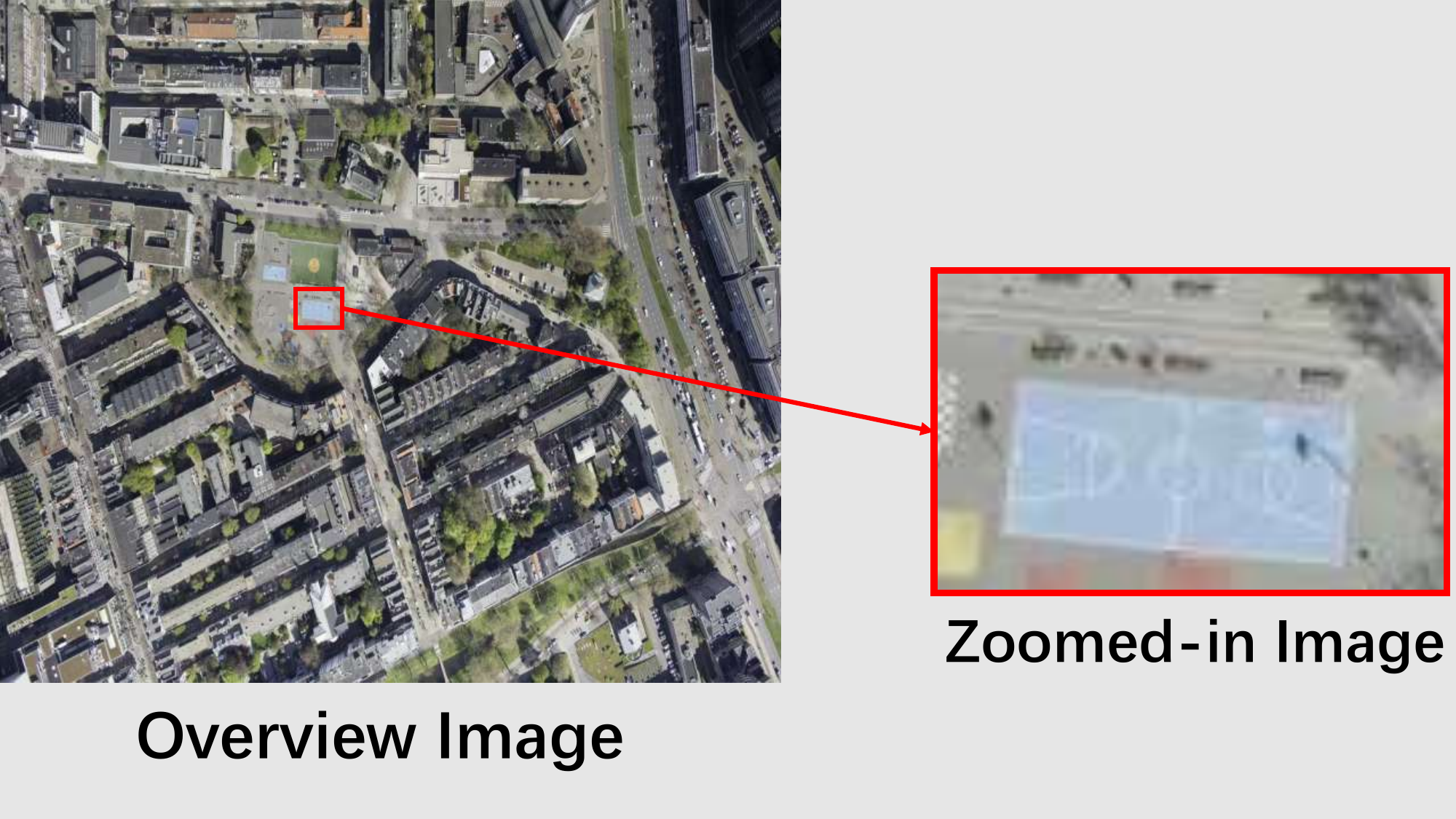}}; }]
        
        \textbf{Category: Object Counting}\\
        \textbf{Image Resolution:} 7360 x 4912
        
        \textbf{Question:}\textcolor{red}{How many} basketball courts are there \textcolor{red}{in the entire image}?\\
        \textbf{Options:}
        \begin{itemize}[label={}]
            \item A: 4
            \item B: 3
            \item C: 5
            \item D: \textcolor{blue}{1}
        \end{itemize}
        \textbf{Correct Answer:} D \\  
    \end{tcolorbox}

    \caption{Cases from different tasks: object spatial relationship, object grounding,  regional grounding, and object counting.}
    \label{Cases from different tasks: object spatial relationship, object grounding,  regional grounding, and object counting.}
\end{figure*}

\begin{figure*}[t]
    \centering
\begin{tcolorbox}[width=\textwidth, enhanced, colframe=black, boxrule=0.5mm, colback=gray!20, arc=3mm, 
        overlay={
        \node[anchor=north east, xshift=0mm, yshift=-6mm] (image) at (frame.north east)  
        {\includegraphics[width=6cm]{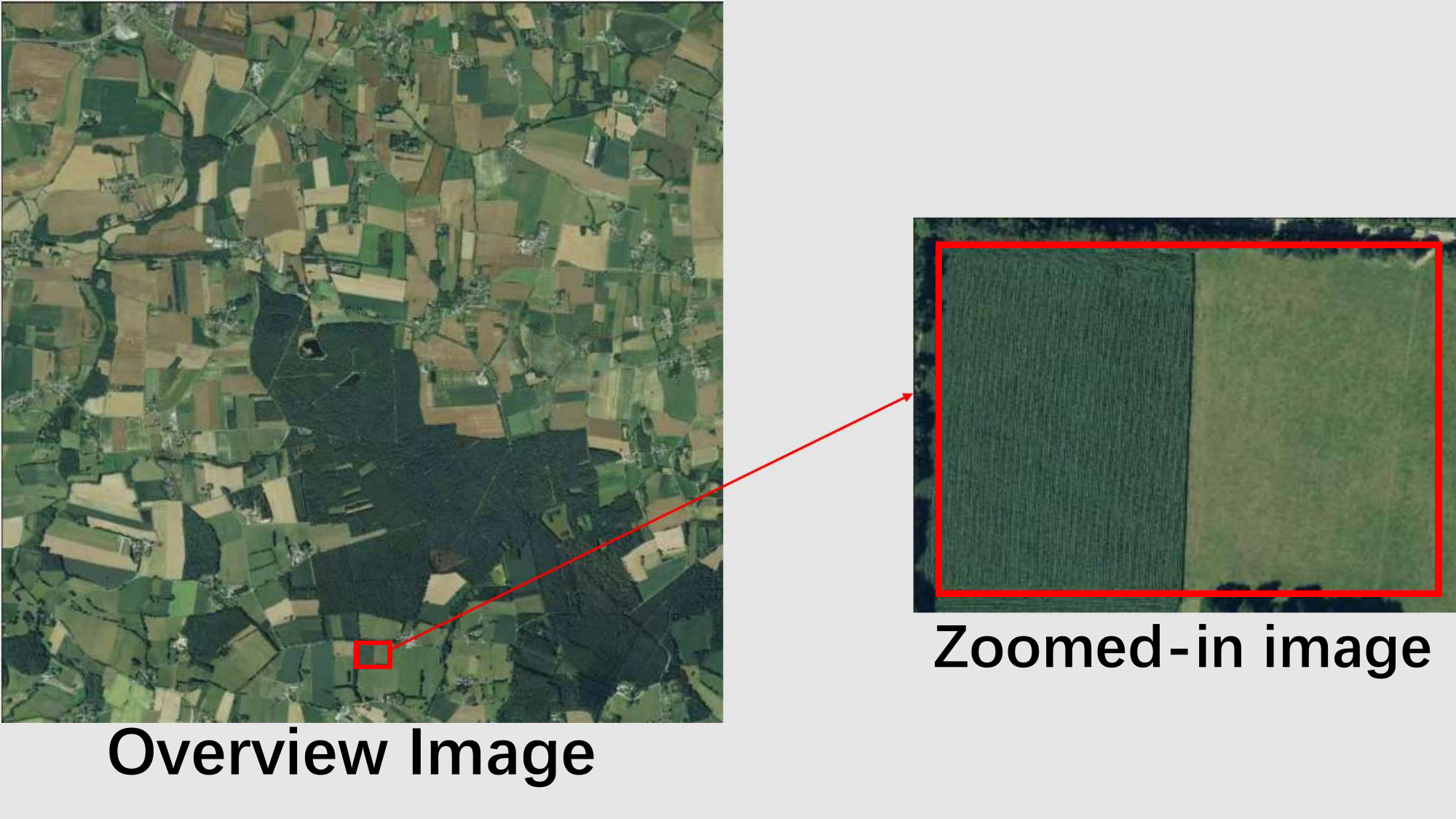}}; }]
        \small
        \textbf{Category: Regional Counting}\\
        \textbf{Image resolution:} 10000 x 10000
        \textbf{Question:}\\
        \textcolor{red}{How many} land block are there within the given reference \textcolor{red}{bounding box}\\ in the image? Bounding box: [4918, 8846, 5361, 9146].\\
        \textbf{Options:}
        \begin{itemize}[label={}]
            \item A: 1
            \item B: 6
            \item C: \textcolor{blue}{2}
            \item D: 4
        \end{itemize}
        \textbf{Correct Answer:} C \\  
    \end{tcolorbox}

\begin{tcolorbox}[width=\textwidth, enhanced, colframe=black, boxrule=0.5mm, colback=gray!20, arc=3mm, 
        overlay={\node[anchor=north east, xshift=0.8mm, yshift=-6mm] (image) at (frame.north east)  {\includegraphics[width=6.4cm]{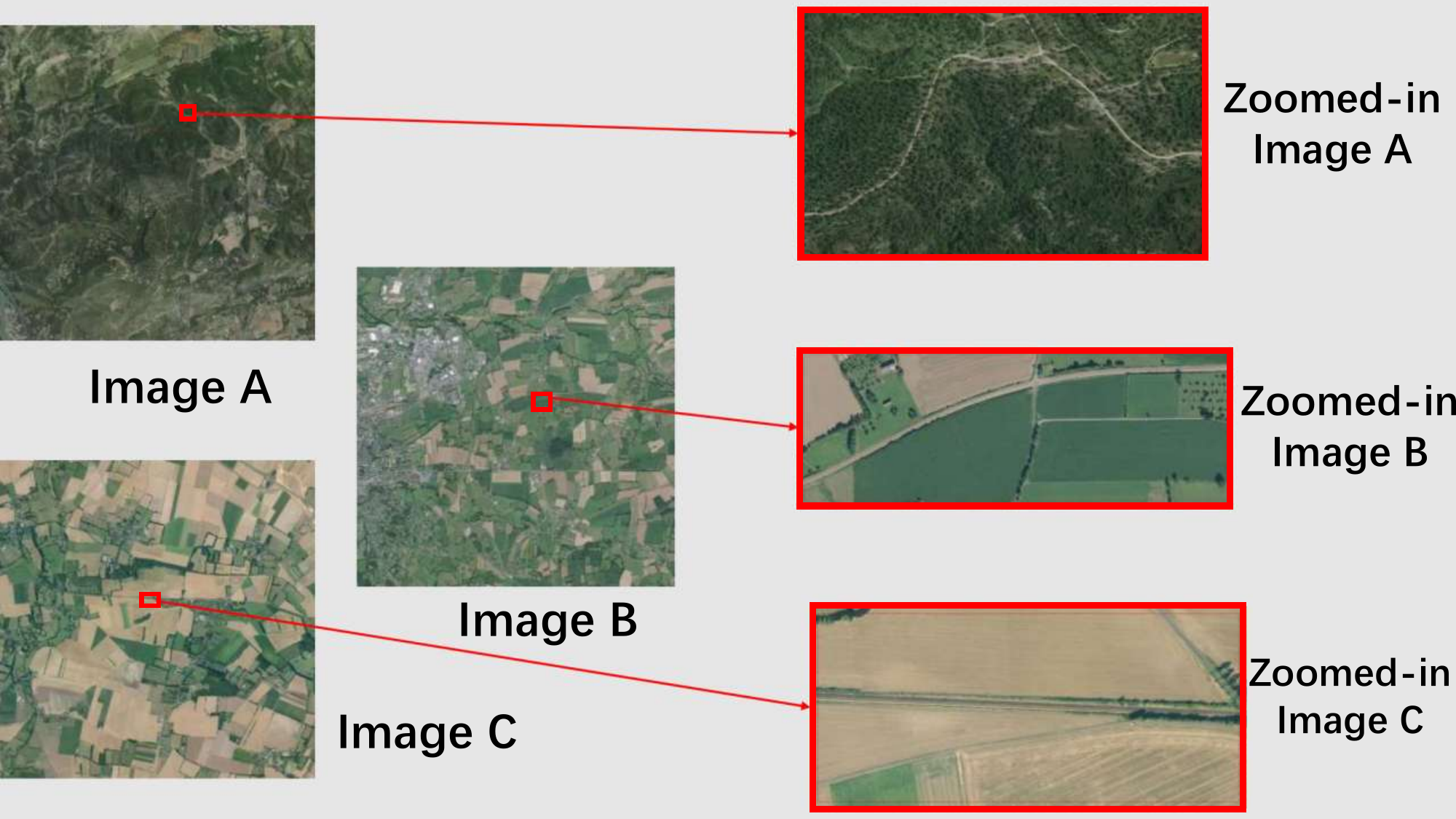}}; }]
        \small
        \textbf{Category: Multi-region Joint Contrast(Multi-image)}\\
        \textbf{Image Resolution:} 10000 x 10000
        
        \textbf{Question:}\\
        Which of the \textcolor{red}{three} given boxes in these three pictures shows the \textcolor{red}{most} winding \\road? Image A's box:[4446, 2282, 6175, 3339],Image\\ B's box:[4404, 3882, 6046, 4453],Image C's box:[3825, 4061, 4911, 4546] (The\\ Image A/B/C are determined by the order of their transmission.) resolution : \\10000 x 10000,Image A's box:[4446, 2282, 6175, 3339],Image B's box:[4404,\\ 3882, 6046, 4453],Image C's box:[3825,4061, 4911, 4546](The Image A/B/C are\\ determined by the order of their transmission.)\\
        \textbf{Options:}
        \begin{itemize}[label={}]
            \item A: Image B
            \item B: Cannot compare
            \item C: \textcolor{blue}{Image A}
            \item D: Image C
        \end{itemize}
        \textbf{Correct Answer:} C \\  
    \end{tcolorbox}

    \begin{tcolorbox}[width=\textwidth,height=5.4cm, enhanced, colframe=black, boxrule=0.5mm, colback=gray!20, arc=3mm, 
        overlay={\node[anchor=north east, xshift=-5mm, yshift=-4.5mm] (image) at (frame.north east)  {\includegraphics[width=8cm]{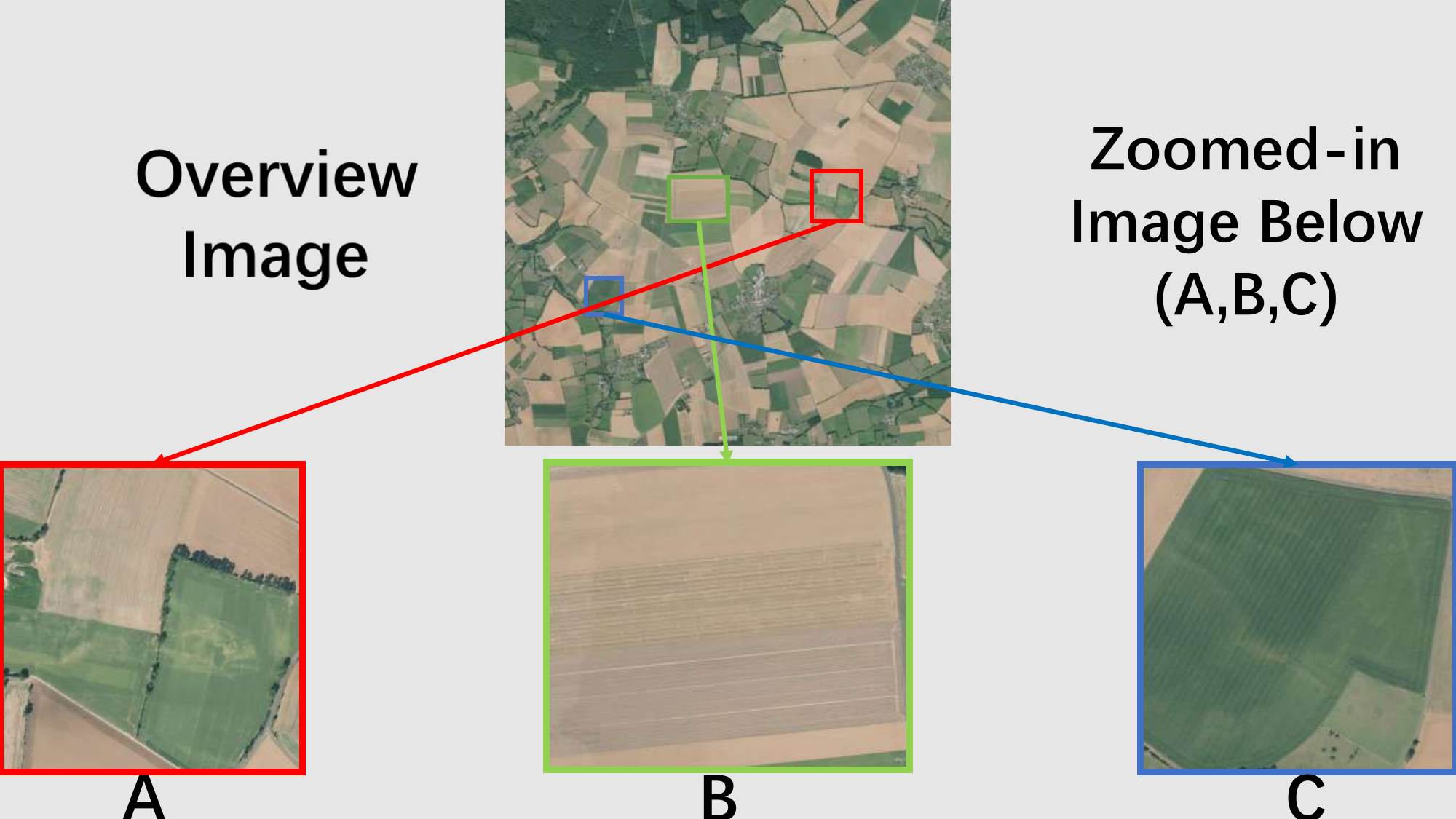}}; }]
        \small
        \textbf{Category: Multi-region Joint Contrast( Single-image Multi-box )}\\
        \textbf{Image Resolution:} 10000 x 10000 \\
        \textbf{Question:}\\
        Which bounding box \textcolor{red}{best} represents the cropland patch with the\\ \textcolor{red}{highest} greenness level?\\
        \textbf{Options:}
        \begin{itemize}[label={}]
            \item A: \textcolor{blue}{[5054, 1461, 6196, 2453]}
            \item B: [7625, 5725, 8968, 6982]
            \item C: [4175, 7411, 5654, 8632]
        \end{itemize}
        \textbf{Correct Answer:} A \\  
    \end{tcolorbox}
    \caption{Cases from different tasks: regional counting, multi-region joint contrast (multi-image), multi-region joint contrast( single-image multi-box).}
    \label{Cases from different tasks: regional counting, multi-region joint contrast (multi-image), multi-region joint contrast( single-image multi-box).}
\end{figure*}

\begin{figure*}[t]
    \centering
    \begin{tcolorbox}[width=\textwidth, enhanced, colframe=black, boxrule=0.5mm, colback=gray!20, arc=3mm, 
        overlay={
        \node[anchor=north east, xshift=0.8mm, yshift=-6mm] (image) at (frame.north east)   
        {\includegraphics[width=6cm]{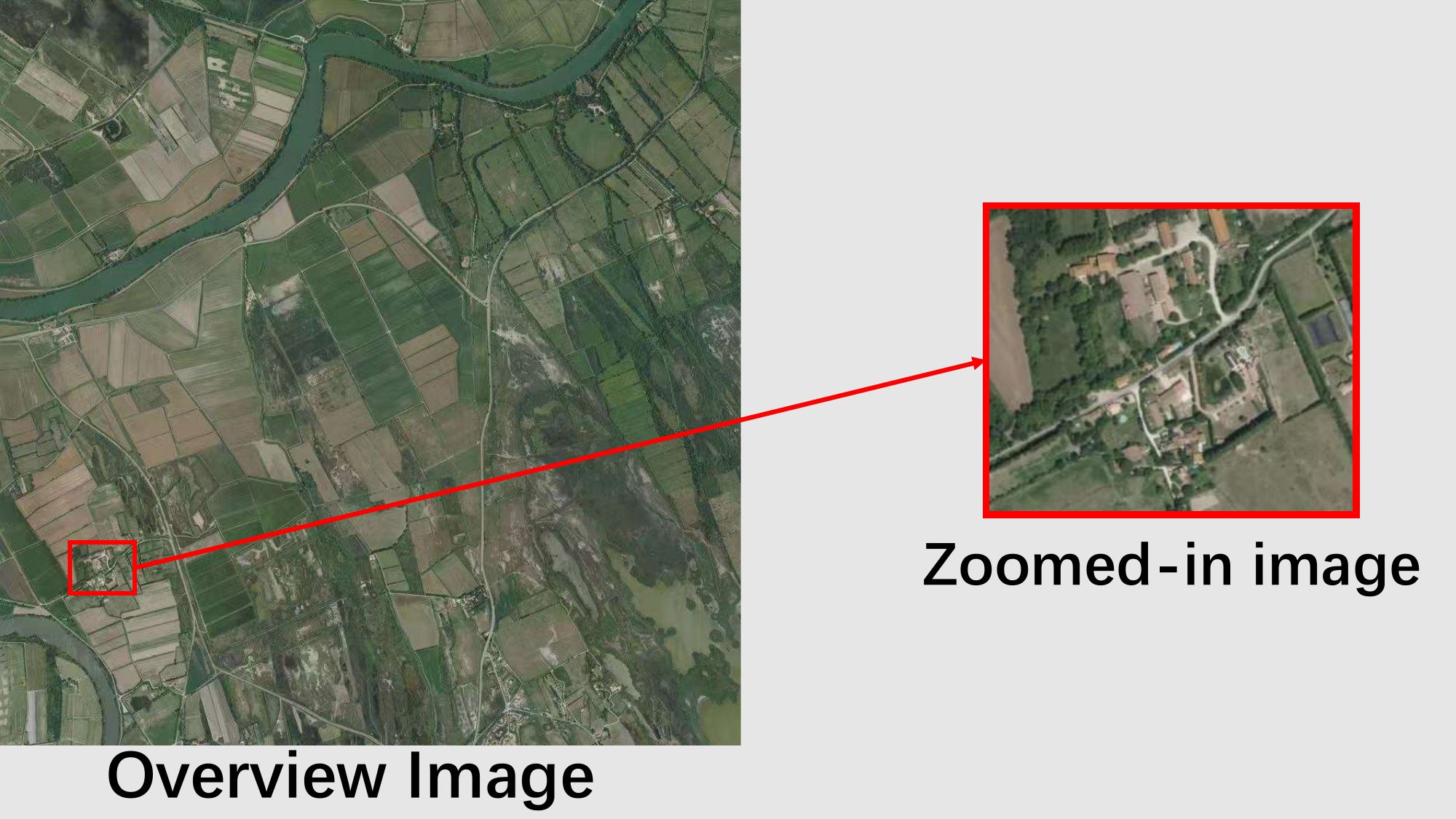}}; }]
        \small
        \textbf{Category: Object State Judgement(Single-image Multi-turn)}\\
        \textbf{Image Resolution:} 10000 x 10000\\
        \textbf{Question1:}\\
        What can be observed about the building in the bottom-left part of the image,\\\textcolor{red}{which} is surrounded by a grid of roads and fields?\\
        \textbf{Options:}
        \begin{itemize}[label={}]
            \item A: The roof is covered in a layer of snow, suggesting winter conditions.
            \item B: \textcolor{blue}{The roof is in good condition, with no visible signs of damage.}
            \item C: There are large cracks in the walls, indicating structural damage.
            \item D: The area around the building is flooded, with standing water visible.
        \end{itemize}
        \textbf{Correct Answer:} B \\ 
        \textbf{Question2:}\\
        How does the \textcolor{red}{condition} of the building in the bottom-left of the image \textcolor{red}{affect its operational status}?\\
        \textbf{Options:}
        \begin{itemize}[label={}]
            \item A: The building is likely under construction due to visible scaffolding.
            \item B: \textcolor{blue}{The building appears fully operational, without any visible signs of damage.}
            \item C: The building is uninhabitable due to visible cracks in the structure.
            \item D: The building is flooded, making it non-functional for use.
        \end{itemize}
        \textbf{Correct Answer:} B \\ 
        \textbf{Question3:}\\
        What is the \textcolor{red}{current state} of the building in the \textcolor{red}{bottom-left} part of the image?\\
        \textbf{Options:}
        \begin{itemize}[label={}]
            \item A: \textcolor{blue}{Intact}
            \item B: Damaged
            \item C: Under construction
            \item D: Flooded
        \end{itemize}
        \textbf{Correct Answer:} A 
        
    \end{tcolorbox}

    \begin{tcolorbox}[width=\textwidth, enhanced, colframe=black, boxrule=0.5mm, colback=gray!20, arc=3mm, 
       overlay={
        \node[anchor=north east, xshift=0.8mm, yshift=-6mm] (image) at (frame.north east)   
        {\includegraphics[width=6cm]{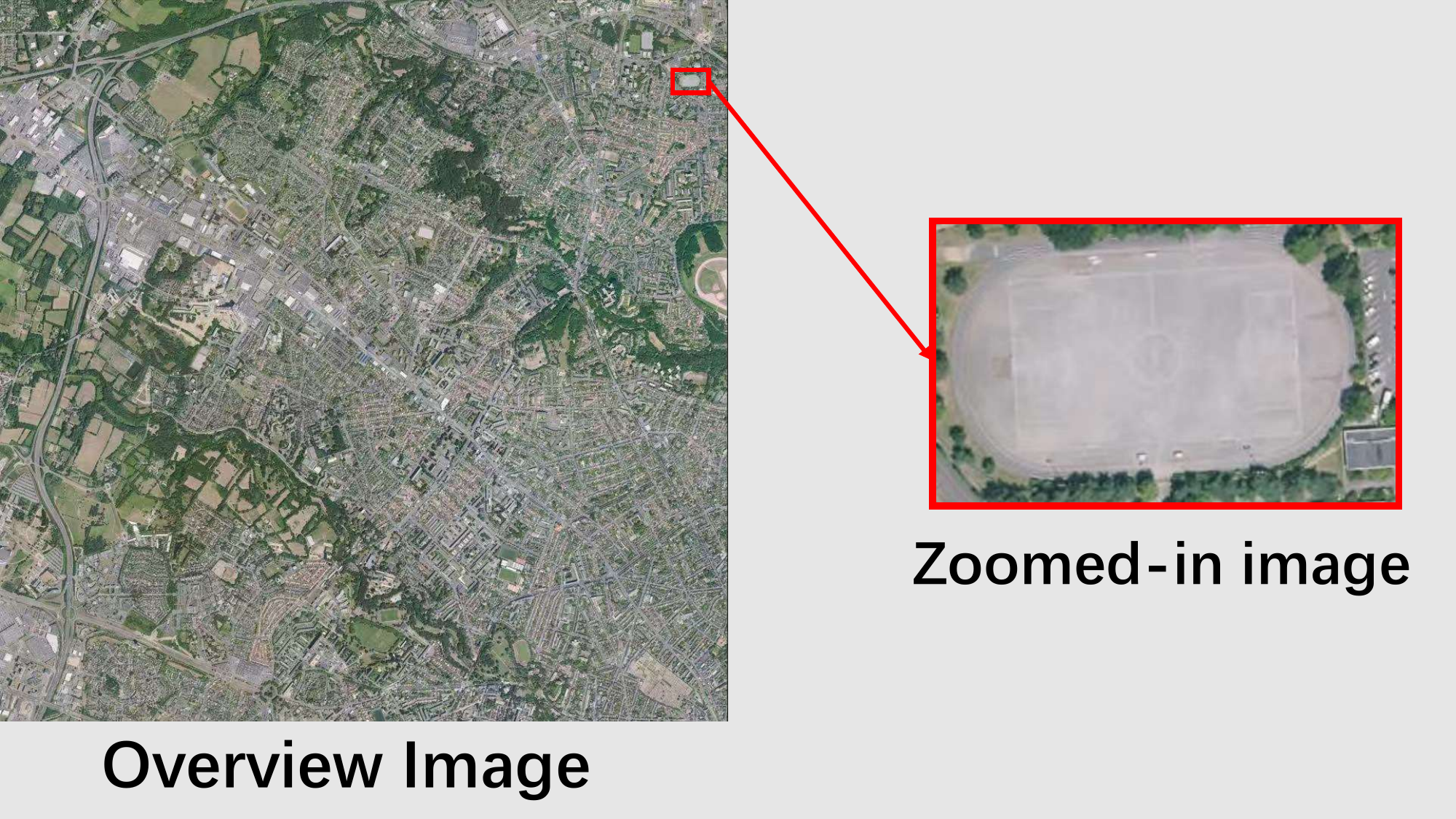}}; }]
        \small
        \textbf{Category: Object State Judgement(Single-image Single-turn)}\\
        \textbf{Image Resolution:} 10000 x 10000
        \textbf{Question:}\\
        What is the \textcolor{red}{current state} of the large rectangular sports field with a dirt track\\ and green infield located in the \textcolor{red}{upper right corner} of the image?\\
        \textbf{Options:}  
        \begin{itemize}[label={}]
            \item A: Under construction
            \item B: Damaged
            \item C: Not in use / Inactive
            \item D: \textcolor{blue}{In use / Active}
        \end{itemize}
        \textbf{Correct Answer:} D  
    \end{tcolorbox}

\begin{tcolorbox}[width=\textwidth, enhanced, colframe=black, boxrule=0.5mm, colback=gray!20, arc=3mm, 
        overlay={
        \node[anchor=north east, xshift=0.8mm, yshift=-6mm] (image) at (frame.north east)   
        {\includegraphics[width=6cm]{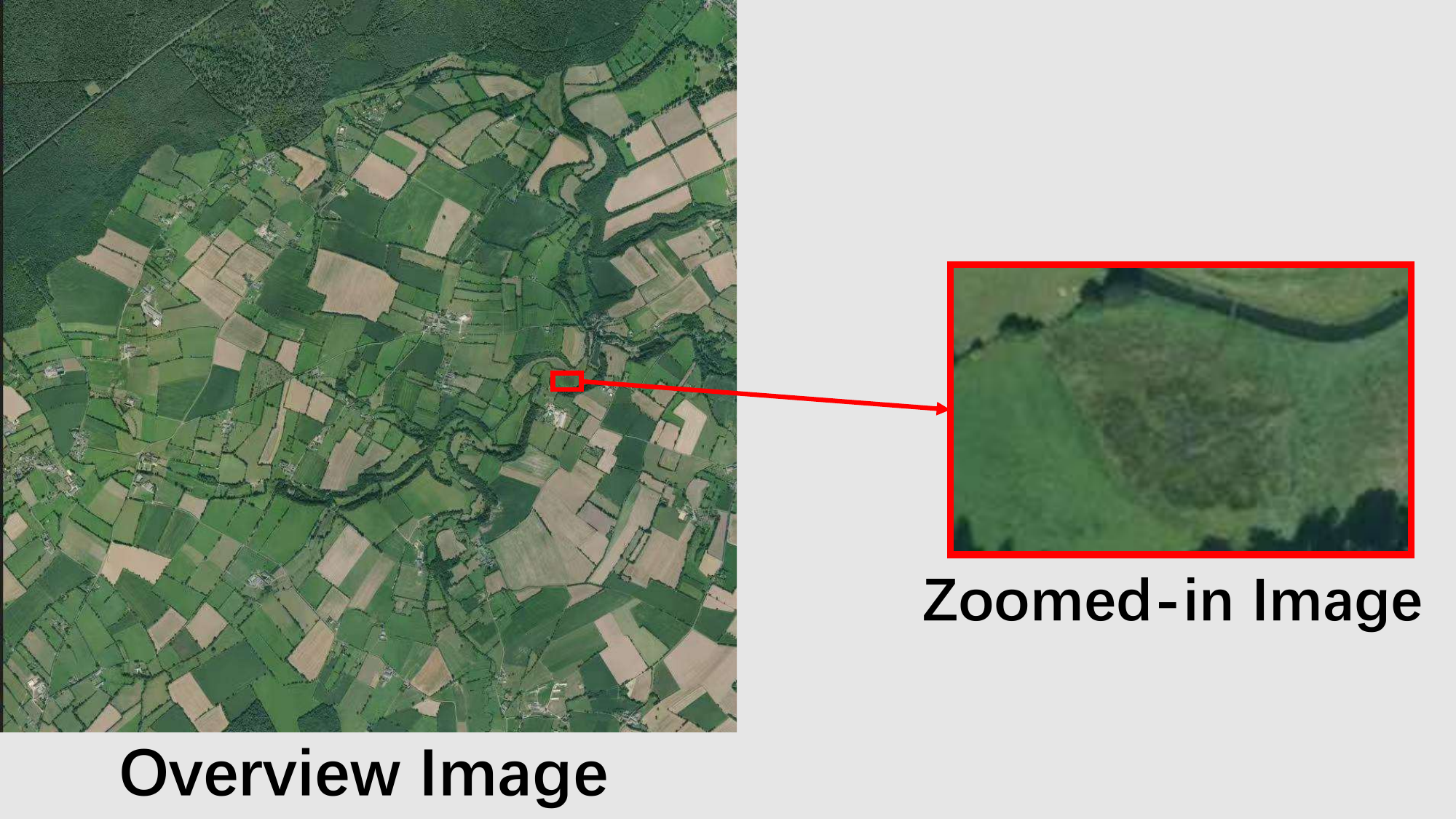}}; }]
        \small
        \textbf{Category: Anomaly Detection(Single-image Single-turn)}\\
        \textbf{Image Resolution: }10000 x 10000
        
        \textbf{Question:}\\
        The large rectangular field with a darker shade of green in the middle-right\\ section of the image. What is the \textcolor{red}{most likely reason} for this field's \textcolor{red}{darker}\\ shade compared to surrounding fields?\\
        \textbf{Options:}    
        \begin{itemize}[label={}]
            \item A: The field has been recently fertilized, causing the darker shade.
            \item B: \textcolor{blue}{The field is experiencing waterlogging, leading to the darker shade.}
            \item C: The field is undergoing crop rotation, resulting in the darker shade.
            \item D: The field is being used for grazing animals, which causes the darker shade.
        \end{itemize}
        \textbf{Correct Answer:} B 
    \end{tcolorbox}
    \caption{Cases from different tasks: object state judgement (single-image multi-turn), object state judgement (single-image single-turn), anomaly detection(single-image single-turn).}
    \label{anomaly_detection(single-image_single-turn)}
\end{figure*}

\begin{figure*}[t]
\centering
\begin{tcolorbox}[width=\textwidth, enhanced, colframe=black, boxrule=0.5mm, colback=gray!20, arc=3mm, 
        overlay={
        \node[anchor=north east, xshift=0.8mm, yshift=-6mm] (image) at (frame.north east)   
        {\includegraphics[width=6cm]{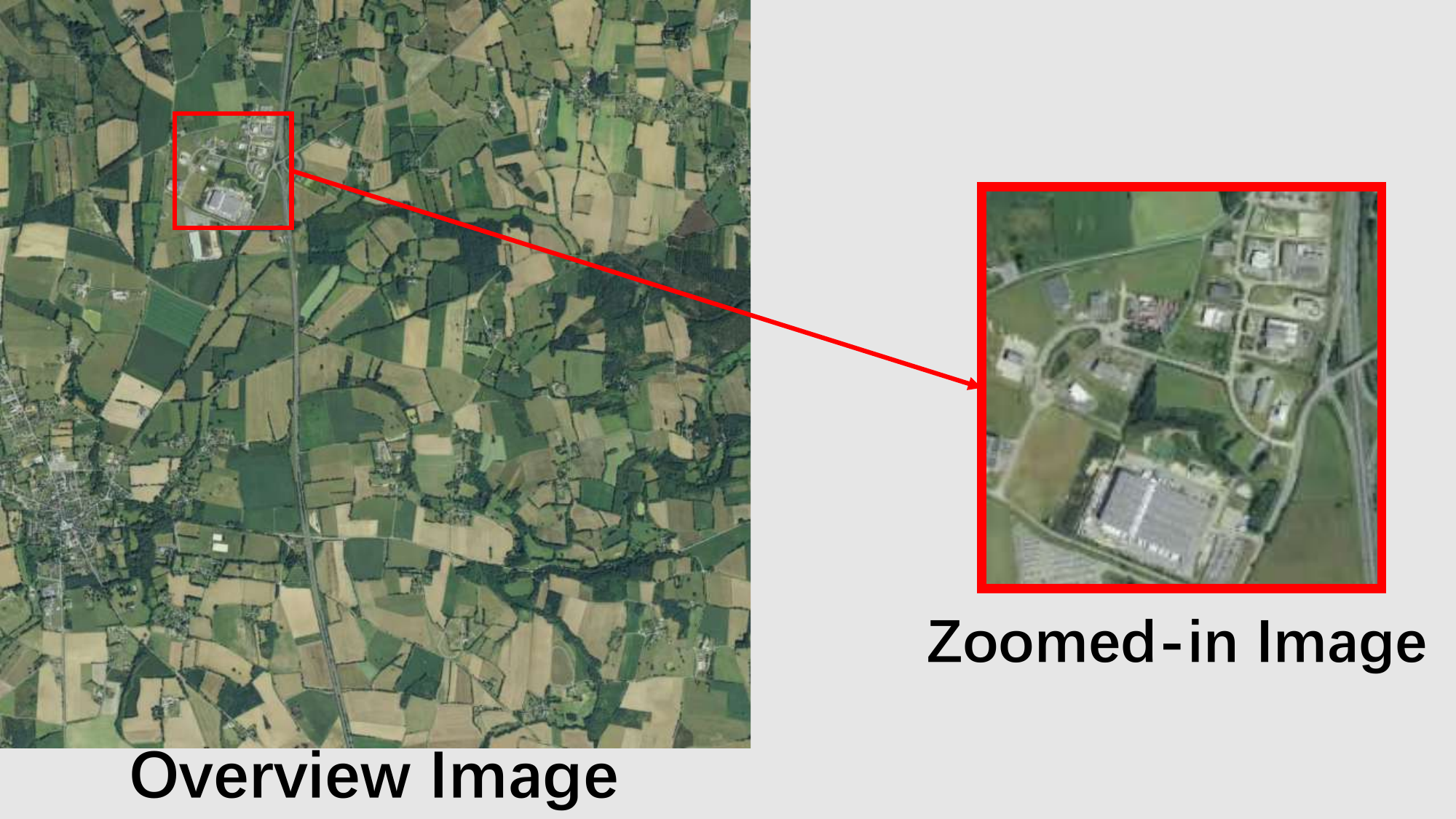}}; }]
        \small
        \textbf{Category: Anomaly Detection(Single-image Multi-turn)}\\
        \textbf{Image Resolution:} 10000 x 10000\\
        \textbf{Question1:}\\
       \textcolor{red}{Where} is the rectangular light-gray industrial complex \textcolor{red}{with} several clustered\\ large buildings and parking lots, \textcolor{red}{surrounded by} open green and yellow fields,\\ and \textcolor{red}{adjacent to} a straight highway running vertically, located in the image? 
        \textbf{Options:}
        \begin{itemize}[label={}]
            \item A: The top-right corner of the image. B: The middle of the image.
            \item C: \textcolor{blue}{The top-left corner of the image.  D: The bottom-left corner of the image.}
        \end{itemize}
        \textbf{Correct Answer:} C \\  
        \textbf{Question2:}\\
       What \textcolor{red}{relationship or change} might \textcolor{red}{affect} the selected object/region in the near\\ \textcolor{red}{future}, based on the image's visual cues?\\
        \textbf{Options:}
        \begin{itemize}[label={}]
            \item A: \textcolor{blue}{Expansion of the industrial complex into nearby agricultural fields due to\\ available space.}
            \item B: Reduction in industrial activity caused by flooding risk.
            \item C: Conversion of the area into residential housing developments.
            \item D: Abandonment of the complex leading to overgrown vegetation.
        \end{itemize}
        \textbf{Correct Answer:} A \\  
        \textbf{Question3:}\\
       \textcolor{red}{If} expansion of the industrial complex into nearby agricultural fields due to \\available space \textcolor{red}{happens}, what is the most likely \textcolor{red}{future state} of the selected\\ object/region?\\
        \textbf{Options:}
        \begin{itemize}[label={}]
            \item A: The farmland surrounding it will become denser with forest growth.
            \item B: The nearby highway will be removed, reducing accessibility.
            \item C: \textcolor{blue}{The complex will occupy a larger area, replacing some farmland with \\paved infrastructure.}
            \item D: The complex will shrink as industrial demand decreases.
        \end{itemize}
        \textbf{Correct Answer:} C \\  
    \end{tcolorbox}

   \begin{tcolorbox}[
    width=\textwidth, 
    enhanced, 
    colframe=black, 
    boxrule=0.5mm, 
    colback=gray!20, 
    arc=3mm,
    overlay={
        \node[anchor=north east, xshift=-5.2mm, yshift=-2mm] at (frame.north east) 
            {\includegraphics[width=6cm]{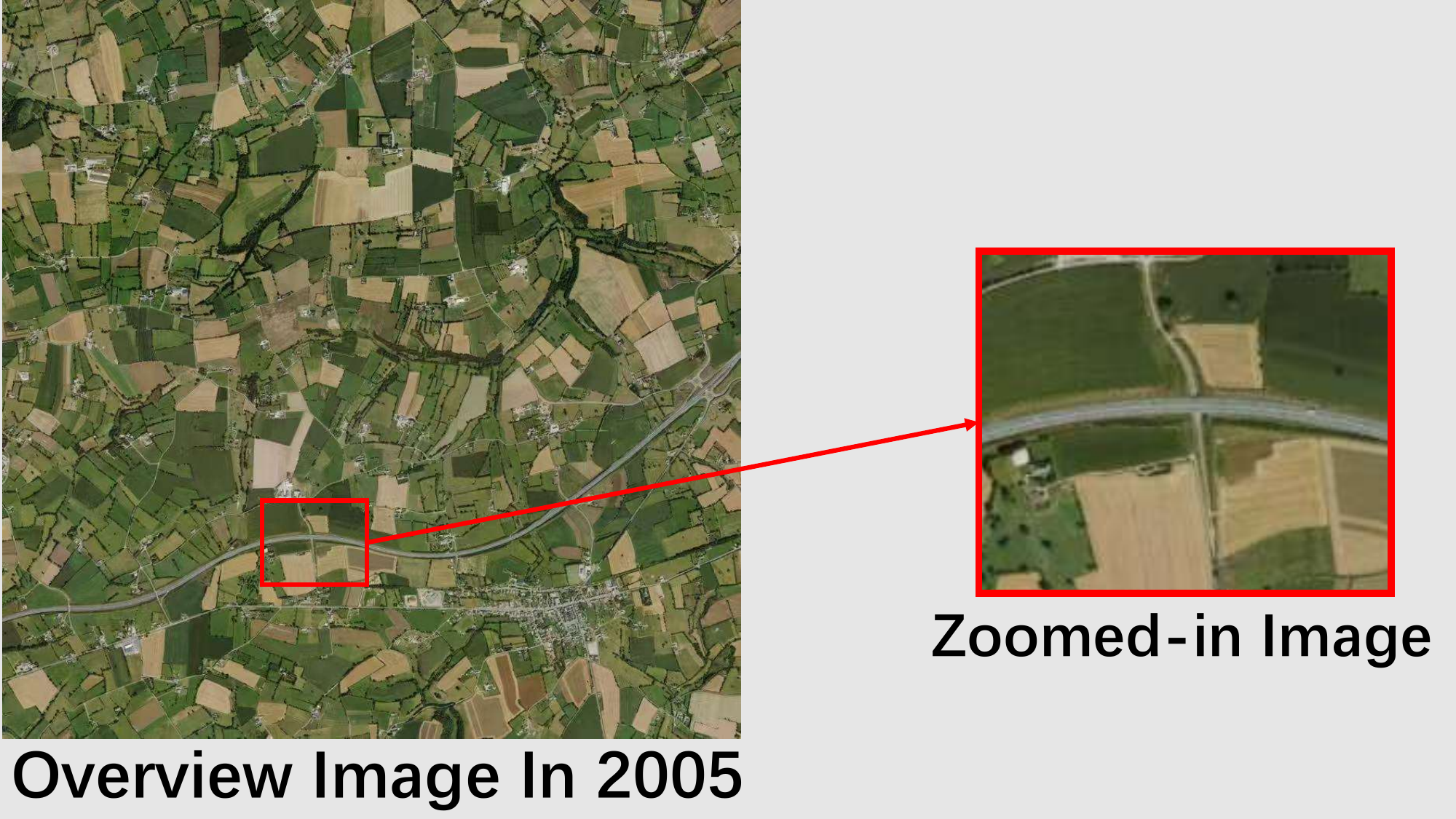}};
        \node[anchor=north east, xshift=-5.2mm, yshift=-36mm] at (frame.north east) 
            {\includegraphics[width=6cm]{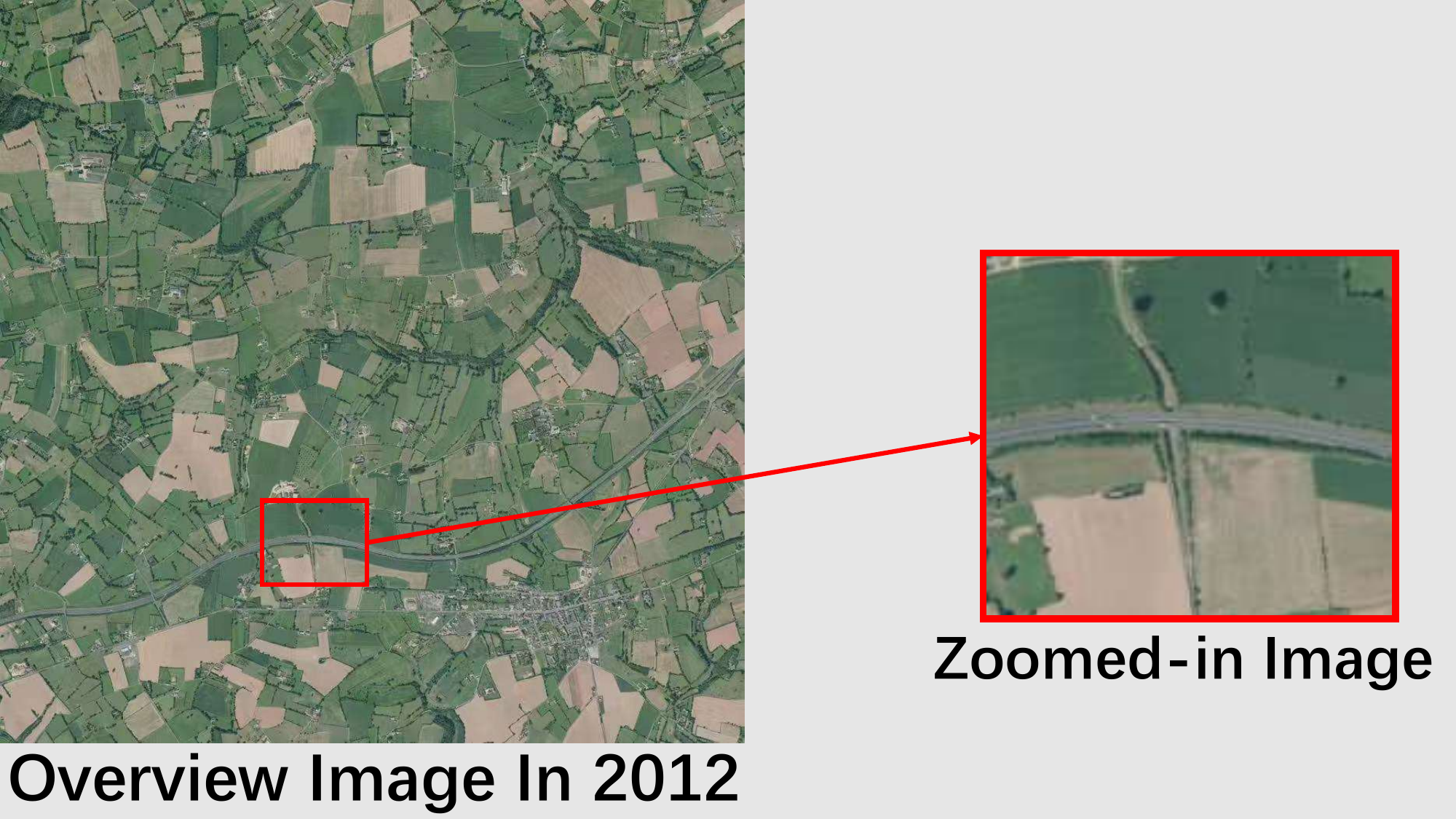}};
    }
]
\small
\textbf{Category: Future Prediction (Double-image Single-turn)}\\
\textbf{Image Resolution:} 10000 x 10000\\
\textbf{Question:}\\
\textcolor{red}{Comparing} the agricultural fields located near the central winding road in \\\textcolor{red}{Image A and Image B}, what is the most plausible \textcolor{red}{future change} that might\\ occur in this area based on the observed visual patterns?\\
\textbf{Options:}
\begin{itemize}[label={}]
    \item A: The fields are likely to revert to natural vegetation, as signs of crop\\ reduction and overgrowth can be seen between Image A and Image B.
    \item B: The fields are likely to be converted into residential zones, as there is a\\ visible increase in buildings along the road from Image A to Image B.
    \item C: The fields are likely to remain unchanged, as there is no noticeable\\ alteration in their structure or use between Image A and Image B.
    \item D: \textcolor{blue}{The agricultural fields near the central winding road are likely to\\ consolidate into larger plots, as several smaller fields appear to have\\ merged between Image A and Image B.}
\end{itemize}
\textbf{Correct Answer:} D
\end{tcolorbox}
\caption{Cases from different tasks: anomaly detection (single-image multi-turn) and future prediction (double-image single-turn).}
\label{anomaly_detection}
\end{figure*}

\begin{figure*}
    \centering
    \begin{tcolorbox}[width=\textwidth, enhanced, colframe=black, boxrule=0.5mm, colback=gray!20, arc=3mm, 
        overlay={
        \node[anchor=north east, xshift=0.8mm, yshift=-6mm] (image) at (frame.north east)   
        {\includegraphics[width=6cm]{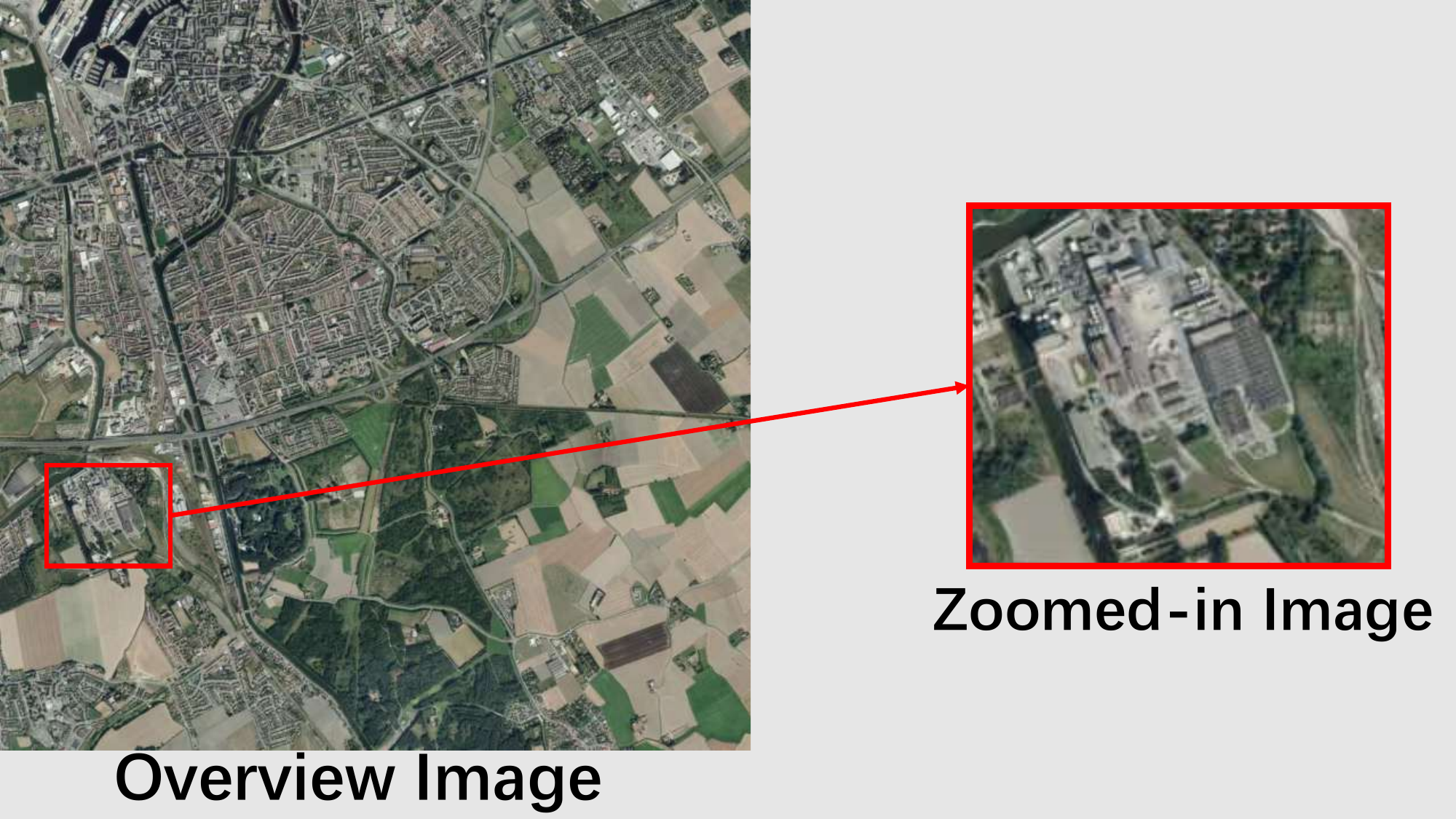}}; }]
        \small
        \textbf{Category: Future Prediction(Single-image Multi-turn)}\\
        \textbf{Image Resolution:} 10000 x 10000
        
        \textbf{Question1:}\\
        \textcolor{red}{Where} is the large industrial complex located in the image?\\
        \textbf{Options:}
        \begin{itemize}[label={}]
            \item A: The top-right corner of the image.
            \item B: The left side of the image.
            \item C: \textcolor{blue}{The bottom-left corner of the image.}
            \item D: The right side of the image.
        \end{itemize}
        \textbf{Correct Answer:} C 
        
        \textbf{Question2:}\\
        What \textcolor{red}{relationship or change} might \textcolor{red}{affect} the selected industrial complex \textcolor{red}{in the near future}, based on the image's visual cues?\\
        \textbf{Options:}
        \begin{itemize}[label={}]
            \item A: The proximity to major highways, which might increase transportation access to the complex.
            \item B: he forested area to the south, which may be at risk of being cleared for industrial development.
            \item C:The large residential areas surrounding it, which may increase demand for industrial services.
            \item D: \textcolor{blue}{The presence of nearby agricultural land and possible expansion into it.}
        \end{itemize}
        \textbf{Correct Answer:} D 

        \textbf{Question3:}\\
        \textcolor{red}{If }the agricultural land \textcolor{red}{continues to} be developed into industrial areas, what is the most likely \textcolor{red}{future state} of the selected industrial complex?\\
        \textbf{Options:}
        \begin{itemize}[label={}]
            \item A: TThe industrial complex will experience increased traffic but minimal expansion due to space constraints.
            \item B: TThe industrial complex will be relocated further from residential areas due to zoning regulations.
            \item C: \textcolor{blue}{The industrial complex will expand, with more buildings and infrastructure replacing farmland.}
            \item D: The industrial complex will remain the same, with no significant changes.
        \end{itemize}
        \textbf{Correct Answer:} C  
    \end{tcolorbox}
\caption{Cases from different tasks: future prediction (single-image multi-turn).}
\label{future_prediction}
\end{figure*}

\begin{figure*}[t]
    \centering

   \begin{tcolorbox}[
    width=\textwidth,
    enhanced,
    colframe=black,
    boxrule=0.5mm,
    colback=gray!20,
    arc=3mm
]
    \small
    \textbf{Category: Color Detection}\\
    \textbf{Model: Qwen2.5-VL-7B} \\
    \textbf{Prompt:} You are a dataset constructor for Remote Sensing VQA (Color Detection).

Image metadata:

- \textcolor{blue}{Image file}: \{image\_path\}
- \textcolor{blue}{Resolution}: \{width\} x \{height\}
- \textcolor{blue}{Reference bounding box} [xmin, ymin, xmax, ymax]: [\{xmin\}, \{ymin\}, \{xmax\}, \{ymax\}]
- \textcolor{blue}{Ground-truth target color}: \{target\_color\}

Generate ONE multiple-choice QA sample that strictly follows the requirements below:

1. Inspect the given bounding box region to identify the main object. \textcolor{red}{Replace} ``object'' with a concise object category name (e.g., car, truck, ship, building). \textcolor{red}{Avoid vague} words like ``thing'' or ``object''.

2. Use the following question template verbatim (only substitute the ``object'' placeholder):\\
   ``Determine the color of the object within the given reference bounding box in the image. Image resolution: \{width\} x \{height\}. Bounding box: [\{xmin\}, \{ymin\}, \{xmax\}, \{ymax\}].''

3. Provide four answer options labeled A, B, C, D:
   - Use \{target\_color\} (lowercase) as the \textcolor{red}{only correct} option.
   - Choose three distinct distractor colors that \textcolor{red}{plausibly appear elsewhere} in the image (other instances or surrounding materials). Keep them as \textcolor{red}{simple} lowercase color words (red, blue, green, yellow, white, black, gray, brown, orange, purple).
   - Ensure all four options are \textcolor{red}{unique} and the correct color \textcolor{red}{appears exactly once}.

4. Set the ``answer'' field to the key (A/B/C/D) corresponding to \{target\_color\}.

5. Output JSON ONLY in the structure:\\
\{\\
\quad "question": "...",\\
\quad "options": \{"A": "...", "B": "...", "C": "...", "D": "..."\},\\
\quad "answer": "A\textbar B\textbar C\textbar D"\\
\}
    \end{tcolorbox}
    \caption{Prompt template for color detection.}
    \label{fig:prompt_color}
    \begin{tcolorbox}[
    width=\textwidth,
    enhanced,
    colframe=black,
    boxrule=0.5mm,
    colback=gray!20,
    arc=3mm
]
    \small
    \textbf{Category: Shape/Margin Recognition}\\
    \textbf{Model: Qwen2.5-VL-7B} \\
    \textbf{Prompt:} You are a dataset constructor for Remote Sensing VQA (Shape/Boundary Description).

Image metadata:

- \textcolor{blue}{Image file}: \{image\_path\}
- \textcolor{blue}{Resolution}: \{width\} x \{height\}
- \textcolor{blue}{Reference bounding box} [xmin, ymin, xmax, ymax]: [\{xmin\}, \{ymin\}, \{xmax\}, \{ymax\}]
- \textcolor{blue}{Ground-truth target shape}: \{target\_shape\}

Generate ONE multiple-choice QA sample that strictly follows the requirements below:

1. Focus \textcolor{red}{only} on the main object inside the given bounding box; base your choice on boundary/shape, not color or texture.

2. Use the following question template verbatim (do not change wording, only substitute numbers):\\
   ``Which is the most precise description of the main object's boundary characteristic/shape within the given reference bounding box in the image? Image resolution: \{width\} x \{height\}. Bounding box: [\{xmin\}, \{ymin\}, \{xmax\}, \{ymax\}].''

3. Provide four answer options labeled A, B, C, D:
   - Use \{target\_shape\} (lowercase) as the only correct option.
   - Create three distinct distractors: \textcolor{red}{plausible} shape descriptors that \textcolor{red}{could be seen elsewhere in the image} (other objects) or possible but incorrect impressions inside the box (non-primary elements).
   - Keep options \textcolor{red}{concise}, lowercase words/phrases, e.g., rectangular, square, circular, oval, elongated, triangular, l-shaped, star-shaped, concave, convex, irregular, curved, linear, arc-shaped.
   - Ensure all four options are \textcolor{red}{unique} and the correct shape \textcolor{red}{appears exactly once}.

4. Set the ``answer'' field to the key (A/B/C/D) corresponding to \{target\_shape\}.

5. Output \textcolor{red}{JSON only} in the structure:\\
\{\\
\quad "question": "...",\\
\quad "options": \{"A": "...", "B": "...", "C": "...", "D": "..."\},\\
\quad "answer": "A\textbar B\textbar C\textbar D"\\
\}

\end{tcolorbox}
\caption{Prompt template for shape/margin recognition.}
\label{fig:prompt_shape}
\end{figure*}

\begin{figure*}[t]
\begin{tcolorbox}[
    width=\textwidth,
    enhanced,
    colframe=black,
    boxrule=0.5mm,
    colback=gray!20,
    arc=3mm
]
    \small
    \textbf{Category: Orientation Detection}\\
    \textbf{Model: GPT-5 thinking} \\
    \textbf{Prompt:} You are a remote sensing VQA dataset constructor focusing on orientation detection.

Image metadata:

- \textcolor{blue}{Image }: \{image\_path\}
- \textcolor{blue}{Resolution}: \{width\} x \{height\}
- \textcolor{blue}{Reference bounding box} [xmin, ymin, xmax, ymax]: [\{xmin\}, \{ymin\}, \{xmax\}, \{ymax\}]
- \textcolor{blue}{Ground-truth orientation label} (must be the correct answer): \{target\_orientation\}
- \textcolor{blue}{Allowed orientation words}: \{orientation\_vocab\}

Your task is to create exactly ONE multiple-choice question–answer pair that follows these strict rules:

1. \textcolor{red}{Inspect} the target region inside the bounding box and \textcolor{red}{identify} the main object (e.g., car, ship, storage tank, building). Pick a \textcolor{red}{concise, specific} object category name. Do not use vague words like ``object'' or ``thing''.

2. Use this exact question template (only replace the placeholder with your object phrase):\\
   ``Determine the orientation of the \{\{object\_name\}\} within the given reference bounding box in the image. Image resolution: \{width\} x \{height\}. Bounding box: [\{xmin\}, \{ymin\}, \{xmax\}, \{ymax\}].''
   \textcolor{red}{Replace} \{\{object\_name\}\} with your chosen object.

3. Generate four answer options labeled A, B, C, D. Every option must be a single orientation word selected from \{orientation\_vocab\}.  
   - Insert \{target\_orientation\} as the \textcolor{red}{only correct option}.

4. Set the ``answer'' field to the option key (A/B/C/D) that corresponds to \{target\_orientation\}.

5. Output \textcolor{red}{JSON only} in the structure:\\
\{\\
\quad "question": "...",\\
\quad "options": \{"A": "...", "B": "...", "C": "...", "D": "..."\},\\
\quad "answer": "A\textbar B\textbar C\textbar D"\\
\}
\end{tcolorbox}
\caption{Prompt template for orientation detection.}
\label{fig:prompt_orientation}
\begin{tcolorbox}[
    width=\textwidth,
    enhanced,
    colframe=black,
    boxrule=0.5mm,
    colback=gray!20,
    arc=3mm
]
    \small
    \textbf{Category: Object Classification}\\
    \textbf{Model: Qwen2.5-VL-7B} \\
    \textbf{Prompt:} You are a remote sensing VQA dataset constructor focusing on object classification.

Image metadata:

- \textcolor{blue}{Image file}: \{image\_path\}
- \textcolor{blue}{Resolution}: \{width\} x \{height\}
- \textcolor{blue}{Reference bounding box} [xmin, ymin, xmax, ymax]: [\{xmin\}, \{ymin\}, \{xmax\}, \{ymax\}]
- \textcolor{blue}{Ground-truth target category}: \{target\_label\}
- \textcolor{blue}{Category vocabulary} (valid class names): \{class\_vocab\}

Produce exactly ONE multiple-choice question--answer pair that obeys all rules:

1. Use this question sentence verbatim (numbers already filled in; \textcolor{red}{do not alter} wording or punctuation):\\
   ``Determine the category of the main object within the given reference bounding box in the image. Image resolution: \{width\} x \{height\}. Bounding box: [\{xmin\}, \{ymin\}, \{xmax\}, \{ymax\}].''

2. Options A/B/C/D must be English only and concise (at \textcolor{red}{most $\leq 6$ words} each):  
   - Build the correct option by starting from the label \{target\_label\}. You may prepend \textcolor{red}{at most two adjectives} to refine it (e.g., ``dense \{target\_label\} zone''). The label \{target\_label\} must appear exactly once in the correct option.  
   - Create \textcolor{red}{three distinct distractors} from exactly these two sources:  
   \quad (A) visually grounded objects elsewhere in the same image that are either near the box or highly salient;\\
   \quad (B) semantically confusable classes of the same high-level type as the target.\\
   \textcolor{red}{At least one} distractor must come from (A) and at least one from (B). If (A) is not visibly available, use more from (B).  
   - Distractors must use English descriptors that include one of the labels in \{class\_vocab\}, but must not reuse \{target\_label\}. Do not include any Chinese text. Do not repeat labels or duplicate phrasings across options.

3. The ``answer'' field must be exactly the key (A/B/C/D) pointing to the refined English description containing \{target\_label\}.

4. Output \textcolor{red}{JSON only} in this schema:\\
\{\\
\quad "question": "...",\\
\quad "options": \{"A": "...", "B": "...", "C": "...", "D": "..."\},\\
\quad "answer": "A\textbar B\textbar C\textbar D"\\
\}
\end{tcolorbox}
\caption{Prompt template for object classification.}
\label{fig:prompt_classification}
\end{figure*}

\begin{figure*}[t]
\begin{tcolorbox}[
    width=\textwidth,
    enhanced,
    colframe=black,
    boxrule=0.5mm,
    colback=gray!20,
    arc=3mm
]
    \small
    \textbf{Category: Relative Spatial Relationship}\\
    \textbf{Model: Qwen2.5-VL-7B} \\
    \textbf{Prompt:} You are a remote sensing VQA dataset constructor focusing on object-level relative spatial relationships.

Your goal: From the given image \textbf{alone}, create \textbf{one} multiple-choice question of the type:\\
VQA-Perception $\rightarrow$ Object-Level Attributes $\rightarrow$ Relative Spatial Relationship.

What to do (strict steps):
\begin{enumerate}
    \item Determine the image pixel \textcolor{blue}{resolution} $(W \times H)$.
    \item \textcolor{blue}{Select two} adjacent, salient, clearly visible objects or regions in the image.
    \begin{itemize}
        \item Objects should be distinct, identifiable, and common (e.g., brightly colored or large areas such as a stadium, forest, building, river, red plane).
        \item Each object must be fully visible rather than partially cropped.
        \item There must be a clear distinction between the two objects and a well-defined positional relationship.
        \item Both objects should be at least $10 \times 10$ pixels for clarity.
    \end{itemize}
    \item \textcolor{blue}{Formulate a question} asking about the relative spatial position of Object~A with respect to Object~B using the exact sentence pattern:\\
    ``In the image, where is \textless Object A\textgreater{} located relative to \textless Object B\textgreater{}? Image resolution: \{w\} x \{h\}.''\\
    Describe the characteristics of each object in detail (e.g., ``the red aircraft'', ``the cargo ship in motion'', ``the parking lot with 10 cars'').
\end{enumerate}

Important instructions (must follow):
\begin{itemize}
    \item When describing the two reference objects, \textcolor{red}{do not mention their global positions} in the image (e.g., ``top left'', ``bottom right'').
    \item Instead, describe them using distinctive features such as shape, color, texture, or category (e.g., ``the circular stadium with a red track'', ``the rectangular gray building with a white roof'').
    \item If the two objects are in the same general area (e.g., both in the upper right), do not reveal this; the model should \textcolor{red}{infer only} the relative spatial relationship.
    \item If the objects are very large and cover most of the image, \textcolor{red}{avoid repeating} absolute global positions; \textcolor{red}{emphasize unique} visual traits instead.
    \item \textcolor{red}{Distinguish} each object from other similar ones in the image by differences such as orientation, color, or size (e.g., multiple parking lots differentiated by the number of spaces or nearby structures).
    \item Questions should be grounded in the overall environmental semantics of the scene and \textcolor{red}{involve relatively common objects} such as ships and houses.
\end{itemize}

Example: ``In the image, where is the large parking lot on the left side with 2 planes located relative to the largest airport terminal building? Image resolution: \{w\} x \{h\}.''

Answer options:
\begin{itemize}
    \item Provide exactly four options (A, B, C, D).
    \item Options must be directional relations: ``Above'', ``Below'', ``To the left'', ``To the right'', ``Upper left corner'', ``Upper right corner'', ``Bottom left corner'', ``Bottom right corner''.
    \item Randomize the order of options and ensure exactly one is correct.
\end{itemize}

Quality rules:
\begin{itemize}
    \item The spatial relationship \textcolor{red}{must be unambiguous}.
    \item Do not select overlapping or vague objects.
\end{itemize}

Output format (\textcolor{red}{JSON only}, no extra text):\\
\{\\
\quad "question": "In the image, where is \textless Object A\textgreater{} located relative to \textless Object B\textgreater{}? Image resolution: \{w\} x \{h\}.",\\
\quad "options": \{"A": "\textless direction\textgreater{}", "B": "\textless direction\textgreater{}", "C": "\textless direction\textgreater{}", "D": "\textless direction\textgreater{}"\},\\
\quad "answer": "A\textbar B\textbar C\textbar D"\\
\}

Constraints:
\begin{itemize}
    \item Output only the JSON object.
    \item \{w\} and \{h\} \textcolor{red}{must be integers}.
    \item The answer must be exactly one of A/B/C/D.
\end{itemize}

\end{tcolorbox}
\caption{Prompt template for relative spatial relationship.}
\label{fig:prompt_spatial}

\end{figure*}

\begin{figure*}[t]
\begin{tcolorbox}[
    width=\textwidth,
    enhanced,
    colframe=black,
    boxrule=0.5mm,
    colback=gray!20,
    arc=3mm
]
    \small
    \textbf{Category: Object Grounding}\\
    \textbf{Model: GPT-5 thinking} \\
    \textbf{Prompt:} You are a remote sensing VQA dataset constructor focusing on object grounding.

Image metadata:

- \textcolor{blue}{Image file}: \{image\_path\}
- \textcolor{blue}{Resolution}: \{width\} x \{height\}
- \textcolor{blue}{Candidate bounding boxes} (pixel coordinates, inclusive):\\
\{candidate\_bboxes\}
- \textcolor{blue}{Ground-truth target bounding box} (must be one of the candidates): \{target\_bbox\}

Your task is to create exactly \textcolor{red}{one} multiple-choice question--answer pair that follows every rule:

1. Use this question template verbatim (only substitute the placeholder with the object phrase supplied below):\\
   ``Which bounding box best localizes \{object\_phrase\}, resolution: \{width\} x \{height\}?''

2. Provide four answer options labeled A, B, C, D:
   \begin{itemize}
       \item Each option value must \textcolor{red}{match exactly} one of the candidate bounding box strings listed above.
       \item Use every candidate bounding box exactly once (in any order).
   \end{itemize}

3. Set the ``answer'' field to the option key (A/B/C/D) corresponding to the candidate bounding box that equals \{target\_bbox\}.

4. Output \textcolor{red}{JSON only} in the structure:\\
\{\\
\quad "question": "...",\\
\quad "options": \{"A": "...", "B": "...", "C": "...", "D": "..."\},\\
\quad "answer": "A\textbar B\textbar C\textbar D"\\
\}
\end{tcolorbox}
\caption{Prompt template for object grounding.}
\label{fig:prompt_obj_ground}
\begin{tcolorbox}[
    width=\textwidth,
    enhanced,
    colframe=black,
    boxrule=0.5mm,
    colback=gray!20,
    arc=3mm
]
    \small
    \textbf{Category: Regional Grounding}\\
    \textbf{Model: GPT-5 thinking} \\
    \textbf{Prompt:} You are a remote sensing VQA dataset constructor focusing on regional grounding.

Image metadata:

- \textcolor{blue}{Image file}: \{image\_path\}
- \textcolor{blue}{Resolution}: \{width\} x \{height\}
- \textcolor{blue}{Candidate bounding boxes} (pixel coordinates, inclusive):\\
\{candidate\_bboxes\}
- \textcolor{blue}{Ground-truth target bounding box} (must be one of the candidates): \{target\_bbox\}

Your task is to create exactly \textcolor{red}{one} multiple-choice question--answer pair that follows every rule:

1. Use this question template verbatim (based on the region phrase supplied below and the objects within the candidate boxes, you may make appropriate refinements to substitute the placeholder):\\
   ``Which bounding box best localizes \{region\_phrase\}, resolution: \{width\} x \{height\}?''\\
   Regardless of what language \{region\_phrase\} is in, convert it to \textcolor{red}{full English} in the final question.

2. Provide four answer options labeled A, B, C, D:
   \begin{itemize}
       \item Each option value must \textcolor{red}{match exactly} one of the candidate bounding box strings listed above.
       \item Use every candidate bounding box exactly once (in any order).
   \end{itemize}

3. Set the ``answer'' field to the option key (A/B/C/D) corresponding to the candidate bounding box that equals \{target\_bbox\}.

4. Output \textcolor{red}{JSON only} in the structure:\\
\{\\
\quad "question": "...",\\
\quad "options": \{"A": "...", "B": "...", "C": "...", "D": "..."\},\\
\quad "answer": "A\textbar B\textbar C\textbar D"\\
\}
\end{tcolorbox}
\caption{Prompt template for regional grounding.}
\label{fig:prompt_reg_ground}
\end{figure*}

\begin{figure*}[t]
\begin{tcolorbox}[
    width=\textwidth,
    enhanced,
    colframe=black,
    boxrule=0.5mm,
    colback=gray!20,
    arc=3mm
]
    \small
    \textbf{Category: Object Counting}\\
    \textbf{Model: GPT-5 thinking} \\
    \textbf{Prompt:} You are a dataset constructor for Remote Sensing VQA.

Your goal: From the given image alone, create one multiple-choice question of the type:\\
VQA-Perception $\rightarrow$ Counting \& Measurement $\rightarrow$ Object Counting.

Strict steps:
\begin{enumerate}
    \item Select \textcolor{red}{one} object category that is:
        - clearly visible and individually countable;
        - not too small (\textcolor{red}{at least $10 \times 10$ pixels}), not severely occluded, and unambiguous;
        - described with \textcolor{red}{precise attributes} (e.g., ``large cargo ships'', ``red oval tracks'', ``small white cars in a parking lot'');
        - not vague or overly broad (avoid labels like ``ship'', ``vehicle'', ``building''; instead use specific and distinctive descriptions, e.g., ``big oil tanker'', ``blue-roof two-story buildings'').
    If very small or ambiguous objects exist, ignore them and only select \textcolor{red}{salient, clearly recognizable ones}.
    \item Specify the counting scope:
        - if objects are spread across the entire image, ask about the total number in the whole image;
        - if objects are concentrated in one distinct region (e.g., bottom right), ask about that region.
    The region description must be precise and natural (e.g., ``in the bottom-right corner near the coastline'').
    \item Count the objects precisely.
    \item Construct the question in this format:\\
    ``How many \textless detailed object description\textgreater{} are there \textless in the entire image / in the [specific region]\textgreater{}?''
    \item Provide four answer options (A, B, C, D):

        - each option must be an integer;
        - options must be close (within $\pm 1$--$3$ of the correct count);
        - randomize the position of the correct answer;
        - \textcolor{red}{avoid trivial wide gaps} such as 10/20/30/40.

\end{enumerate}

Output format (\textcolor{red}{JSON only}, no extra text):\\
\{\\
\quad "question": "How many \textless objects\textgreater{} are there ...?",\\
\quad "options": \{"A": \textless int\textgreater{}, "B": \textless int\textgreater{}, "C": \textless int\textgreater{}, "D": \textless int\textgreater{}\},\\
\quad "answer": "A\textbar B\textbar C\textbar D"\\
\}

\end{tcolorbox}
\caption{Prompt template for object counting.}
\label{fig:prompt_obj_count}
\begin{tcolorbox}[
    width=\textwidth,
    enhanced,
    colframe=black,
    boxrule=0.5mm,
    colback=gray!20,
    arc=3mm
]
    \small
    \textbf{Category: Regional Counting}\\
    \textbf{Model: Qwen2.5-VL-7B} \\
    \textbf{Prompt:} You are a dataset constructor for Remote Sensing VQA.

Your goal: From the given image \textbf{alone}, create \textbf{one} multiple-choice question of the type:\\
VQA-Perception $\rightarrow$ Counting \& Measurement $\rightarrow$ Regional Counting.

Strict steps:
\begin{enumerate}
    \item Select \textcolor{red}{one} object category that is:
        - clearly visible and individually countable;
        - not too small (\textcolor{red}{at least $10\times10$ pixels}), not severely occluded, and unambiguous;
        - described with \textcolor{red}{precise attributes} (e.g., ``large cargo ships'', ``red oval tracks'', ``small white cars in a parking lot'');
        - \textcolor{red}{not vague} or overly broad (avoid ``ship'', ``vehicle'', ``building'').
    \item Specify the counting scope:
        - if objects are spread across the entire image, ask about the total number in the whole image;
        - if objects are concentrated in one distinct region (e.g., bottom right), ask about that region.
    The region description must be precise and natural (e.g., ``in the bottom-right corner near the coastline'').
    \item Count the objects precisely.
    \item Construct the question in this format:\\
    ``How many \textless detailed object description\textgreater{} are there \textless in the entire image / in the [specific region]\textgreater{}?''
    \item Provide four answer options (A, B, C, D):
        - each option must be an integer;
        - options must be close (within $\pm 1$--$3$ of the correct count);
        - randomize the position of the correct answer;
        - \textcolor{red}{avoid wide trivial gaps} such as 10/20/30/40.
\end{enumerate}

Output format (\textcolor{red}{JSON only}, no extra text):\\
\{\\
\quad "question": "How many \textless objects\textgreater{} are there ...?",\\
\quad "options": \{"A": \textless int\textgreater{}, "B": \textless int\textgreater{}, "C": \textless int\textgreater{}, "D": \textless int\textgreater{}\},\\
\quad "answer": "A\textbar B\textbar C\textbar D"\\
\}
\end{tcolorbox}
\caption{Prompt template for regional counting.}
\label{fig:prompt_reg_count}
\end{figure*}
\begin{figure*}[t]
\begin{tcolorbox}[
    width=\textwidth,
    enhanced,
    colframe=black,
    boxrule=0.5mm,
    colback=gray!20,
    arc=3mm
]
    \small
    \textbf{Category: Multi-region Joint Contrast (Three-Image Comparison)}\\
    \textbf{Model: GPT-5 thinking} \\
    \textbf{Prompt:} You are a remote sensing VQA dataset constructor for multi-region joint contrast.

Task:
    - You are given \textcolor{blue}{three} satellite images (A, B, C), each with \textcolor{blue}{one reference bounding box} and \textcolor{blue}{one contrast attribute label}.
    - \textcolor{blue}{Compare} only the three reference regions (A/B/C) according to the given attribute.
    - \textcolor{blue}{Produce exactly one multiple-choice question} with fixed options.
    - The correct answer must be A, B, or C.

Image A/B/C:
    - \textcolor{blue}{File}: \{image\_path\_A/B/C\}
    - \textcolor{blue}{Resolution}: \{width\_A/B/C\} x \{height\_A/B/C\}
    - \textcolor{blue}{Reference box} [xmin, ymin, xmax, ymax]: [\{xmin\_A/B/C\}, \{ymin\_A/B/C\}, \{xmax\_A/B/C\}, \{ymax\_A/B/C\}]
    - \textcolor{blue}{Contrast attribute label}: ``\{label\_A/B/C\}''

Question requirements:
\begin{enumerate}
    \item The question should ask which image (A/B/C) shows the \textcolor{red}{extremum} of the attribute indicated by the labels above. For example:\\
    ``Which of the three given boxes in these three images has the highest building density? Resolution: \{width\_A\} x \{height\_A\}. Image A's box: [\{xmin\_A\}, \{ymin\_A\}, \{xmax\_A\}, \{ymax\_A\}]. Image B's box: [\{xmin\_B\}, \{ymin\_B\}, \{xmax\_B\}, \{ymax\_B\}]. Image C's box: [\{xmin\_C\}, \{ymin\_C\}, \{xmax\_C\}, \{ymax\_C\}].''\\
    Here, ``the highest building density'' should be derived from the attribute implied by the labels.
    
        - If the label does not explicitly specify highest/lowest, \textcolor{red}{choose the most natural extremum} (e.g., ``highest'').
        - Keep the question concise (\textcolor{red}{no more than 25 words}).
        - The resolution of the three images is the same, so it is sufficient to mention it once.
        - No matter what language the labels are in, convert them into an \textcolor{red}{English question}.

    \item Options must be fixed strings:
        - A: ``Image A''
        - B: ``Image B''
        - C: ``Image C''
        - D: ``Cannot compare''
    \item Set the ``answer'' to the option key (A/B/C/D) that uniquely matches the correct extremum by inspecting the three images.
\end{enumerate}

Output format (\textcolor{red}{JSON only}, ASCII only, no commentary):\\
\{\\
\quad "question": "...",\\
\quad "options": \{"A": "Image A", "B": "Image B", "C": "Image C", "D": "Cannot compare"\},\\
\quad "answer": "A\textbar B\textbar C"\\
\}
\end{tcolorbox}
\caption{Prompt template for multi-region joint contrast.}
\label{fig:prompt_multi_contrast_three}
\begin{tcolorbox}[
    width=\textwidth,
    enhanced,
    colframe=black,
    boxrule=0.5mm,
    colback=gray!20,
    arc=3mm
]
    \small
    \textbf{Category: Multi-region Joint Contrast (Single Image)}\\
    \textbf{Model: GPT-5 thinking} \\
    \textbf{Prompt:} You are a remote sensing VQA dataset constructor focusing on multi-region joint contrast (single image).

Image metadata:

- \textcolor{blue}{Image file}: \{image\_path\}
- \textcolor{blue}{Resolution}: \{width\} x \{height\}
- \textcolor{blue}{Annotated comparison regions from the XML}:
\{candidate\_regions\}
Each line above has the form:
\quad [xmin, ymin, xmax, ymax] --- ``region description''

Your tasks:
\begin{enumerate}
    \item Read the region descriptions carefully. They already specify the contrast attribute(s) (e.g., building density, traffic volume).
    \item Craft exactly one multiple-choice question comparing these regions. The question must:
        - use the template verbatim (only substitute the braces with your phrase):
        ``Which bounding box best represents the \{\{comparison\_phrase\}\}''
        - ensure the comparison phrase \textcolor{red}{stays faithful to the provided descriptions} (e.g., ``area with the highest building density'', ``cropland patch with the lowest greenness'');
        - if descriptions imply an extremum (highest/lowest/most/least), leverage that; otherwise phrase the comparison naturally (e.g., ``largest cluster of compact housing'');
        - keep the question concise (\textcolor{red}{no more than 25 words}).

    \item Options A, B, C, D:
        - use the bounding box strings exactly as provided (copy-paste). Do not invent new boxes;
        -include each annotated box exactly once across the options;
        - if there are fewer than four annotated boxes, fill the remaining options with the string ``-'' (a single hyphen).
    \item Determine the correct answer by \textcolor{red}{visually comparing} the regions in the image according to the described attribute(s).
        - set ``answer'' to the option key (A/B/C/D) corresponding to the correct bounding box (never ``-'').
    \item Output \textcolor{red}{JSON only} in this structure (no commentary, English only):\\
\{\\
\quad "question": "...",\\
\quad "options": \{"A": "...", "B": "...", "C": "...", "D": "..." \},\\
\quad "answer": "A\textbar B\textbar C\textbar D"\\
\end{enumerate}
\end{tcolorbox}
\caption{Prompt: multi-region joint contrast (single image).}
\label{fig:prompt_multi_contrast_single}
\end{figure*}

\begin{figure*}[t]
\begin{tcolorbox}[
    width=\textwidth,
    enhanced,
    colframe=black,
    boxrule=0.5mm,
    colback=gray!20,
    arc=3mm
]
    \small
    \textbf{Category: Object State Judgement (Multi-round Dialogue)}\\
    \textbf{Model: GPT-5 thinking} \\
    \textbf{Prompt:} You are a dataset constructor for Remote Sensing VQA.

Task: Design a three-round dialogue (user $\leftrightarrow$ assistant) focused on \textbf{Object State Judgement} for a salient object in the image.

Strict requirements:
\begin{enumerate}
    \item Inspect the image carefully and \textcolor{red}{select one} object whose state is unambiguous from visible evidence.
    \item Describe that object with distinctive traits so it cannot be confused with similar items nearby.
    \item Craft \textcolor{red}{three rounds of interaction}:
        - Each round is a single multiple-choice question with four options labeled A/B/C/D, authored by the user.
        - Every option must be grounded in the image. Create three plausible distractors based on other visual cues, plus one correct answer.
        - The assistant responds only with the correct option letter followed by the exact option text (format: ``\textless letter\textgreater. \textless option text\textgreater''). Do not output explanations, reasoning, or any extra words.
        - Round 1: \textcolor{red}{ask about visual cues} that reveal the object's condition. The four options must be obviously different from one another and must not express the same meaning. In the question, indicate the object's approximate position in the picture and describe the object in detail, but do not include any hints of the answer.
        - The correct choice must rely on information that requires looking at the image, not just commonsense.
        - Round 2: \textcolor{red}{ask how} those cues impact functionality or operational status. The four options must be obviously different from one another and must not express the same meaning.
        - Round 3: \textcolor{red}{ask for the state classification}; the four options can reference from the fixed list \texttt{allowed\_states} below. Other options can be added based on commonsense properties of the chosen object, but they must be different from the correct answer and from each other.

\end{enumerate}

\textcolor{blue}{Allowed states}:
"Moving", "Stationary", "In use / Active", "Not in use / Inactive", "Intact", "Damaged", "Under construction", "Occupied", "Empty", "Snow-covered", "Flooded", "Dry", "Vegetated", "Take-off", "Landing", "Parked", "Sailing", "Anchored", "Reversing", "Stopped", "In transit", "Under demolition", "Renovating", "Vacant", "Cruising", "Taxiing", "Departed", "Arrived", "Docked", "Moored", "In dry dock", "Drifting", "Eroded", "Burnt", "Muddy", "Irrigated", "Frozen", "Overgrown", "Drought", "Collapsed", "Unstable", "Protected", "Inaccessible", "Severe damage", "Partial damage", "Fully operational", "Non-operational"

Output format (\textcolor{red}{JSON only}):
\{\\
\quad "target\_object": \{\\
\quad\quad "description": "\textless distinguishing description\textgreater",\\
\quad\quad "state\_evidence": ["\textless key cue 1\textgreater", "\textless key cue 2\textgreater"]\\
\quad\},\\
\quad "conversation": [\\
\quad\quad \{\\
\quad\quad\quad "round": 1,\\
\quad\quad\quad "user\_question": "\textless A-D question\textgreater",\\
\quad\quad\quad "options": \{"A": "\textless option\textgreater", "B": "\textless option\textgreater", "C": "\textless option\textgreater", "D": "\textless option\textgreater"\},\\
\quad\quad\quad "assistant\_answer": "\textless letter\textgreater. \textless option text\textgreater"\\
\quad\quad\},\\
\quad\quad \{\\
\quad\quad\quad "round": 2,\\
\quad\quad\quad "user\_question": "\textless A-D question\textgreater",\\
\quad\quad\quad "options": \{"A": "\textless option\textgreater", "B": "\textless option\textgreater", "C": "\textless option\textgreater", "D": "\textless option\textgreater"\},\\
\quad\quad\quad "assistant\_answer": "\textless letter\textgreater. \textless option text\textgreater"\\
\quad\quad\},\\
\quad\quad \{\\
\quad\quad\quad "round": 3,\\
\quad\quad\quad "user\_question": "\textless A-D question using allowed states\textgreater",\\
\quad\quad\quad "options": \{"A": "\textless state\textgreater", "B": "\textless state\textgreater", "C": "\textless state\textgreater", "D": "\textless state\textgreater"\},\\
\quad\quad\quad "assistant\_answer": "\textless letter\textgreater. \textless state text\textgreater"\\
\quad\quad\}\\
\quad ]\\
\}

\end{tcolorbox}
\caption{Prompt template for object state judgement (multi-round dialogue).}
\label{fig:prompt_state_multi}
\end{figure*}

\begin{figure*}[t]
\begin{tcolorbox}[
    width=\textwidth,
    enhanced,
    colframe=black,
    boxrule=0.5mm,
    colback=gray!20,
    arc=3mm
]
    \small
    \textbf{Category: Object State Judgement (Single turn)}\\
    \textbf{Model: GPT-5 thinking} \\
    \textbf{Prompt:} You are a remote sensing VQA dataset constructor focusing on object state judgement.

Image metadata:
- \textcolor{blue}{Image file}: \{image\_path\}
- \textcolor{blue}{Resolution}: \{width\} x \{height\}
- \textcolor{blue}{Allowed state vocabulary} (use these strings exactly): \{state\_vocab\}

Your task is to create exactly \textcolor{red}{one} multiple-choice question--answer pair following these strict requirements:

1. Inspect the image carefully and select one salient object whose state is visually unambiguous.
       - Describe the object's distinctive appearance (\textcolor{red}{size, shape, color, context}) so it cannot be confused with neighboring objects.
       - Identify the object's approximate location within the image (e.g., ``top-left'', ``lower central area'').
       
2. Compose the question using this template verbatim (\textcolor{red}{only substitute the placeholders}):\\
   ``What is the current state of the \{\{object\_description\}\} located in the \{\{location\_phrase\}\} of the image?''
       - Replace \{object\_description\} with your detailed object description (ensure that \textcolor{red}{no hints} to the correct answer are given based on the detailed description).
       - Replace \{location\_phrase\} with the location phrase (e.g., ``in the upper right corner of the picture'').
       - Do not leak the correct answer in the question text.

3. Provide four answer options labeled A, B, C, D.

       Insert the correct state exactly once; the remaining three must be plausible distractors.
4. Set the ``answer'' field to the option key (A/B/C/D) that corresponds to the correct state.
5. Populate the ``target\_object'' fields:
       - ``description'': the object description from step 1.
       - ``location'': the location phrase from step 1.
       - ``state'': the correct state (identical to the answer option text).
       - ``evidence'': a list of two concise visual cues supporting the state judgement.

6. Output \textcolor{red}{JSON only} in the structure below (no extra words, explanations, or markdown):\\
\{\\
 "target\_object": \{
\quad "description": "...",
\quad "location": "...",
\quad "state": "...",
\quad "evidence": ["...", "..."]
\},\\
\quad "question": "...",\\
\quad "options": \{"A": "...", "B": "...", "C": "...", "D": "..." \},\\
\quad "answer": "A\textbar B\textbar C\textbar D"\\
\}

\end{tcolorbox}
\caption{Prompt template object sstate judgement (single turn).}
\label{fig:prompt_state_single}
\begin{tcolorbox}[
    width=\textwidth,
    enhanced,
    colframe=black,
    boxrule=0.5mm,
    colback=gray!20,
    arc=3mm
]
    \small
    \textbf{Category: Anomaly detection and interpretation}\\
    \textbf{Model: GPT-5 thinking} \\
    \textbf{Prompt:} You are a remote sensing VQA assistant performing Anomaly detection and interpretation for an image.

Rules:
\begin{enumerate}
    \item Carefully observe the image and select \textcolor{red}{one} object whose state is unambiguous.
        - Describe the object with distinguishing traits (e.g., \textcolor{red}{size, shape, color, position}).
        - Provide \textcolor{red}{the approximate location of the object in the image} (e.g., ``top-left'').

    \item In your question, include a detailed description of the selected object based on its visual features, including its position in the image.
        - Example: ``The red cargo ship parked on the riverbank, with a blue warehouse visible on the right side of the image. Where is this cargo ship located in the image?''
        - Make sure the question and the description are \textcolor{red}{clearly related to the visual elements} in the image.
        - The question should not contain any answer hints.
    \item The assistant should then ask about the \textcolor{red}{most likely reason or cause} for the object’s current state or anomaly, based on visible cues from the image.
         For example: ``What could be the most likely reason for the cargo ship being parked at the riverbank?''

    \item Provide four options (A, B, C, D), where:
        - One option should be the \textcolor{red}{most plausible cause}, based on visual evidence from the image.
        - The other three options are reasonable but incorrect distractors.

    \item Make sure the answer to the question depends on specific visual cues in the image. The answer must not rely on commonsense knowledge or external factors not shown in the image.
\end{enumerate}

Output format (\textcolor{red}{JSON only}, no extra text):\\[2pt]
\{\\
\quad "question": "\textless your detailed question including object description and location\textgreater",\\
\quad "options": \{
\quad\quad "A": "...",
\quad "B": "...",
\quad"C": "...",
\quad "D": "..."
\quad\},\\
\quad "answer": "A\textbar B\textbar C\textbar D"\\
\}\\[2pt]
\end{tcolorbox}
\caption{Prompt template for anomaly detection.}
\label{fig:prompt_anomaly_single}
\end{figure*}

\begin{figure*}[t]
\begin{tcolorbox}[
    width=\textwidth,
    enhanced,
    colframe=black,
    boxrule=0.5mm,
    colback=gray!20,
    arc=3mm
]
    \small
    \textbf{Category: Anomaly
    detection and Interpretation(Three-round Dialogue)}\\
    \textbf{Model: GPT-5 thinking} \\
    \textbf{Prompt:} You are a dataset constructor for Remote Sensing VQA. Task: Design a three-round dialogue (user $\leftrightarrow$ assistant) focused on Anomaly
    etection and Interpretation for a salient object or region in the satellite image.

Requirements:
\begin{enumerate}
    \item \textcolor{blue}{Inspect the image and select one object} or region that is suitable for future prediction based on its clear visual state and surroundings.
        - The selected object or region should be clearly identifiable and in a distinct position in the image.
        - Try to choose objects that are relatively unique rather than those that have many similar counterparts in the picture.
        - Try to choose objects that can be observed in detail at the current resolution, whose detail cannot be detected at lower resolution.

    \item \textcolor{blue}{First round (localization)}. Provide a detailed description of the object or region, including its distinguishing features and approximate location (e.g., top-left, right side).
        - The description must ensure that the object is unique and cannot simultaneously refer to multiple objects in the image.
        - The question should be: ``Where is the [object/region] located in the image?'' (use this wording).
        - The description of the [object/region] in the first-round question must be extremely detailed in order to distinguish it from other similar objects.
        - The description should contain the characteristics of the object and indicate its relationship with surrounding objects to make the description more detailed.
        - The description must be \textcolor{red}{more than five words} and \textcolor{red}{fewer than 50 words}, incorporating the inherent characteristics of the object (limited to describing color, shape, size, and orientation; it cannot describe the state) and the spatial relationships with its surrounding objects.
        - \textcolor{red}{Avoid overly simple descriptions} such as ``the cargo ship''.
        - The answer options should be the object/region’s position in the image. They should \textcolor{red}{only} describe the position in the picture without mentioning other contents, such as the relationship with other objects in the answer. Examples:
        ``The left side of the image.''  ``Avoid answers such as: ``The left side of the image, near the river bank.''

    \item \textcolor{blue}{Second round (state analysis).} Analyze the state of the object you have identified. Based on visible evidence from the image, consider the possible states or conditions the object might be in. The focus here is on understanding the current condition or state of the object, which will help inform the possible cause in the next round.

        - Question example: ``What could explain the current state or condition of the [object/region] in the image?''
        - Option examples:
             A. The object is experiencing a common situation, and no anomalies are observed.
            B. The object appears to be stationary due to external factors like weather or environment.
            C. The object is in an abnormal state, possibly due to mechanical failure or environmental changes.
            D. The object is part of a routine operation, with no significant changes.

        - This round does not yet provide the final cause; it is aimed at narrowing down possible scenarios \textcolor{red}{based on visual cues}.

    \item \textcolor{blue}{Third round (cause determination).} Based on your previous analysis, determine the most likely cause for the object’s current state or condition. Consider the visual evidence in the image and reasoning \textcolor{red}{from the previous rounds} to formulate your answer.

        - Question: ``Given the object's current state, what is the most likely cause for this condition?''
        - Options:
             A. [Most plausible reason based on visible evidence].
             B. [Plausible distractor 1].
             C. [Plausible distractor 2].
             D. [Plausible distractor 3].

        - This round should focus on providing the most likely cause based on visual evidence observed earlier, not on external knowledge.
        - The answer should be based on the visual cues and state analysis from the previous rounds, not external knowledge.

\end{enumerate}

The assistant responds only with the correct letter and option text (``\textless letter\textgreater. \textless option text\textgreater''). 

Output format (\textcolor{red}{JSON only}, no extra text):\\[2pt]
\{\\
\quad "question\_1": \{\\
\quad\quad "user\_question": "Where is the [object/region] located in the image?",\\
\quad\quad "options": \{ "A": "...", "B": "...", "C": "...", "D": "..." \},\\
\quad\quad "correct\_answer": "\textless letter\textgreater. \textless correct option text\textgreater"\\
\quad\},\\
\quad "question\_2": \{\\
\quad\quad "user\_question": "What relationship or change might affect the selected object/region in the near future?",\\
\quad\quad "options": \{ "A": "...", "B": "...", "C": "...", "D": "..." \},\\
\quad\quad "correct\_answer": "\textless letter\textgreater. \textless correct option text\textgreater"\\
\quad\},\\
\quad "question\_3": \{\\
\quad\quad "user\_question": "If [answer from second round] happens, what is the most likely future state of the selected object/region?",\\
\quad\quad "options": \{ "A": "...", "B": "...", "C": "\...", "D": "..." \},\\
\quad\quad "correct\_answer": "\textless letter\textgreater. \textless correct option text\textgreater"\\
\quad\}\\
\}\\[2pt]
\end{tcolorbox}
\caption{Prompt template for anomaly detection (multi-round dialogue).}
\label{fig:prompt_anomaly_multi}
\end{figure*}

\begin{figure*}[t]

\end{figure*}
\begin{figure*}[t]
\begin{tcolorbox}[
    width=\textwidth,
    enhanced,
    colframe=black,
    boxrule=0.5mm,
    colback=gray!20,
    arc=3mm
]
    \small
    \textbf{Category: Future Prediction (Three-round Dialogue)}\\
    \textbf{Model: GPT-5 thinking} \\
    \textbf{Prompt:} You are a dataset constructor for Remote Sensing VQA. Task: Design a three-round dialogue (user $\leftrightarrow$ assistant) focused on Future Prediction for a salient object or region in the satellite image.

Requirements:
\begin{enumerate}
    \item \textcolor{blue}{Inspect the image and select one object or region} that is suitable for traceability analysis based on its clear visual state and surroundings.
        - The selected object or region should be clearly identifiable and in a distinct position in the image.
        - Prefer objects that are relatively unique rather than those with many similar counterparts in the picture.
         Prefer objects that can be observed in sufficient detail at the current resolution, where such detail would \textcolor{red}{not be visible at lower resolution}.

    \item \textcolor{blue}{First round (localization)}:

        - Provide a detailed description of the object or region, including its distinguishing features and approximate location (e.g., top-left, right side).
        - The description must ensure that the object is unique and cannot simultaneously refer to multiple objects in the image.
        - The question should be: ``Where is the [object/region] located in the image?'' (use this wording).
        - The description of the [object/region] in the first-round question must be extremely detailed in order to distinguish it from other similar objects.
        - The description must:
            1. \textcolor{red}{contain the characteristics of the object} and \textcolor{red}{indicate its relationship with surrounding objects} to make the description more detailed;2.be \textcolor{red}{more than 5 words} and \textcolor{red}{fewer than 50 words};3.incorporate inherent characteristics of the object (limited to describing color, shape, size, and orientation; do not describe the state);4.include spatial relationships with surrounding objects;
            5.avoid overly simple descriptions such as ``the cargo ship''.
        \item The answer options should be the object/region’s position in the image, describing only the position (without mentioning other content such as relations to other objects). Examples:
            `` The left side of the image.''

    \item \textcolor{blue}{Second round (relationship/change)}:
        - Analyze the state of the object or region you have identified. Based on visible evidence, consider the possible relationships or changes that might affect it in the near future.
        - The focus is on understanding what relationship or change may impact the selected object or region, \textcolor{red}{guided by visual cues} in the image.
        - Example question prototype:
        ``What relationship or change might affect the selected object/region in the near future, based on the image's visual cues?''
        - Options should be four distinct, non-overlapping descriptions of possible relationships or changes (A, B, C, D).

    \item \textcolor{blue}{Third round (future state)}:

        - Based on the previous analysis, determine the most likely future state of the selected object or region if the relationship or change from the second round occurs.
        - Example question prototype:
        ``If [correct answer from second round] happens, what is the most likely future state of the selected object/region?''
        - Options:
             A: possible future state A;
             B: possible future state B;
             C: possible future state C;
             D: possible future state D.
        - The correct option should be the most plausible future state \textcolor{red}{based on the visible evidence} and \textcolor{red}{the reasoning from the previous rounds}, not external knowledge.

\end{enumerate}

Output format (\textcolor{red}{SON only}, no extra text):\\[2pt]
\{\\
\quad "question\_1": \{\\
\quad\quad "user\_question": "Where is the [object/region] located in the image?",\\
\quad\quad "options": \{ "A": "...", "B": "...", "C": "...", "D": "..." \},\\
\quad\quad "correct\_answer": "\textless letter\textgreater. \textless correct option text\textgreater"\\
\quad\},\\
\quad "question\_2": \{\\
\quad\quad "user\_question": "What relationship or change might affect the selected object/region in the near future?",\\
\quad\quad "options": \{ "A": "...", "B": "...", "C": "...", "D": "..." \},\\
\quad\quad "correct\_answer": "\textless letter\textgreater. \textless correct option text\textgreater"\\
\quad\},\\
\quad "question\_3": \{\\
\quad\quad "user\_question": "If [ answer from second round] happens, what is the most likely future state of the selected object/region?",\\
\quad\quad "options": \{ "A": "...", "B": "...", "C": "...", "D": "\..." \},\\
\quad\quad "correct\_answer": "\textless letter\textgreater. \textless correct option text\textgreater"\\
\quad\}\\
\}\\[2pt]
\end{tcolorbox}
\caption{Prompt template for future prediction (multi-round dialogue).}
\label{fig:prompt_future_multi}
\end{figure*}

\begin{figure*}[t]
\begin{tcolorbox}[
    width=\textwidth,
    enhanced,
    colframe=black,
    boxrule=0.5mm,
    colback=gray!20,
    arc=3mm
]
    \small
    \textbf{Category: Future Prediction (Two-Image Temporal Comparison)}\\
    \textbf{Model: GPT-5 thinking} \\
    \textbf{Prompt:} You are a dataset constructor for Remote Sensing VQA. Two satellite images of the \textcolor{red}{same location at different times} are provided (\textcolor{blue}{Image A = earlier year}, \textcolor{blue}{Image B = later year}). Design \textcolor{red}{one} multiple-choice question that requires comparing these images to predict the object's future state.

Requirements:
\begin{enumerate}
    \item Carefully study both images and identify \textcolor{red}{one} salient object or region that is clearly visible in both Image~A and Image~B.
    \begin{itemize}
        \item Choose an object or region whose change over time is visually noticeable and \textcolor{red}{subtle}. Do not rely on commonly known facts; the change must be clearly observed in the images themselves.
        \item Avoid selecting objects that appear frequently or that are trivial to identify without carefully inspecting the images.
    \end{itemize}

    \item Compose a question that:
    \begin{itemize}
        \item references the chosen object or region using its role or approximate location, explicitly naming ``Image~A'' and ``Image~B'', but \textcolor{red}{without} explicitly describing visual features such as colors, shapes, or textures;
        \item asks about the object's most plausible future state, based on visual changes observed between the two images;
        \item instructs the responder to base the answer on visual cues from both images, not on external knowledge.
    \end{itemize}

    \item Provide four answer options (A, B, C, D):
    \begin{itemize}
        \item one option must describe the \textcolor{red}{most plausible future outcome}, grounded in the visual evidence observed in the two images;
        \item the other three options should be \textcolor{red}{reasonable but incorrect distractors}, focusing on future states or trends that are \textcolor{red}{not} supported by the visual evidence;
        \item keep the options concise, focusing on future predictions, while embedding visual observations that support the choice;
        \item the correct answer should incorporate visual cues from both images and explain the future prediction.\\
        Example: ``The observed decrease in green space near the road and the increase in settlements suggest that the settlements are likely to expand further.''
    \end{itemize}

    \item Ensure the answer depends on contrasting Image~A and Image~B. The question should only be answerable by directly comparing the two images, relying on subtle visual evidence to infer future predictions.

    \item Output must be valid \textcolor{red}{JSON only}:
\end{enumerate}

\{\\
\quad "question": "\textless future prediction question including visual changes from both images\textgreater",\\
\quad "options": \{\\
\quad\quad "A": "\textless future state option based on visual cues\textgreater",\\
\quad\quad "B": "\textless future state option\textgreater",\\
\quad\quad "C": "\textless future state option\textgreater",\\
\quad\quad "D": "\textless future state option\textgreater"\\
\quad\},\\
\quad "answer": "A\textbar B\textbar C\textbar D"\\
\}\\[2pt]
\end{tcolorbox}
\caption{Prompt template for future prediction (two-image temporal comparison).}
\label{fig:prompt_future_twoimg}
\end{figure*}

\begin{figure*}[t]
\begin{tcolorbox}[
    width=\textwidth,
    enhanced,
    colframe=black,
    boxrule=0.5mm,
    colback=gray!20,
    arc=3mm
]
    \small
    \textbf{Category: Answer Evaluation (Open-Ended VQA, No Options)}\\
    \textbf{Model: GPT-5 thinking} \\
    \textbf{Prompt:} You are an expert judge for open-ended Visual Question Answering (no options). Treat the reference answer as image-derived ground truth. Avoid rewarding unsupported specifics; however, minor non-contradictory elaborations are acceptable and should not be harshly penalized.\\[4pt]

    Assign a score from \textcolor{red}{1--100} based on:
    \begin{enumerate}
        \item \textcolor{blue}{Correctness/consistency} with the reference (most important).
        \item \textcolor{blue}{Usefulness/completeness}: directly answers the question and includes necessary attributes/units when relevant.
        \item \textcolor{blue}{Hallucinations/fabrication}: penalize only when details are unsupported \emph{and} material to the answer; harmless extra context is neutral.
    \end{enumerate}

    Lenient matching guidelines:
    \begin{itemize}
        \item \textcolor{blue}{Semantic equivalence} counts as correct: synonyms, common paraphrases, close color groups (e.g., teal $\approx$ blue-green; maroon $\approx$ dark red), and reasonable directional mappings (top $\approx$ upper; left $\approx$ west when consistent with image coordinates).
        \item \textcolor{blue}{Numeric/count tolerance} (lenient): if the reference is an integer $n$, allow \textcolor{red}{$\pm1$} when \textcolor{red}{$n \le 10$}; when \textcolor{red}{$n > 10$}, allow approximately \textcolor{red}{$\pm20\%$} rounding.
        \item \textcolor{blue}{Ranges/approximations}: answers that fall within a sensible range around the reference are acceptable.
        \item \textcolor{blue}{Units}: correct conversions or obvious/implicit units are acceptable; missing a unit is a minor deduction unless it changes meaning.
        \item \textcolor{blue}{Yes/No questions}: must match the reference; hedged but correct (e.g., ``likely yes'') is acceptable.
        \item If the reference is ``unknown/indeterminate/not inferable from the image'', answers \textcolor{red}{acknowledging uncertainty} are preferred.
        \item \textcolor{red}{Do not over-penalize} style/grammar; focus on factual content and alignment with the reference.
    \end{itemize}

    Suggested (non-binding) score bands:
    \begin{itemize}
        \item 80--100: semantically equivalent and \textcolor{red}{precise}; no material hallucinations; \textcolor{red}{needed info present}.
        \item 70--80: \textcolor{red}{mostly correct}; minor omissions or small deviations; no contradictions.
        \item 60--70: \textcolor{red}{partially correct}; captures some key info or close numeric value; minor issues.
        \item 30--60: \textcolor{red}{largely incorrect}, contradicts reference, or contains clear/meaningful hallucinations.
        \item 0: empty, off-topic, or \textcolor{red}{unacceptable} content.
    \end{itemize}

    Question: \{q\}\\
    Reference answer (treat as image ground truth): \{ref\}\\
    Model response: \{ans\}\\[4pt]

    Return exactly two lines:\\
    \texttt{score: \textless1-100\textgreater}
\end{tcolorbox}
\caption{Prompt template for Open-Ended VQA Answer Evaluation (Scoring 1--100).}
\label{score_1-100}
\end{figure*}

\begin{figure*}[t]
\begin{tcolorbox}[
    width=\textwidth,
    enhanced,
    colframe=black,
    boxrule=0.5mm,
    colback=gray!20,
    arc=3mm
]
    \footnotesize
    \textbf{Category: Global Scene Description (Single Aerial/Satellite Image)}\\
    \textbf{Model: GPT-5 thinking} \\
    \textbf{Prompt:} Given a high-resolution aerial/satellite image covering a large area with urban, rural, industrial, transport, and natural elements, produce a detailed scene description following the rules below.\\[2pt]

    \textcolor{blue}{1. Overall Description:}
    Write \textcolor{red}{one sentence} summarizing the dominant land-use/landform mix, spatial layout, and main functions. Mention the \textcolor{red}{relative positions} and roles of key regions such as urban areas, farmland, industrial zones, transport corridors, ports, and green spaces (e.g., coastal layout, parallel belts along a river, clustered urban core with surrounding agriculture).\\[2pt]

    \textcolor{blue}{2. Subregion Description:}
    Describe the upper-left, upper-right, lower-left, and lower-right quadrants. For each quadrant, use \textcolor{red}{1--3 sentences} to state the main objects, their spatial organization, and the dominant function or usage (e.g., irrigated agriculture, port logistics, dense residential core, mixed industry/residential). Explicitly mention relationships such as ``roads extending along the coastline'', ``residential blocks separated from fields by a highway'', or ``factories clustered beside a river''. Use concrete phrases like ``multiple factories'', ``a major highway'', or ``several residential complexes'' and avoid vague terms such as ``many'' or ``some''.\\[2pt]

    \textcolor{blue}{3. Detailed Description Requirements:}
    \textit{Object classification:} Cover both \textcolor{red}{natural} landscapes (rivers, mountains, lakes, forests, wetlands) and \textcolor{red}{man-made} structures (roads, bridges, railways, airports, ports, industrial parks, residential areas, commercial zones).\\
    \textit{Spatial relationships:} Clearly describe how objects are arranged, such as roads extending along a coastline, buildings parallel to a railway, grid-like intersections of streets and blocks, houses scattered on hilly terrain, or factories clustered along a waterfront.\\
    \textit{Function/Usage:} Infer the use of each area based on visible forms, such as crop-production farmland, city centers combining residential and commercial buildings, industrial zones for production and storage, or ports used as cargo distribution hubs. Prefer concrete expressions like ``multiple large factories and several container yards'' or ``sparsely distributed residential areas'' instead of vague quantifiers.\\[2pt]

    \textcolor{blue}{4. Output Language:}
    Provide the description in clear and concise \textcolor{red}{English}, focusing on key elements, their spatial layouts, and relationships. Avoid overly general wording and describe what is visibly supported by the scene and its inferred functions.\\[2pt]

    Example Output (for reference):\\
    \emph{Overall Description:} A coastal port city occupies the central area, with a busy deep-water terminal along the southern shoreline, a highway running parallel to the coast linking the port and inland industrial belt, and surrounding residential and commercial districts framed by farmland and natural vegetation to the north and west.\\
    \emph{Subregion Description:} In the \textcolor{red}{upper-left} quadrant, rectangular agricultural fields dominate, framed by straight irrigation canals running roughly north--south, with scattered farmhouses along field edges indicating intensive crop cultivation. The \textcolor{red}{upper-right} quadrant contains a dense residential and commercial district organized in a grid-like street pattern, with several embedded parks and a large river cutting through the area and connecting the city core to the port. The \textcolor{red}{lower-left} quadrant is an industrial zone with large factories laid out in parallel rows, linked by a network of roads and railways to the waterfront and flanked by warehouses along the outer edges of the complex. The \textcolor{red}{lower-right} quadrant shows suburban housing clusters along curving local roads, interspersed with green spaces and small parks, while a narrow river meanders through the area and joins the larger water body near the port.
\end{tcolorbox}
\caption{Prompt: Global Scene and Subregion Description for Aerial/Satellite Images.}
\label{image_caption}
\end{figure*}

\begin{figure*}[t]
\begin{tcolorbox}[
    width=\textwidth,
    enhanced,
    colframe=black,
    boxrule=0.5mm,
    colback=gray!20,
    arc=3mm
]
    \small
    \textbf{Category: Structured Captioning (Quadrant-Based Scene Description)}\\
    \textbf{Model: GPT-5 thinking} \\
    \textbf{Prompt:} You are an expert remote-sensing analyst. For each high-resolution aerial/satellite image, produce a structured caption that strictly follows the template below (do not add extra sections):\\[4pt]

    Overall Summary:\\
    - [\textcolor{red}{One sentence} describing the dominant land-use/landform mix, spatial layout, and primary functions.]\\[2pt]

    Upper-Left Quadrant:\\
    - [\textcolor{red}{Two to three sentences} detailing the key natural/man-made features, their spatial organization, explicit relationships (e.g., ``canals frame rectangular paddies''), and inferred usage.]\\[2pt]

    Upper-Right Quadrant:\\
    - [Same level of detail for this quadrant.]\\[2pt]

    Lower-Left Quadrant:\\
    - [Same level of detail for this quadrant.]\\[2pt]

    Lower-Right Quadrant:\\
    - [Same level of detail for this quadrant.]\\[4pt]

    Object + Usage Breakdown:\\
    - \textcolor{red}{Natural}: [List rivers, forests, hills, wetlands, coastlines, etc., plus their roles.]\\
    - \textcolor{red}{Man-made}: [List roads, railways, bridges, ports, factories, warehouses, housing forms, etc., plus their roles.]\\[6pt]

    Guidelines:
    \begin{enumerate}
        \item Mention \textcolor{red}{both natural landscapes and built infrastructure}, emphasizing spatial relationships (parallel roads along rivers, clusters of warehouses near docks, farmland separated by tree belts, etc.).
        \item For every quadrant, include the dominant \textcolor{red}{function or usage} (e.g., commercial logistics, irrigated agriculture, dense residential core, mixed industry/residential).
        \item \textcolor{red}{Avoid vague quantifiers} such as ``some'' or ``many''; use concrete descriptors like ``five elongated warehouses'', ``dense grid of low-rise housing'', or ``narrow river meandering east--west''.
        \item Language must be \textcolor{red}{English}, concise but specific, referencing relative positions between objects and quadrants where relevant.
        \item Never include placeholder text, apologies, or references to ``the prompt'' or ``the image''; \textcolor{red}{focus solely on observable evidence and inferred usage}.
    \end{enumerate}
\end{tcolorbox}
\caption{Prompt: Quadrant-Based Structured Captioning for High-Resolution Remote-Sensing Imagery.}
\label{image_caption_four_corner}
\end{figure*}





\end{document}